\documentclass[hidelinks]{article}

\newif\ifanonymous

\usepackage{arxiv}

\usepackage{blindtext}
\usepackage{graphicx}
\usepackage{hyperref}
\usepackage{url}
\usepackage{optidef}
\usepackage{times}
\usepackage{caption}
\usepackage{subcaption}
\usepackage[super]{nth}
\usepackage{ifdraft}
\usepackage{multirow}
\usepackage{placeins}
\usepackage{comment}
\usepackage{tabularx}
\usepackage[colorinlistoftodos,prependcaption]{todonotes}
\usepackage{amssymb}
\usepackage{amsthm}
\usepackage{algorithm}
\usepackage{algpseudocode}

\graphicspath{{./figures/}}
\usepackage{amsmath,amsfonts,bm,mathtools,xargs}

\def\Tableref#1{Table~\ref{#1}}
\def\Twotablerefs#1#2{Table~\ref{#1} and~\ref{#2}}
\def\Figref#1{Figure~\ref{#1}}

\def\Twofigref#1#2{Figures~\ref{#1} and~\ref{#2}}
\def\Threefigref#1#2#3{Figures~\ref{#1},~\ref{#2}, and~\ref{#3}}

\def\Figrangeref#1#2{Figures~\ref{#1} to~\ref{#2}}
\def\Figrangetworef#1#2{Figures~\ref{#1}~and~\ref{#2}}

\def\Secref#1{Section~\ref{#1}}
\def\Twosecrefs#1#2{Sections~\ref{#1} and~\ref{#2}}
\def\Secrefs#1#2#3{Sections~\ref{#1},~\ref{#2} and~\ref{#3}}
\def\eqref#1{equation~\ref{#1}}

\def\Eqref#1{Equation~\ref{#1}}

\def\Appref#1{Appendix~\ref{#1}}

\def\Twoappref#1#2{Appendices~\ref{#1} and~\ref{#2}}
\def\1{\bm{1}}

\def\rvx{{\mathbf{x}}}

\def\rvz{{\mathbf{z}}}

\def\vone{{\bm{1}}}

\def\vtheta{{\bm{\theta}}}
\def\vphi{{\bm{\phi}}}

\def\vx{{\bm{x}}}

\def\vz{{\bm{z}}}

\def\mA{{\bm{A}}}
\def\mB{{\bm{B}}}
\def\mC{{\bm{C}}}
\def\mD{{\bm{D}}}

\def\mI{{\bm{I}}}

\def\mQ{{\bm{Q}}}

\def\mX{{\bm{X}}}
\def\mY{{\bm{Y}}}

\DeclareMathAlphabet{\mathsfit}{\encodingdefault}{\sfdefault}{m}{sl}
\SetMathAlphabet{\mathsfit}{bold}{\encodingdefault}{\sfdefault}{bx}{n}

\newcommand{\N}{\mathcal{N}}
\newcommand{\E}{\mathbb{E}}

\newcommand{\R}{\mathbb{R}}

\newcommand{\KL}{D_{\mathrm{KL}}}
\newcommandx{\D}[5][1,2]{D^{#1}_{#2{#3}}(#4||#5)}

\DeclareMathOperator{\ELBO}{\mathcal{L}(\vtheta, \vphi; \rvx)}
\DeclarePairedDelimiter\norm{\lVert}{\rVert}

\newcommand{\coo}{\ensuremath{\mathrm{CO_2}}}
\newcommandx{\suggestion}[2][1=]{\todo[linecolor=ProcessBlue,backgroundcolor=ProcessBlue!25,bordercolor=ProcessBlue,#1]{#2}}
\newcommandx{\donelast}[2][1=]{\todo[linecolor=gray,backgroundcolor=gray!25,bordercolor=gray,#1]{#2}}
\renewcommand\labelenumi{(\roman{enumi})}
\renewcommand\theenumi\labelenumi
\anonymousfalse

\title{How do Variational Autoencoders Learn?\\ Insights from Representational Similarity}

\author{Lisa Bonheme \& Marek Grzes
    School of Computing\\
    University of Kent\\
    Canterbury, UK\\
    \texttt{\{lb732, m.grzes\}@kent.ac.uk}
}

\begin{document}
    \maketitle

    \begin{abstract}
        The ability of Variational Autoencoders (VAEs) to learn disentangled representations has made them popular for
        practical applications. However, their behaviour is not yet fully understood. For example, the questions of when they
        can provide disentangled representations, or suffer from posterior collapse are still areas of active research.
        Despite this, there are no layerwise comparisons of the representations learned by VAEs, which would further our understanding of these models.
        In this paper, we thus look into the internal behaviour of VAEs using representational similarity techniques.
        Specifically, using the CKA and Procrustes similarities, we found that the encoders' representations
        are learned long before the decoders', and this behaviour is independent of hyperparameters, learning objectives, and datasets.
        Moreover, the encoders' representations in all but the mean and variance layers are similar across hyperparameters and learning objectives.
    \end{abstract}

    \section{Introduction}\label{sec:intro}

Variational Autoencoders (VAEs) are considered state-of-the-art techniques to learn unsupervised disentangled representations,
and have been shown to be beneficial for fairness~\citep{Locatello2019b}.
As a result, VAEs producing disentangled representations have been extensively studied in the last few years~\citep{Locatello2019a,Mathieu2019,Rolinek2019,Zietlow2021},
but they still suffer from poorly understood issues such as posterior collapse~\citep{Dai2020}.
~While some work using explainability techniques has been done to shed light on the behaviour of VAEs~\citep{Liu2020},
a comparison of the representations learned by different methods is still lacking~\citep{Zietlow2021}.~Moreover,
the layer-by-layer similarity of the representations within models has yet to be investigated.\par

Fortunately, the domain of deep representational similarity is an active area of research
and metrics such as SVCCA~\citep{Raghu2017,Morcos2018}, Procrustes distance~\citep{Schonemann1966}, or Centred Kernel
Alignment (CKA)~\citep{Kornblith2019} have proven very useful in analysing the learning dynamics of various
models~\citep{Wang2019a,Kudugunta2019,Raghu2019,Neyshabur2020}, and even helped to design UDR~\citep{Duan2020},
an unsupervised metric for model selection for VAEs.\par

The fact that good models are more similar to each other than bad ones in the context of classification~\citep{Morcos2018}
generalised well to Unsupervised Disentanglement Ranking (UDR)~\citep{Rolinek2019,Duan2020}.
However, such a generalisation may not always be possible, and without sufficient evidence, it would be wise to expect substantial
differences between the representations learned by supervised and unsupervised models.
~In this paper, our aim is to take a first step toward investigating the representational similarity of generative models
by analysing the similarity scores obtained for a variety of VAEs learning disentangled representations, and by providing some
insights into why various VAE-specific methods preventing posterior collapse~\citep{Bowman2016,He2019} or providing
better reconstruction~\citep{Liu2021} are successful.\par
%the properties specific to VAEs, such as posterior collapse.\par

Our contributions are as follows:
\begin{enumerate}
    \itemsep 0em
    \item We provide the first experimental study of the representational similarity between VAEs, and have released more than
          45 million similarity scores
          \ifanonymous(\url{https://t.ly/0GLe3}\footnote{Due to their size and to preserve anonymity, the 300 models trained for this paper will be released after the review.}\else and 300 trained models specifically designed for measuring representational similarity (\url{https://data.kent.ac.uk/428/} and \url{https://data.kent.ac.uk/444/}, respectively\fi).
    \item We have released the library created for this experiment (\ifanonymous\url{https://t.ly/VMIm}\else\url{https://github.com/bonheml/VAE_learning_dynamics}\fi).
          It can be reused with other similarity metrics or models for further research in the domain.
    \item During our analysis, we found that (1) the encoder is learned before the decoder; (2) all the layers of the encoder,
          except the mean and variance layers, learn very similar representations regardless of the learning objective and regularisation strength used; and
          (3) linear CKA could be an efficient tool to track posterior collapse.
\end{enumerate}

    \section{Background}\label{sec:background}

\subsection{Variational Autoencoders}\label{subsec:bg-VAEs}
Variational Autoencoders (VAEs)~\citep{Kingma2013,Rezende2015}
are deep probabilistic generative models based on variational inference.~The encoder, $q_\vphi(\rvz|\rvx)$, maps some input $\rvx$ to
a latent representation $\rvz$, which the decoder, $p_\vtheta(\rvx|\rvz)$, uses to attempt to reconstruct $\rvx$.
This can be optimised by maximising $\mathcal{L}$, the evidence lower bound (ELBO)
\begin{equation}
    \label{eq:elbo}
    \ELBO = \underbrace{\E_{q_\vphi(\rvz|\rvx)}[\log p_\vtheta(\rvx|\rvz)]}_{\text{reconstruction term}} -
    \underbrace{\KL\left(q_\vphi(\rvz|\rvx) || p(\rvz)\right)}_{\text{regularisation term}},
\end{equation}
where $p(\rvz)$ is generally modelled as a multivariate Gaussian distribution $\N(0, \mI)$ to permit closed
form computation of the regularisation term~\citep{Doersch2016}. We refer to the regularisation term of \Eqref{eq:elbo} as regularisation in the rest of the paper, and we do not tune any other forms of regularisation (e.g., L1, dropout).

\paragraph{Polarised regime and posterior collapse}
The polarised regime, also known as selective posterior collapse, is the ability of VAEs to ``shut down''
superfluous dimensions of their sampled latent representations while providing a high precision on the remaining
ones~\citep{Rolinek2019,Dai2020}.
The existence of the polarised regime is a necessary condition for
the VAEs to provide a good reconstruction~\citep{Dai2018,Dai2020}.~However, when
the weight on the regularisation term of the ELBO given in~\Eqref{eq:elbo} becomes too large, the
representations collapse to the prior~\citep{Lucas2019,Dai2020}.
Recently, \citet{Bonheme2021} have also shown that the passive variables, which are ``shut down'' during training, are very
different in mean and sampled representations (see \Appref{sec:app-vae}). This indicates that representational similarity could be a valuable tool in the study of posterior collapse.

\subsection{Representational similarity metrics}\label{subsec:bg-similarity}
Representational similarity metrics aim to compare the geometric similarity between two representations.
In the context of deep learning, these representations correspond to $\R^{n \times p}$ matrices of activations, where
$n$ is the number of data examples and $p$ the number of neurons in a layer.~Such metrics can provide various information on
deep neural networks (e.g., the training dynamics of neural networks, common and specialised layers between models).

\paragraph{Centred Kernel Alignment} Centred Kernel Alignment (CKA)~\citep{Cortes2012,Cristianini2002} is
a normalised version of the Hillbert-Schmit Independence Criterion (HSIC)~\citep{Gretton2005}. As its name suggests,
it measures the alignment between the $n \times n$ kernel matrices of two representations, and works well with linear kernels~\citep{Kornblith2019} for representational similarity of centred layer activations.
We thus focus on the linear CKA, also known as RV-coefficient~\citep{Robert1976}.
Given the centered layer activations $\mX \in \R^{n \times m}$ and $\mY \in \R^{n \times p}$ taken over $n$ data examples,
linear CKA is defined as:
\begin{equation*}
    CKA(\mX, \mY) = \frac{\norm{\mY^T\mX}_F^2}{\norm{\mX^T\mX}_F\norm{\mY^T\mY}_F},
\end{equation*}
where $\norm{\cdot}_F$ is the Frobenius norm.
For conciseness, we will refer to linear CKA as CKA in the rest of this paper.
Note that CKA takes values between 0 (not similar) and 1 ($\mX = \mY$).

\paragraph{Orthogonal Procrustes} The aim of orthogonal Procrustes~\citep{Schonemann1966} is to align a matrix $\mY$ to a matrix $\mX$ using orthogonal transformations $\mQ$
such that
\begin{equation}\label{eq:p-mini}
    \min_{\mQ} \norm{\mX - \mY\mQ}^2_F \quad\mathrm{s.t.}\quad \mQ^T\mQ = \mI.
\end{equation}
%\begin{mini}|s|
%{\mQ}{\norm{\mX - \mY\mQ}^2_F}{}{}
%\label{eq:p-mini}
%\addConstraint{\mQ^T\mQ = \mI}.
%\end{mini}
The Procrustes distance, $P_d$, is the difference remaining between $\mX$ and $\mY$ when $\mQ$ is optimal,
\begin{equation}\label{eq:pd}
    P_d(\mX, \mY) = \norm{\mX}^2_F + \norm{\mY}^2_F - 2\norm{\mY^T\mX}_*,
\end{equation}
where $\norm{\cdot}_*$ is the nuclear norm (see \citet[pp.~327-328]{Golub2013} for the full derivation from \Eqref{eq:p-mini} to \Eqref{eq:pd}).~To easily compare the results of \Eqref{eq:pd} with CKA,
we first bound its results between 0 and 2 using normalised $\dot{\mX}$ and $\dot{\mY}$, as detailed in \Appref{sec:xp-setup}.
Then, we transform the result to a similarity metric ranging from 0 (not similar) to 1 ($\mX = \mY$),
\begin{equation}\label{eq:ps}
    P_s(\mX, \mY) = 1 - \frac{1}{2} \left(\norm{\dot{\mX}}^2_F + \norm{\dot{\mY}}^2_F - 2\norm{\dot{\mY}^T\dot{\mX}}_*\right).
\end{equation}
We will refer to \Eqref{eq:ps} as Procrustes similarity in the following sections.

\subsection{Limitations of CKA and Procrustes similarities}\label{subsec:bg-limitations}
While CKA and Procrustes lead to accurate results in practice, they suffer from some limitations that need to be taken into account in
our study.~Before we discuss these limitations, we should clarify that, in the rest of this paper, $sim(\cdot, \cdot)$ represents a similarity metric in general, while
$CKA(\cdot, \cdot)$ and $P_s(\cdot, \cdot)$ specifically refer to CKA and Procrustes similarities.

\paragraph{Sensitivity to architectures}
\citet{Maheswaranathan2019} have shown that similarity metrics comparing the geometry of representations
were overly sensitive to differences in neural architectures.~As CKA and Procrustes belong to this metrics family,
we can expect them to underestimate the similarity between activations coming from layers of different type (e.g., convolutional
and deconvolutional).

\begin{figure}[ht!]
    \centering
    \subcaptionbox{$sim(\mA,\mB)$\label{fig:sim}}{
        \includegraphics[width=0.3\linewidth]{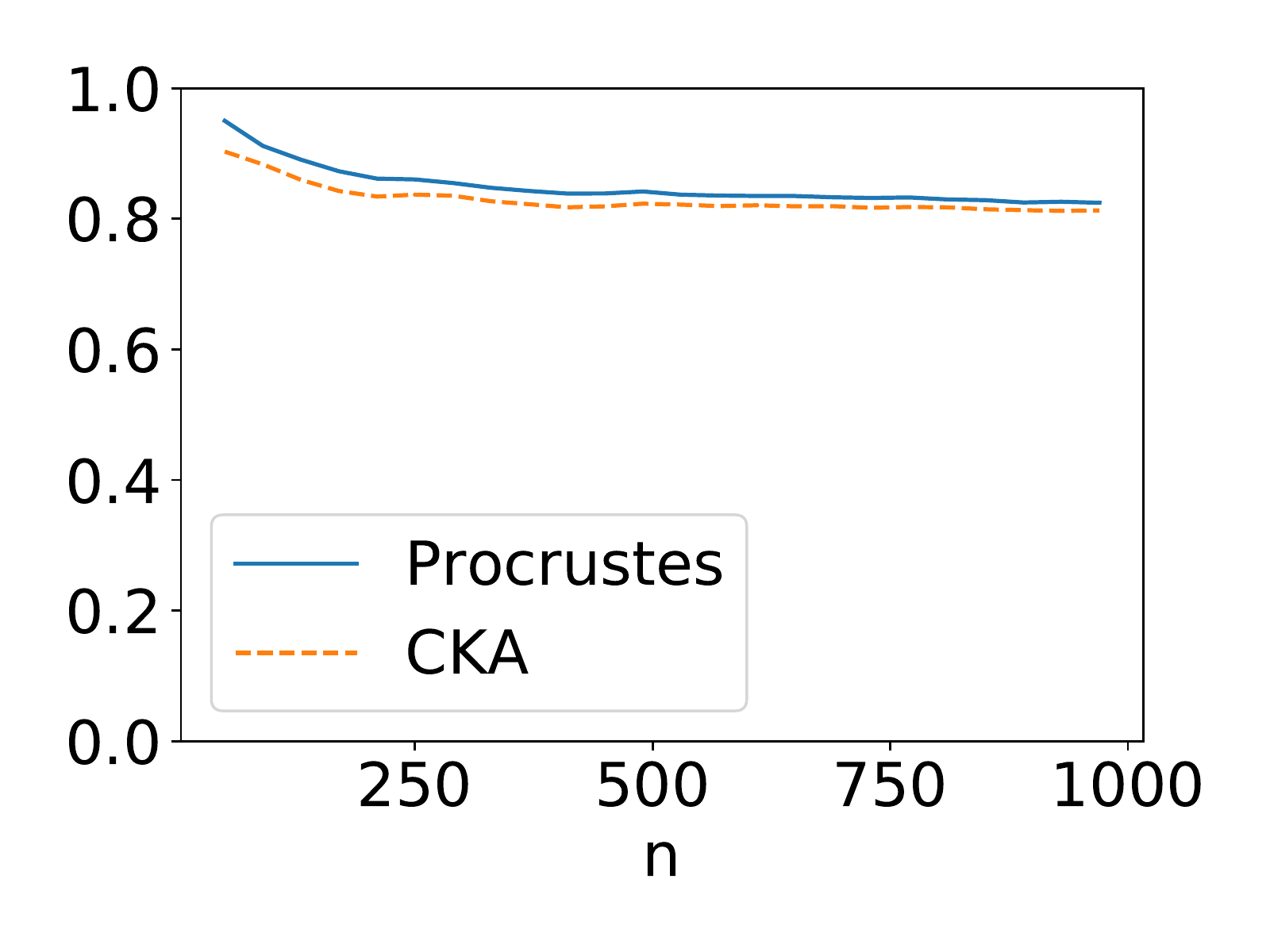}
    }%
    \hfill
    \subcaptionbox{$sim(\mA,\mC)$\label{fig:mid-sim}}{
        \includegraphics[width=0.3\linewidth]{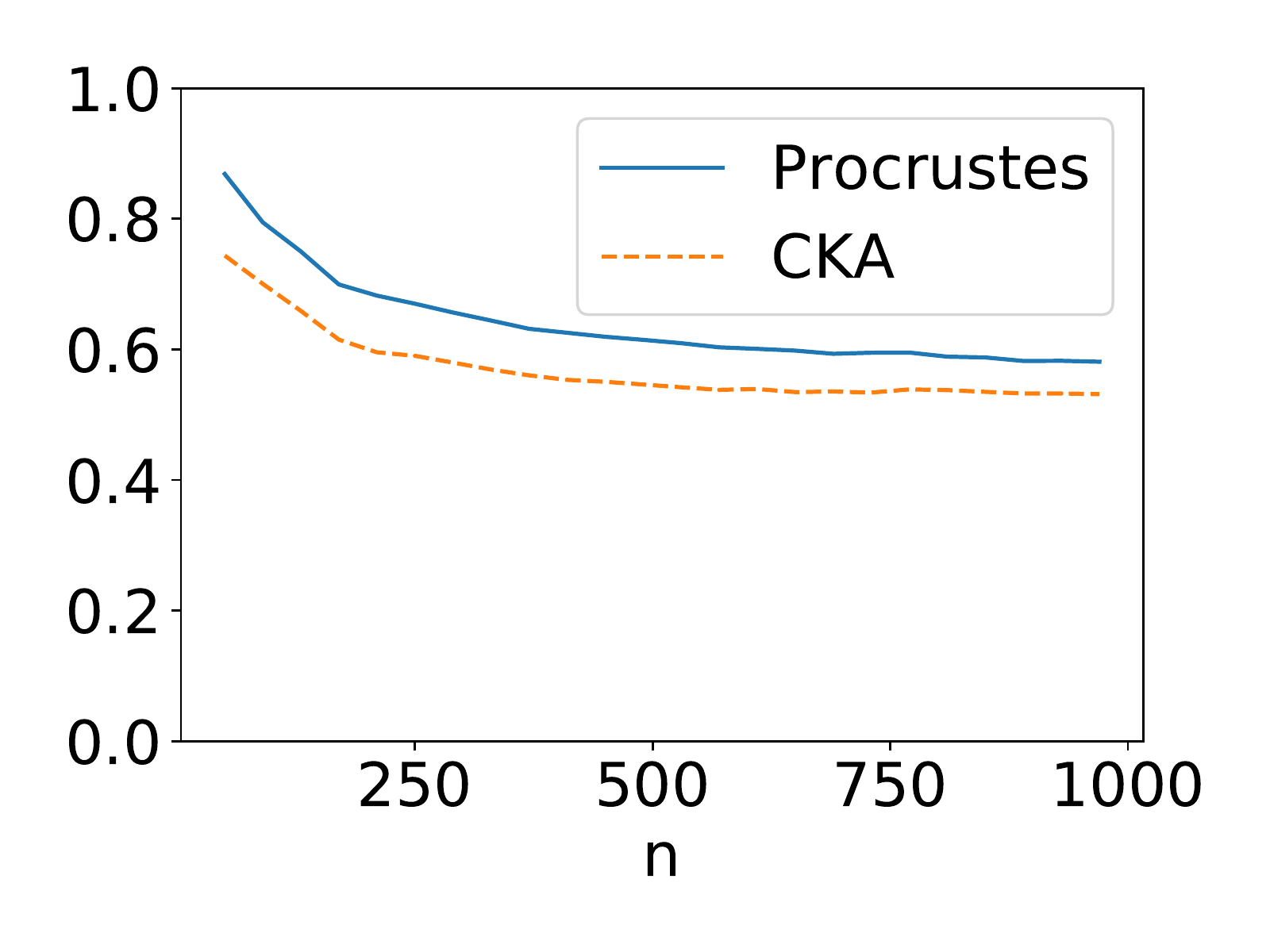}
    }%
    \hfill
    \subcaptionbox{$sim(\mA,\mD)$\label{fig:diff}}{
        \includegraphics[width=0.3\linewidth]{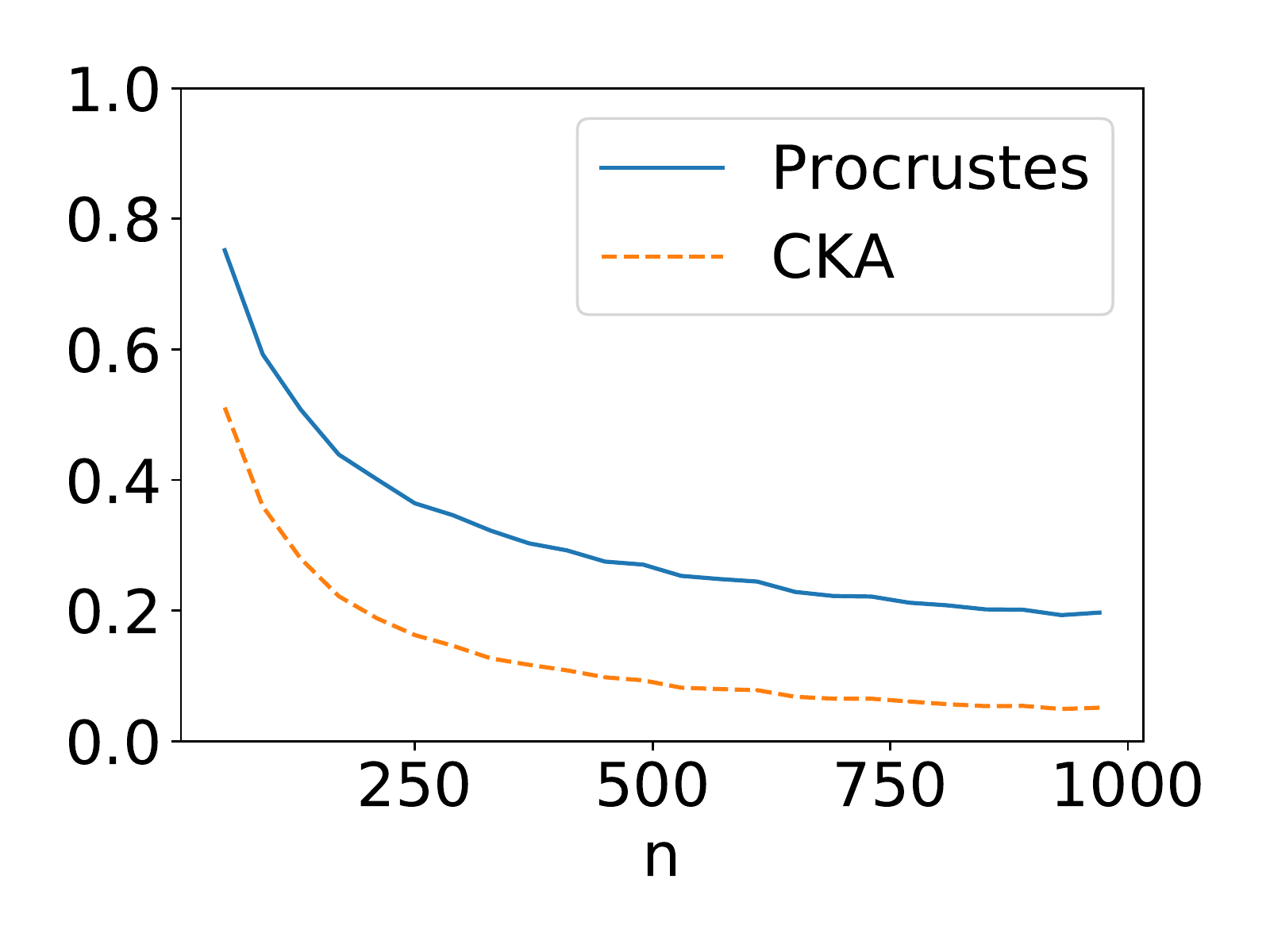}
    }%
   \caption{We compute CKA and Procrustes similarity scores with an increasing number of data examples $n$, and different
    similarity strength: $\mB$ retains 80\% of $\mA$'s features, $\mC$ 50\%, and $\mD$ 0\%.~Both metrics agree in
    (a) and (b), but Procrustes overestimates similarity in (c).}
    \label{fig:cka-procrustes}
\end{figure}

\paragraph{Procrustes is sensitive to the number of data examples}
As we may have representations with high dimensional features (e.g., activations of convolutional layers),
we checked the impact of the number of data examples on CKA and Procrustes.
To do so, we created four increasingly different matrices
$\mA, \mB, \mC$, and $\mD$ with 50 features each: $\mB$ retains 80\% of $\mA$'s features, $\mC$ 50\%, and $\mD$ 0\%.
~We then computed the similarity scores given by CKA and Procrustes while varying the number
of data examples.~As shown in \Twofigref{fig:sim}{fig:mid-sim}, both metrics agree for $sim(\mA,\mB)$ and $sim(\mA,\mC)$,
giving scores that are close to the fraction of common features between the two matrices.
However, we can see in~\Figref{fig:diff} that Procrustes highly overestimates
$sim(\mA, \mD)$ while CKA scores rapidly drop.

\paragraph{CKA ignores small changes in representations}
When considering a sufficient number of data examples for both Procrustes and CKA, if two representations do not have
dramatic differences (i.e., their 10\% largest principal components are the same),
CKA may overestimate similarity, while Procrustes remains stable, as observed by~\citet{Ding2021}.

\begin{figure}[ht!]
    \centering
    \subcaptionbox{Representation $\mA$\label{fig:aligned}}{
        \includegraphics[width=0.4\linewidth]{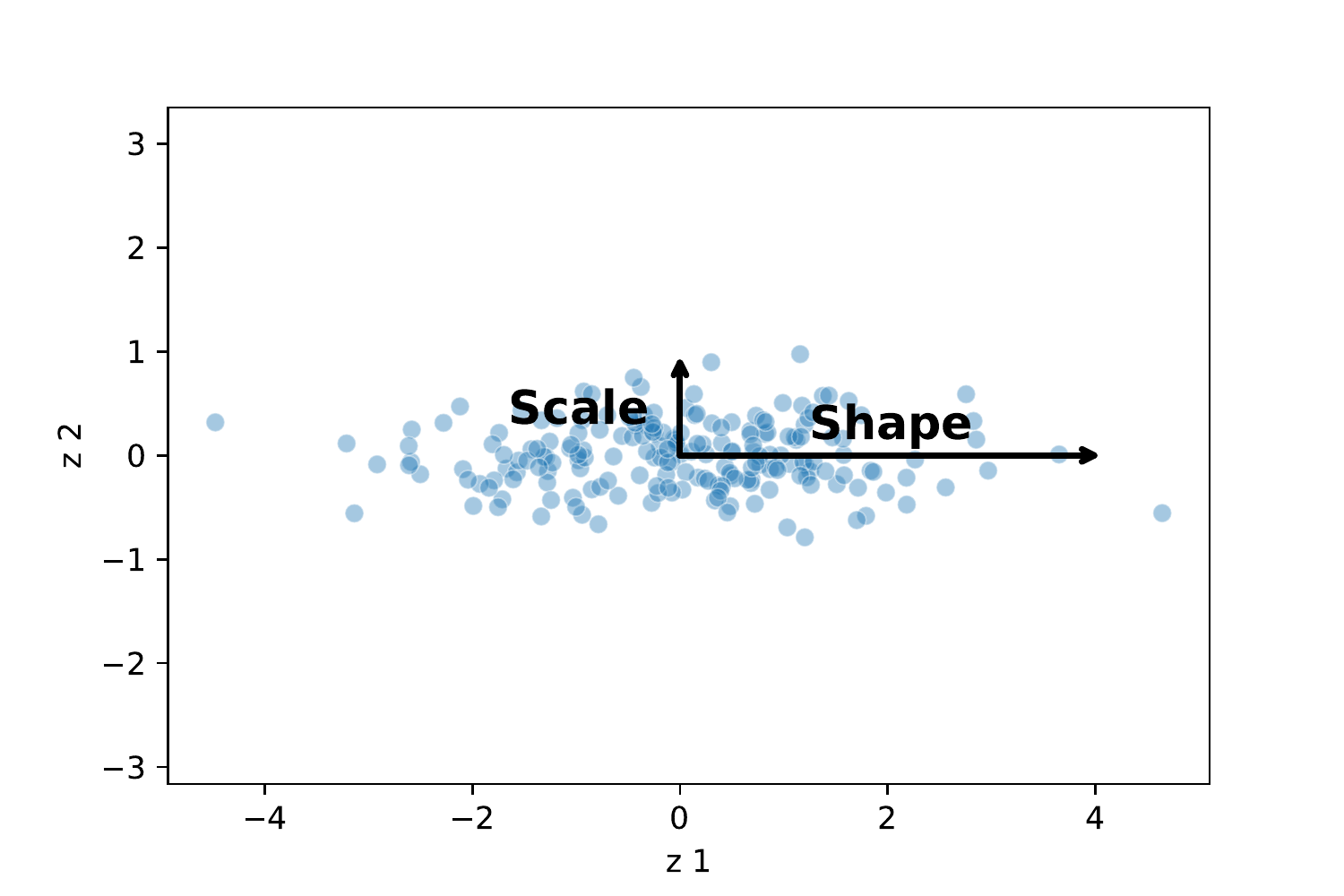}
    }%
    \hfill
    \subcaptionbox{Representation $\mB$\label{fig:rotated}}{
        \includegraphics[width=0.4\linewidth]{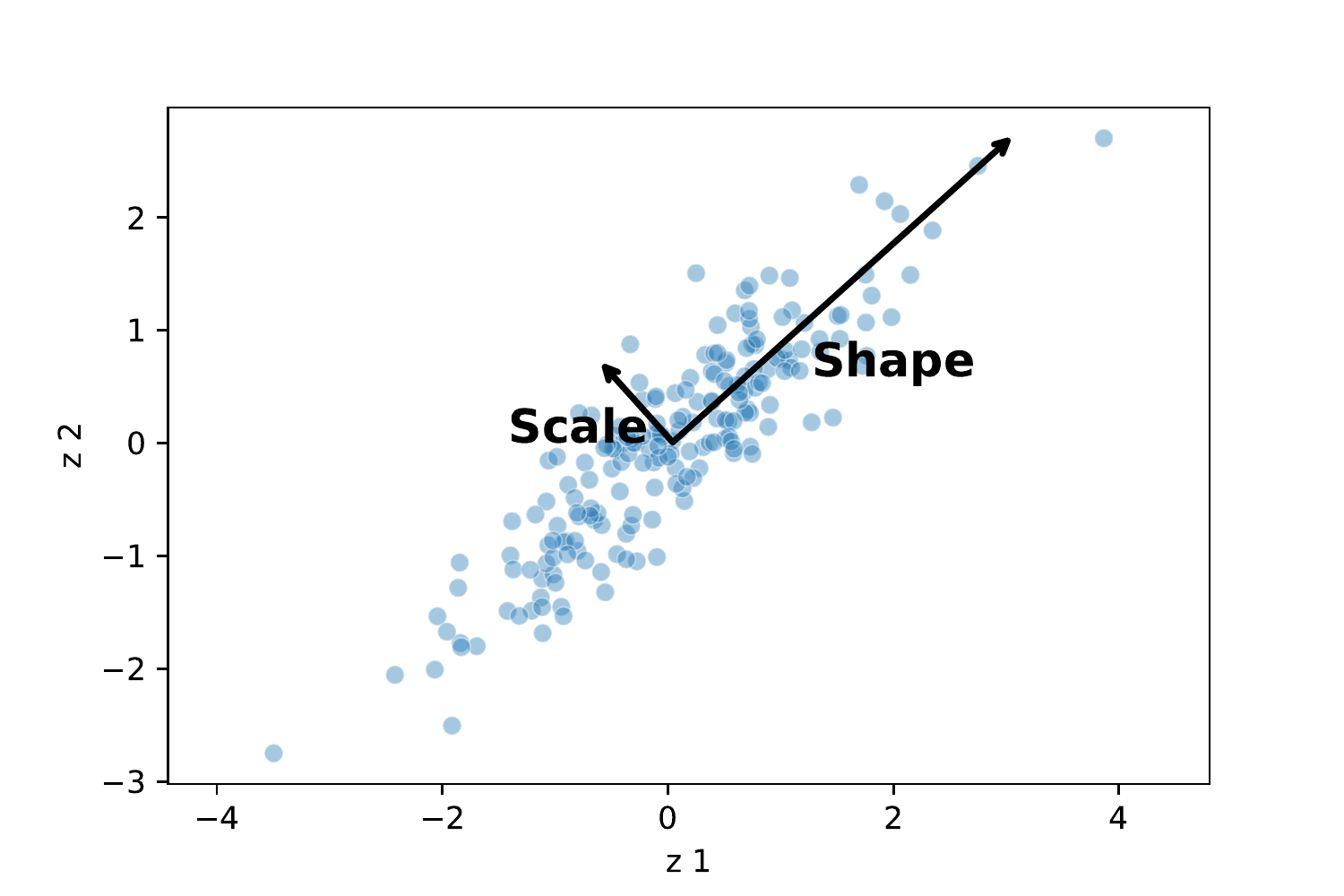}
    }%
    \caption{(a) shows a disentangled representation $\mA$ where the latent dimensions $\vz_1$ and $\vz_2$ are aligned
    with some ground truth factors shape and scale.~(b) shows a representation $\mB$ which
    corresponds to an orthogonal transformation of $\mA$. It is entangled as $\vz_1$ and $\vz_2$ are not aligned
    with the ground truth factors anymore, but $sim(\mA, \mB) = 1$ because Procrustes and CKA are invariant under orthogonal transformations.}
    \label{fig:disentanglement}
\end{figure}

\paragraph{Similarity and disentanglement} It is important to keep in mind that CKA and Procrustes similarities are
invariant to orthogonal transformation.~Thus, they will consider a disentangled representation similar to a rotated
(and possibly entangled) representation, as illustrated in~\Figref{fig:disentanglement}.
Note that CKA and Procrustes are also invariant to isotropic scaling, though this
does not affect disentanglement.

\paragraph{Ensuring accurate analysis}~Given the limitations previously mentioned, we take three remedial actions
to guarantee that our analysis is as accurate as possible.
Firstly, as both metrics will likely underestimate the similarity between different layer types,
we will only discuss the variation of similarity when analysing such cases.
For example, we will not compare $sim(\mA, \mB)$ and $sim(\mA, \mC)$ if $\mA$ and $\mB$ are convolutional layers but
$\mC$ is deconvolutional.~We will nevertheless analyse the changes of $sim(\mA, \mC)$ at different steps of training.
Secondly, when both metrics disagree, we know that one of them is likely overestimating the similarity: Procrustes if
the number of data examples is not sufficient, CKA if the difference between the two representations is not large enough.
~Thus, we will always use the smallest of the two results for our interpretations.
Lastly, when we say that representations are similar, we mean that they are similar up to orthogonal rotations and
isotropic scaling.~Hence, two similar representations may not be equally disentangled.

    \section{Experimental setup}\label{sec:experiment}
The goal of this experiment is to study the training dynamics of VAEs and the impact of initialisation, learning objectives, and
regularisation on the representations learned by each layer.
To do so, we measure the representational similarity of:
\begin{enumerate}
    \itemsep 0em
    %\item Two models with identical learning objective and regularisation strength but different initialisations at the same epoch.
    \item One model at different epochs.
    \item Two models with the same learning objective and different regularisation strengths at the same epoch.
    \item Two models with different learning objectives and equivalent regularisation strength at the same epoch.
\end{enumerate}

\paragraph{Learning objectives} We will focus on learning objectives whose goal is to produce disentangled
representations.~Specifically, we use $\beta$-VAE~\citep{Higgins2017}, $\beta$-TC VAE~\citep{Chen2018}, Annealed VAE~\citep{Burgess2018}, and DIP-VAE II~\citep{Kumar2018}.
A description of these methods can be found in~\Appref{sec:app-disentanglement}.
~To fairly provide complementary insights into previous observations of such models~\citep{Locatello2019a, Bonheme2021},
we will follow the experimental design of~\citet{Locatello2019a} regarding the architecture, learning objectives, and regularisation used.
Moreover, \texttt{disentanglement lib}\footnote{\url{https://github.com/google-research/disentanglement_lib}} will be used as a codebase for our experiment.
The complete details are available in \Appref{sec:xp-setup}.

\paragraph{Datasets} We use three datasets which, based on the results of~\citet{Locatello2019a}, are increasingly difficult for VAEs in terms of reconstruction loss:
dSprites\footnote{Licensed under an Apache 2.0 licence.}~\citep{Higgins2017}, cars3D~\citep{Reed2015}, and smallNorb~\citep{LeCun2004}.

\paragraph{Training process} We trained five models with different initialisations for 300,000 steps for each (learning objective, regularisation strength, dataset)
triplet, and saved intermediate models to compare the similarity within individual models at different epochs.
\Appref{sec:app-epochs} explains our epoch selection methodology.

\paragraph{Similarity measurement} Given the computational complexity detailed below, for every dataset, we sampled 5,000
data examples, and we used them to compute all the similarity measurements.
We compute the similarity scores between all pairs of layers of the different models following the different combinations outlined above.
We will refer to the similarity scores of a group of one or more layers with itself as \textit{self-similarity}.
~As Procrustes similarity takes significantly longer to compute compared to CKA (see below), we only used
it to validate CKA results, restricting its usage to one dataset: cars3D.
We obtained similar results for the two metrics on cars3D, thus we only reported CKA results in the main paper.
Procrustes results can be found in~\Appref{sec:procrustes}.

\paragraph{Computational considerations} Overall we trained 300 VAEs using 4 learning objectives, 5 different initialisations,
5 regularisation strengths, and 3 datasets, which took around 6,000 hours on an NVIDIA A100 GPU.
We then computed the CKA scores for the 15 layer activations (plus the input) of each model combinations
considered above at 5 different epochs, resulting in 470 million similarity scores and approximately 7,000 hours of computation
on an Intel Xeon Gold 6136 CPU.~As Procrustes is slowed down by the computation of the nuclear norm for high dimensional activations, the same number of similarity
scores would have been prohibitively long to compute, requiring 30,000 hours on an NVIDIA A100 GPU.~We thus only computed the Procrustes
similarity for one dataset, reducing the computation time to 10,000 hours.
Overall, based on the estimations of~\citet{Lacoste2019}, the computations done for this experiment amount to
2,200 Kg of \coo, which corresponds to the \coo~produced by one person over 5 months.~To mitigate the negative environmental impact of our work,
we released \ifanonymous all our metric scores at \url{https://t.ly/0GLe3}\else all our trained models and metric scores at \url{https://data.kent.ac.uk/428/}, and \url{https://data.kent.ac.uk/444/}, respectively\fi.
We hope that this will help to prevent unnecessary recomputation should others wish to reuse our results.

    \section{Results}\label{sec:results}

\subsection{How are representations learned as training progresses?}\label{subsec:res-training}
%We have seen in \Secref{subsec:res-general} that models have different representations at different epochs.
%However, in \Figref{fig:tsne}, we only computed the similarity between models trained for the same amount of epochs,
%which does not provide any insights into how one model changes over the training period.
In this section, we will analyse the learning dynamics of VAEs to answer objective (i) of \Secref{sec:experiment}.
To monitor this, we will compare the representations learned by the first and last recorded snapshot of VAEs.
Note that our choice of snapshots and snapshot frequency does not influence the results as verified
in~\Twoappref{sec:app-epochs}{sec:app-convergence}.

\begin{figure}[ht!]
    \centering
    \subcaptionbox{Trained on cars3D\label{fig:heatmap-1}}{
        \includegraphics[width=0.33\textwidth]{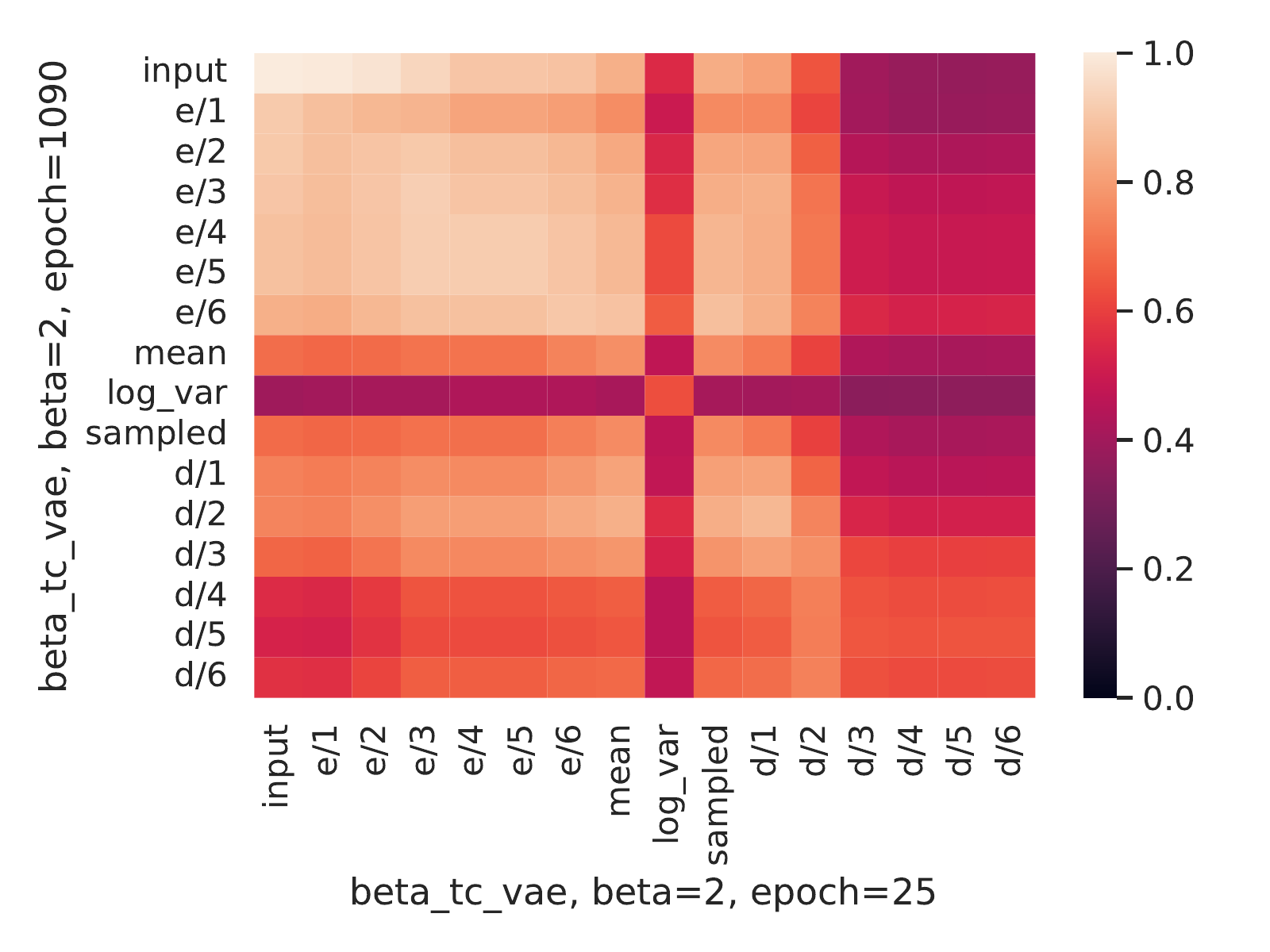}
    }%
    \subcaptionbox{Trained on dSprites\label{fig:heatmap-2}}{
        \includegraphics[width=0.33\textwidth]{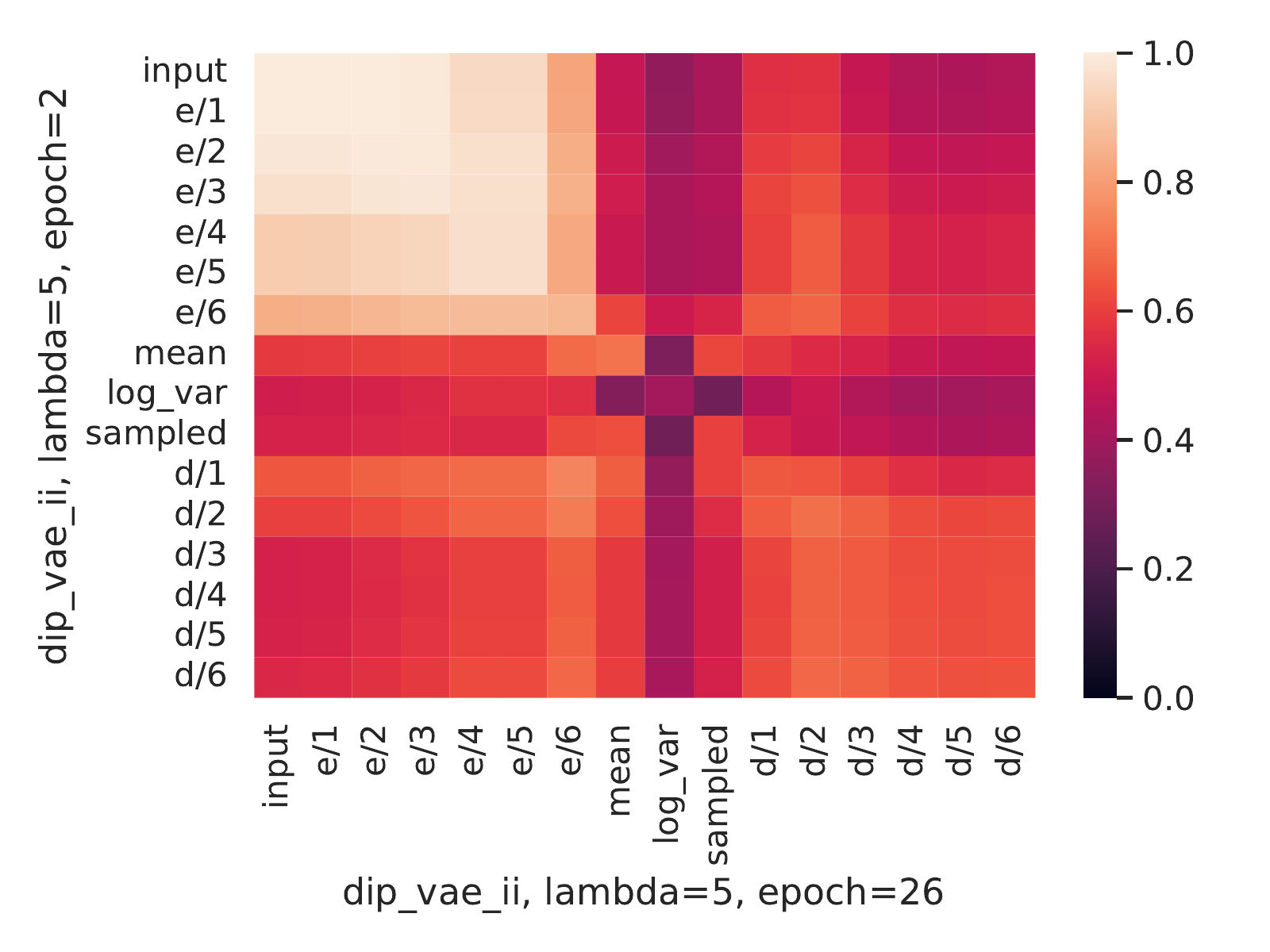}
    }%
    \subcaptionbox{Trained on smallNorb\label{fig:heatmap-3}}{
        \includegraphics[width=0.33\textwidth]{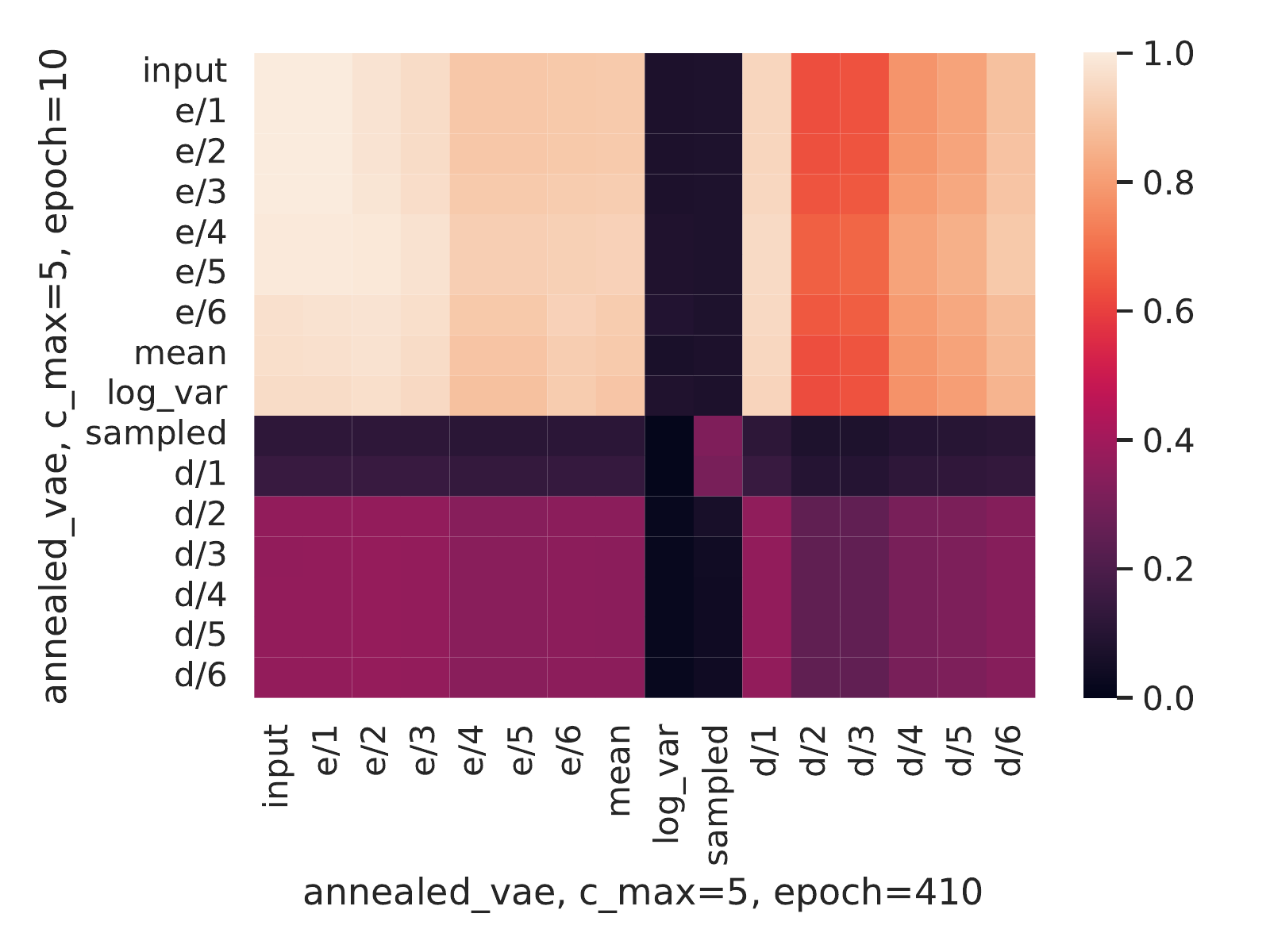}
    }%
    \caption{(a) shows the CKA similarity scores of activations at epochs 25 and 1090 of $\beta$-TC VAE trained on cars3D with $\beta=2$.
        (b) shows the CKA similarity scores of activations at epochs 2 and 26 of DIP-VAE II trained on dSprites with $\lambda=5$.
        (c) shows the CKA similarity scores of activations at epochs 10 and 410 of Annealed VAE trained on smallNorb with $c_{max}=5$.
        We can see that there is a high similarity between the representations learned by the encoder early in the training and after complete training
        (see the bright cells in the top-left quadrants in (a), (b), and (c)).
        However, the self-similarity between the decoder, mean and variance representations is lower,
        indicating some further changes in the representations after the first few epochs of training (see the dark cells
        in the bottom-right quadrant of parts (a), (b), and (c)). These results are averaged over 5 seeds.}
    \label{fig:training-sim}
\end{figure}

\paragraph{VAE learning is bottom-up} As shown in \Figref{fig:training-sim}, the encoder is learned first,
and the representations of its layers become similar to the input after a few epochs.~The decoder then progressively
learns representations that gradually become closer to the input. We observed a similar
trend on fully-connected architectures, as reported in~\Appref{sec:app-fc}.
~This result can explain why, when a decoder has access to
the input (as it is the case in some autoregressive VAEs), it ignores the latent representations~\citep{Bowman2016,Li2019a}.
Indeed, in this case the decoder is not constrained to wait for the encoder to converge before improving its reconstruction
 since it has direct access to the input sequences.
~In this paper, we used standard VAEs, where the decoder is not autoregressive and must rely on the latent representations as the
sole source of information about the input.~It thus needs to wait for the encoder to converge before being able to
converge itself, effectively preventing posterior collapse when the regularisation is not too strong.

\paragraph{Implications} The concurrent learning of encoder and decoder representations can be problematic.
Indeed, if the encoder learns poor representations, the decoder will not be able to provide an accurate reconstruction.
As a result, the encoder may struggle to provide better representations and this may lead to non-optimal results.
~We believe methods that facilitate the incremental learning of the encoder and decoder (e.g., by slowly increasing the
complexity of the data, as in Progressive GANs~\citep{Karras2018}) as well as methods where the encoder is explicitly
learned first~\citep{He2019} are promising ways to mitigate this issue. Moreover, the early convergence of the encoder
can explain the success of ``warm-up'' methods such as annealing~\citep{Bowman2016} or aggressive inference~\citep{He2019}
in the case where the encoder (quickly) converges to a bad local minima resulting in posterior collapse.

\subsection{What is the influence of the hyperparameters on the learned representations?}\label{subsec:res-hyperparameters}
%We have seen in \Secref{subsec:res-general} that models with the same learning objective but different
%regularisation strength have different representations. However, as we compared entire similarity matrices in~\Figref{fig:tsne-reg},
%we cannot see which specific layers are different.
So far, we have compared different snapshots of the same model.~In this section, we will perform a fine-grained analysis
of the impact of hyperparameters to answer objective (ii) of \Secref{sec:experiment}.

\begin{figure}[ht!]
    \centering
    \subcaptionbox{$\beta$-VAE\label{fig:reg-1}}{
        \includegraphics[width=0.35\textwidth]{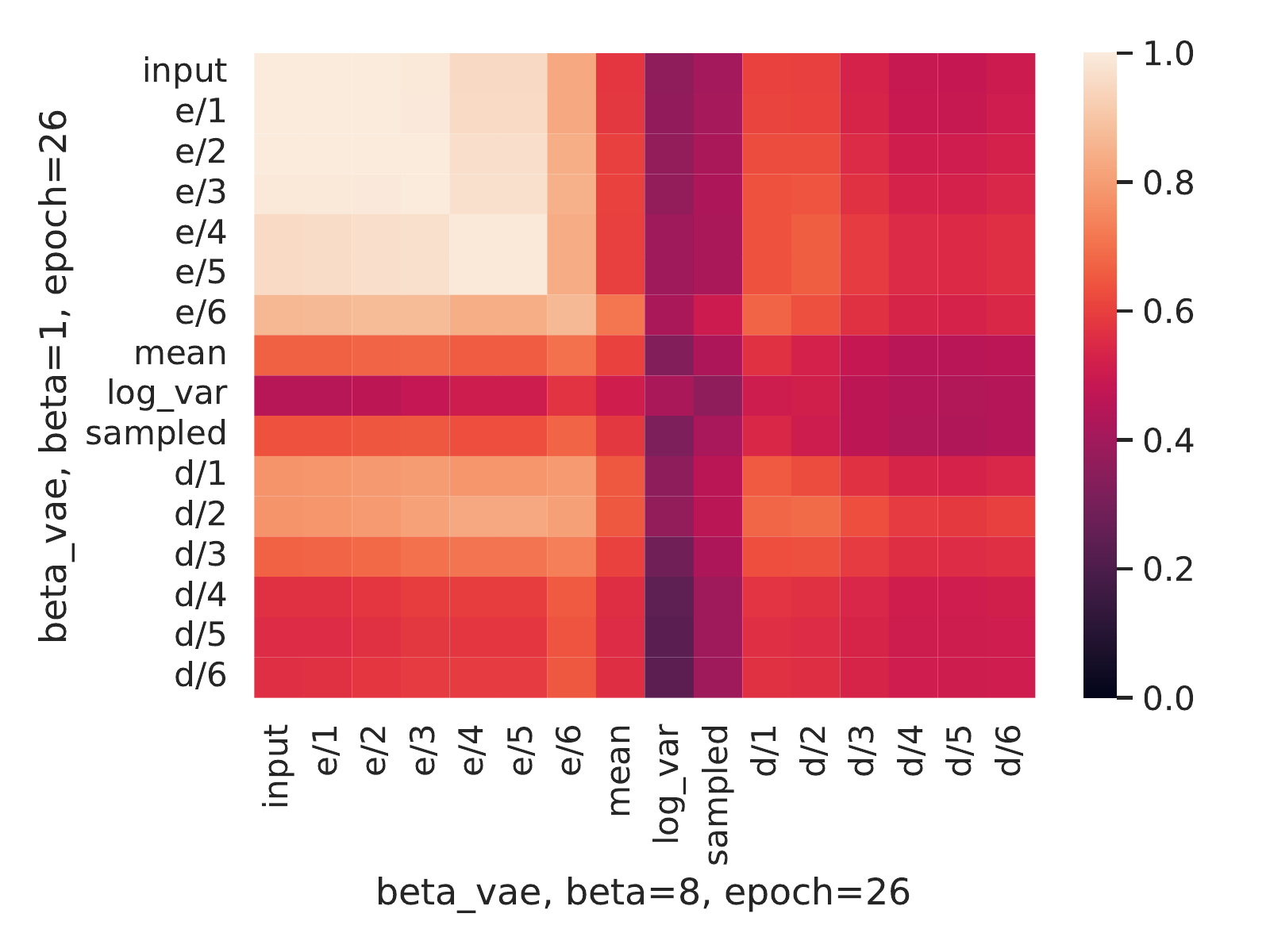}
    }%
    \hfill
    \subcaptionbox{DIP-VAE II\label{fig:reg-2}}{
        \includegraphics[width=0.35\textwidth]{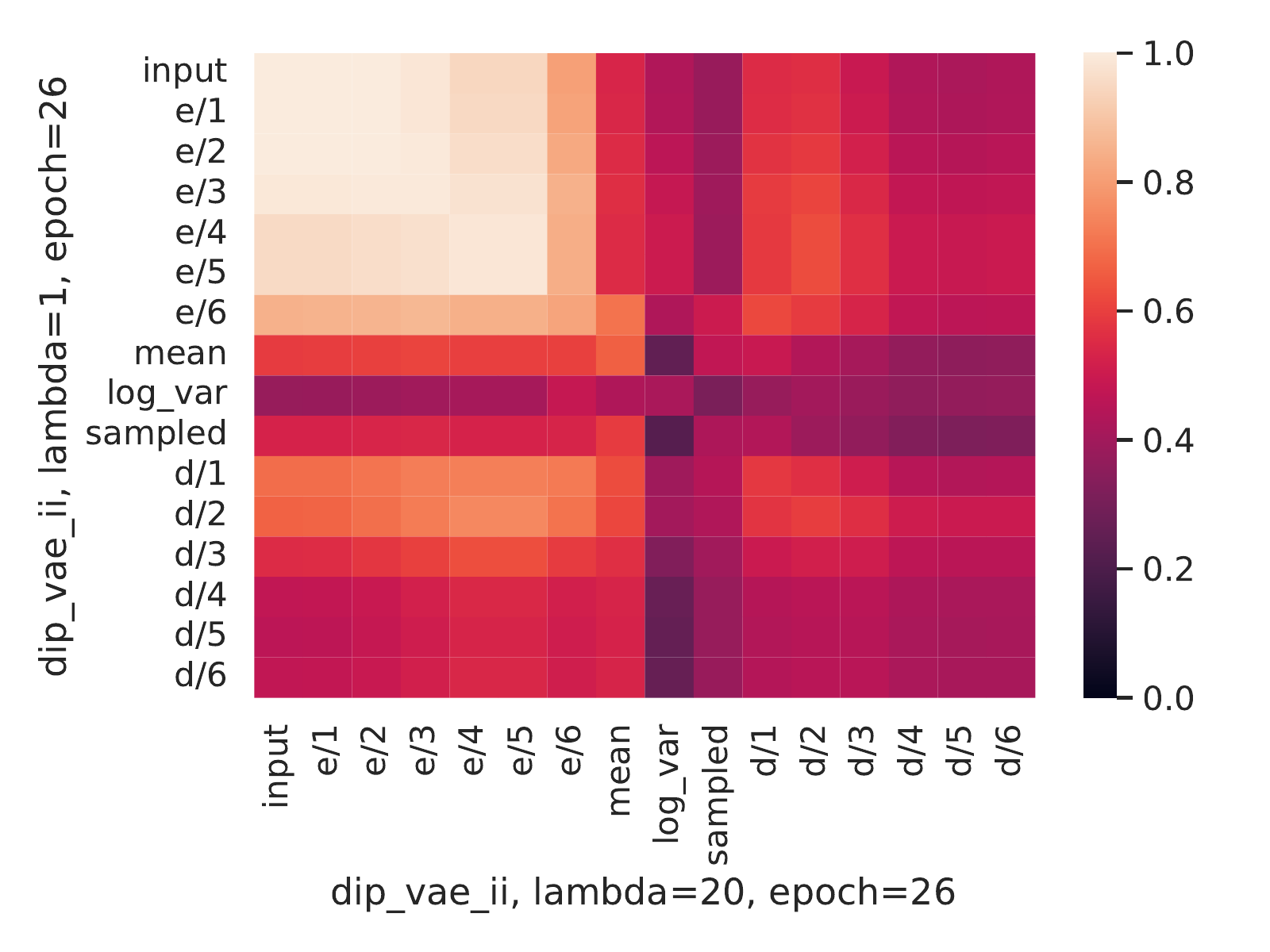}
    }%
    \caption{(a) shows the CKA similarity scores between the activations of two $\beta$-VAEs trained on dSprites with $\beta=1$, and $\beta=8$.
        (b) shows the CKA similarity scores between the activations of two DIP-VAE II trained on dSprites with $\lambda=1$, and $\lambda=20$.
        For (a) and (b), the activations are taken after complete training, and all the results are averaged over 5 seeds.
        In both figures, we can see that the self-similarity of decoder representations is low between models trained with low and high regularisation
        (dark cells in the bottom-right quadrants) while the encoder representations stay very similar (bright cells in the top-left quadrants),
        except for the mean, variance and sampled representations.}
    \label{fig:reg-impact}
\end{figure}

\begin{figure}[ht!]
    \centering
    \subcaptionbox{CKA (mean, sampled)\label{fig:mean-sampled}}{
        \includegraphics[width=0.35\textwidth]{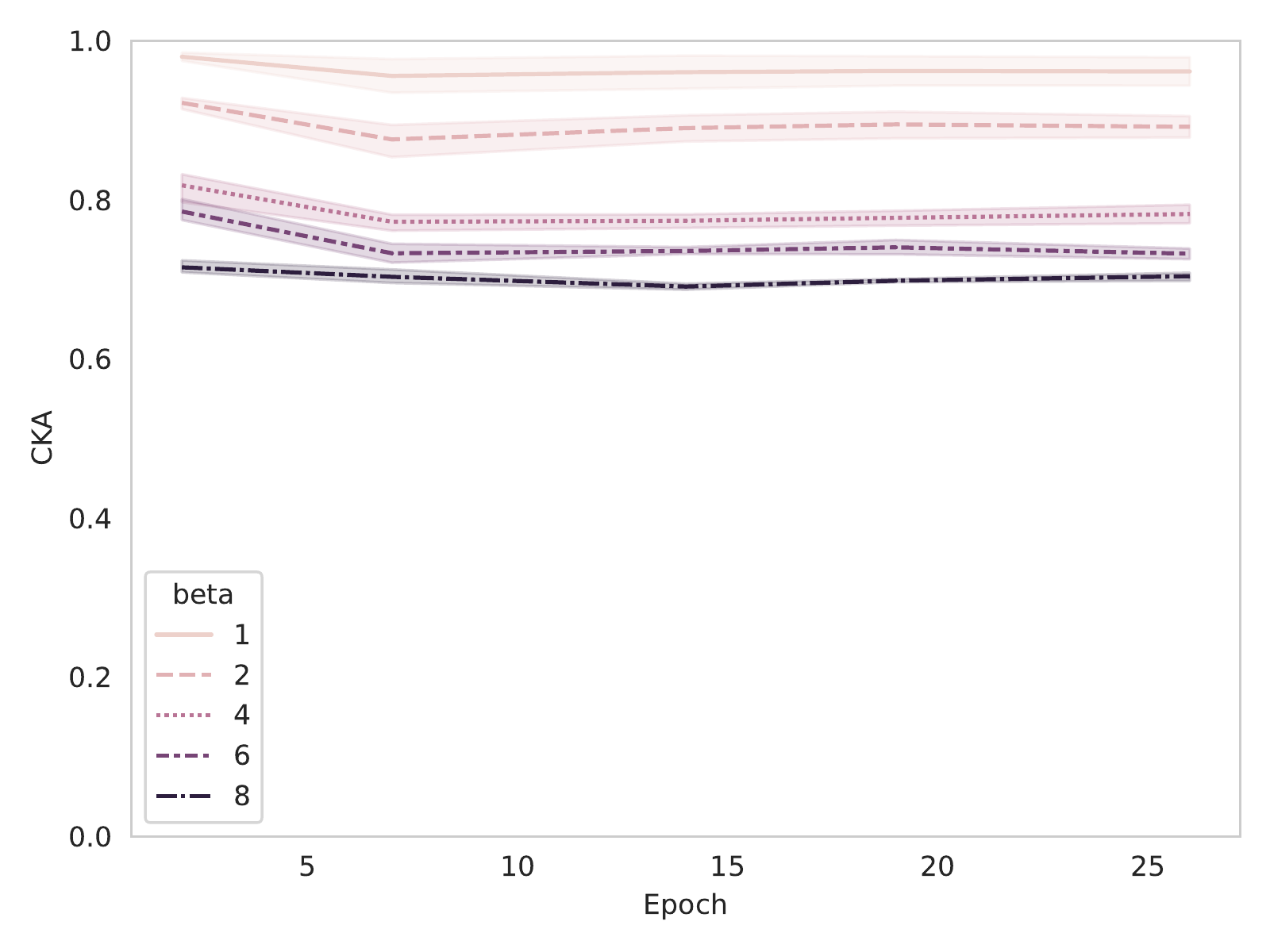}
    }%
    \hfill
    \subcaptionbox{CKA (input, sampled)\label{fig:input-sampled}}{
        \includegraphics[width=0.35\textwidth]{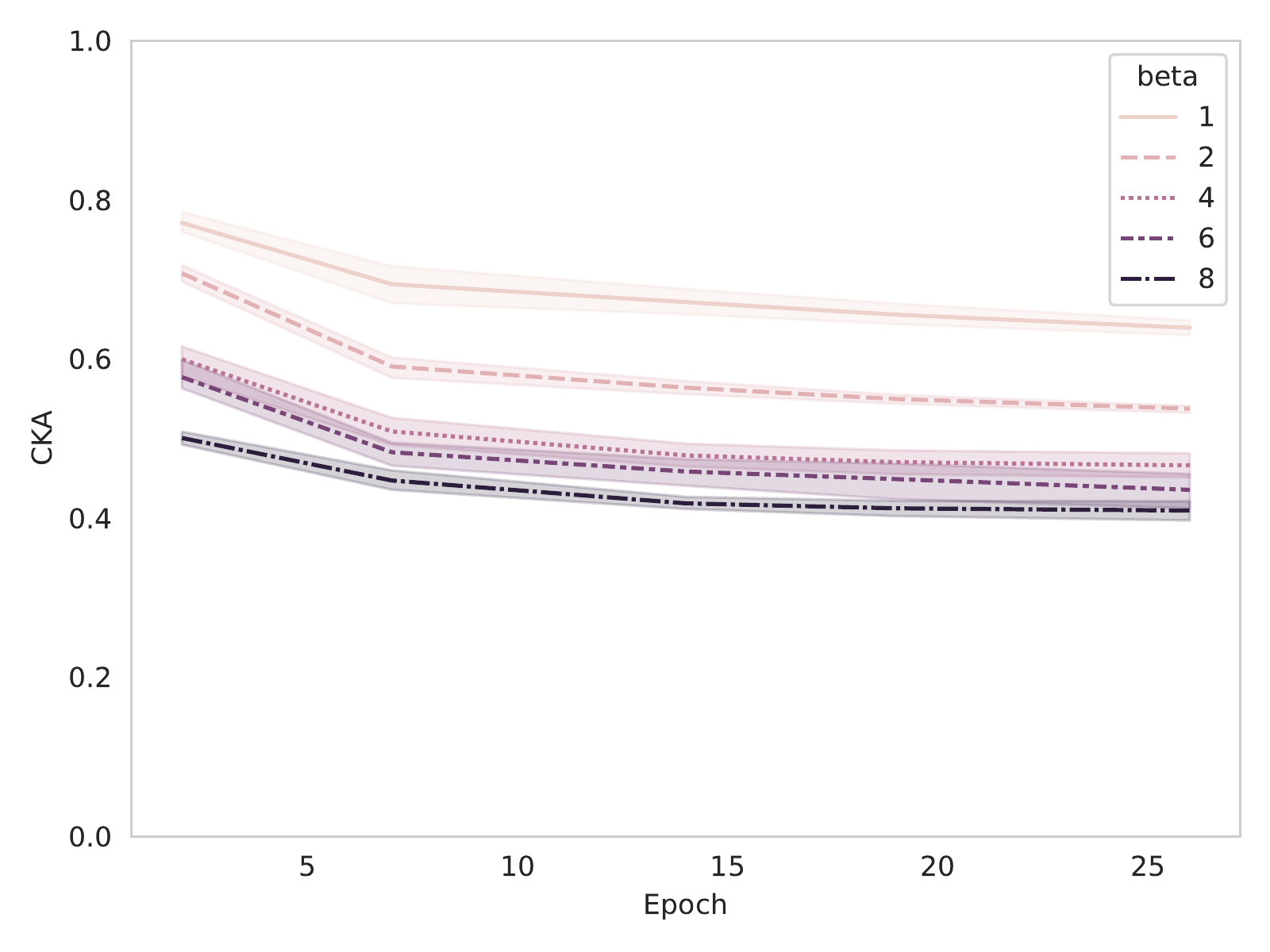}
    }%
    \caption{(a) and (b) show the CKA scores of $\beta$-VAEs trained on dSprites with $\beta$ from 1 to 8.
        (a) is the CKA between mean and sampled representations while (b) is between inputs and sampled representations.
        Both figures are consistent with posterior collapse caused by over-regularisation~\citep{Rolinek2019,Bonheme2021}
        where the mean and sampled representations present a growing number of passive variables which,
        in the case of sampled representations, leads to high dissimilarity with the input in (b).}
    \label{fig:posterior-collapse}
\end{figure}

\begin{figure}[ht!]
    \centering
    \subcaptionbox{$\beta=1$\label{fig:enc-1}}{
        \includegraphics[width=0.33\textwidth]{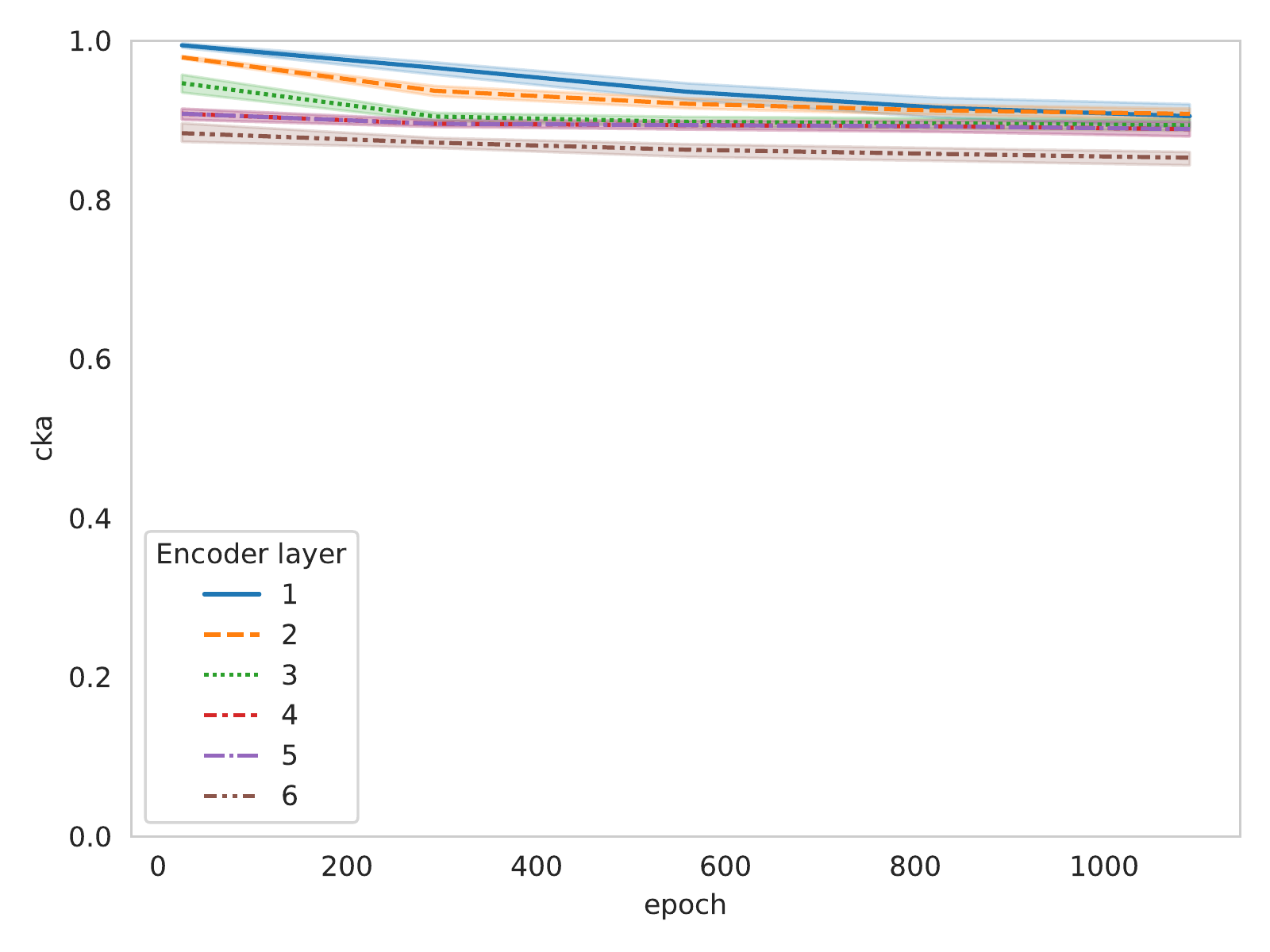}
    }%
    \hfill
    \subcaptionbox{$\beta=8$\label{fig:enc-2}}{
        \includegraphics[width=0.33\textwidth]{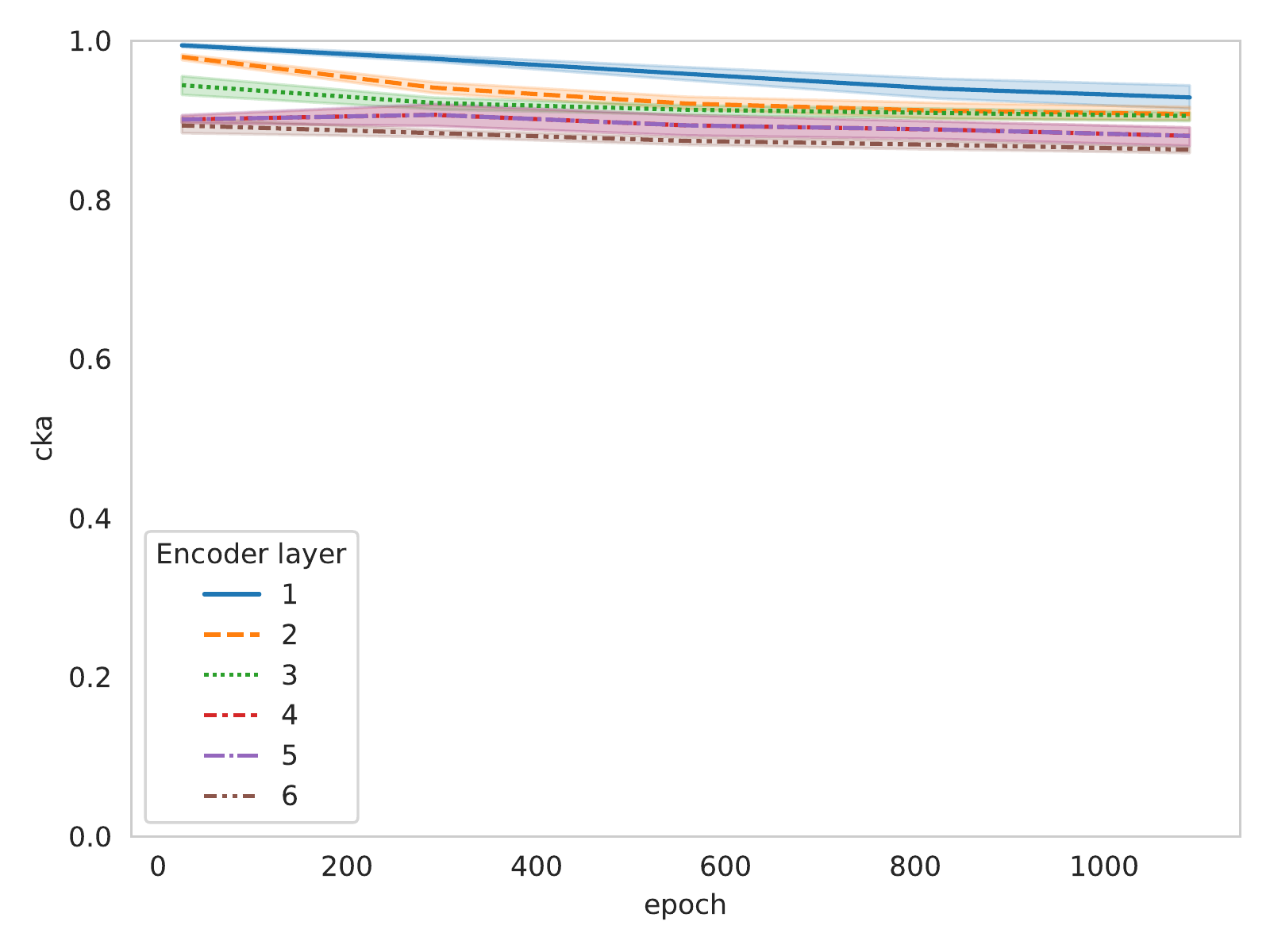}
    }%
     \caption{(a) shows the CKA scores between the inputs and the activations of the first 6 layers of the encoder of
        $\beta$-VAEs trained on cars3D with $\beta=1$.~(b) shows the scores between the same representations with $\beta=8$.
        We can see that we have low variance across seeds (i.e., the shaded area corresponding to the variance interval around each line is small) and the encoder layers have an equivalent similarity
        with the input, regardless of the regularisation strength.}
    \label{fig:reg-encoder}
\end{figure}

\paragraph{Impact of regularisation}~As discussed in~\citep{Bonheme2021}, passive variables have different values in the
mean and sampled representations.~This phenomenon is increasingly visible with higher regularisation, as more
variables must become passive in order to lower the KL divergence~\citep{Rolinek2019,Dai2020}.~However, little is known about how this
behaviour impacts the other layers of the encoder and decoder.~As shown in~\Figref{fig:reg-impact},
the decoder representations change more than the encoder representations with the increased regularisation strength.
~This can be explained by~\Figref{fig:posterior-collapse} where the sampled representations drift away
from the input and mean representations, which is consistent with posterior collapse~\citep{Rolinek2019,Dai2018,Dai2020}.
~This is further confirmed by the fact that posterior collapse was already reported in~\citep{Bonheme2021} for these
configurations.~Thus, our results indicate that CKA, which is quick to compute, could be a good tool to monitor
the polarised regime and posterior collapse --- its pathological counterpart --- and differentiate well the two behaviours (see~\Appref{sec:app-latents}).
Interestingly, apart from the mean and variance, the representations learned by the encoder in~\Figref{fig:reg-impact}
stay very similar.~We can further see in~\Figref{fig:reg-encoder} that this is consistent across seeds and very
different regularisation strengths, and can also be observed with fully-connected architectures in~\Appref{sec:app-fc}.

\paragraph{Implications}~The polarised regime~\citep{Rolinek2019,Zietlow2021,Dai2020} and posterior collapse
~\citep{Dai2018,Lucas2019,Lucas2019b,Dai2020} do not seem to affect the representations learned
by the encoder before the mean and variance layers (see \Figref{fig:reg-encoder}).
Intuitively, this would imply that the encoder learns similar representations, regardless of the regularisation
strength, and then ``fine-tunes'' them in its mean and variance layers.

\subsection{What is the influence of the learning objectives on the learned representations?}\label{subsec:res-objective}
We have seen in \Secref{subsec:res-hyperparameters} that models with the same learning objective (e.g., $\beta$-VAE, DIP-VAE II or any other learning objective defined in~\Appref{sec:app-disentanglement}) but different
regularisation were generally learning similar representations in the encoder except for the mean and variance
layers.
Starting from these layers and moving toward the decoder, the representational similarity decreases as the regularisation strength increases.
Indeed, passive variables have different mean and variance representations than active variables~\citep{Rolinek2019,Bonheme2021} and, as their number
grows, so does the dissimilarity with the representations obtained at lower regularisation, which have fewer passive variables.
Now we can wonder whether this pattern is also occuring in the context of objective (iii) of \Secref{sec:experiment}.
For an equivalent regularisation strength but different learning objectives, do the models still learn similar
representations in the layers of the encoder that are before the mean and variance layers?

\begin{figure}[ht!]
    \centering
    \subcaptionbox{Trained on cars3D\label{fig:methods-cars}}{
        \includegraphics[width=0.33\textwidth]{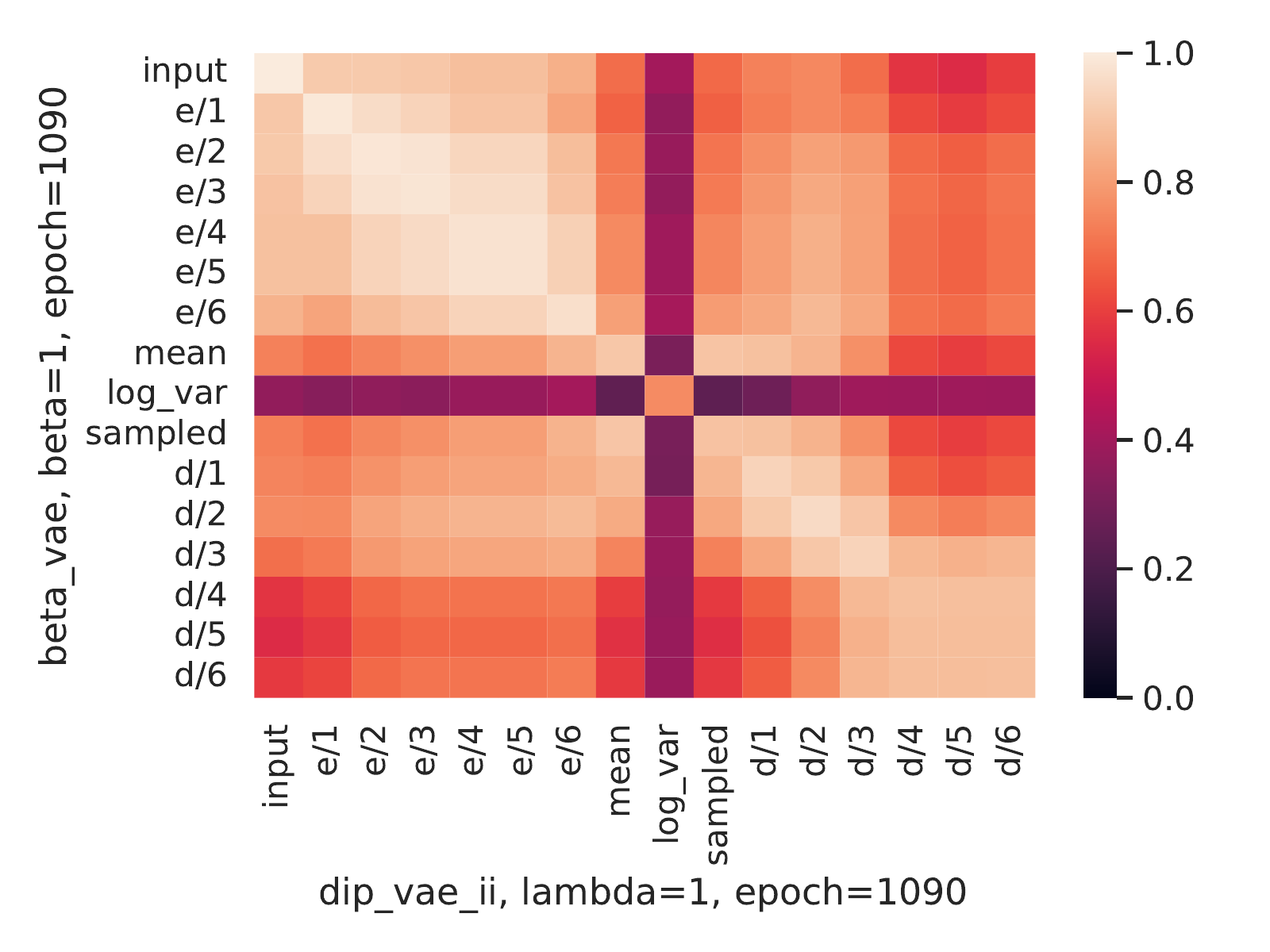}
    }%
    \subcaptionbox{Trained on dSprites\label{fig:methods-dsprites}}{
        \includegraphics[width=0.33\textwidth]{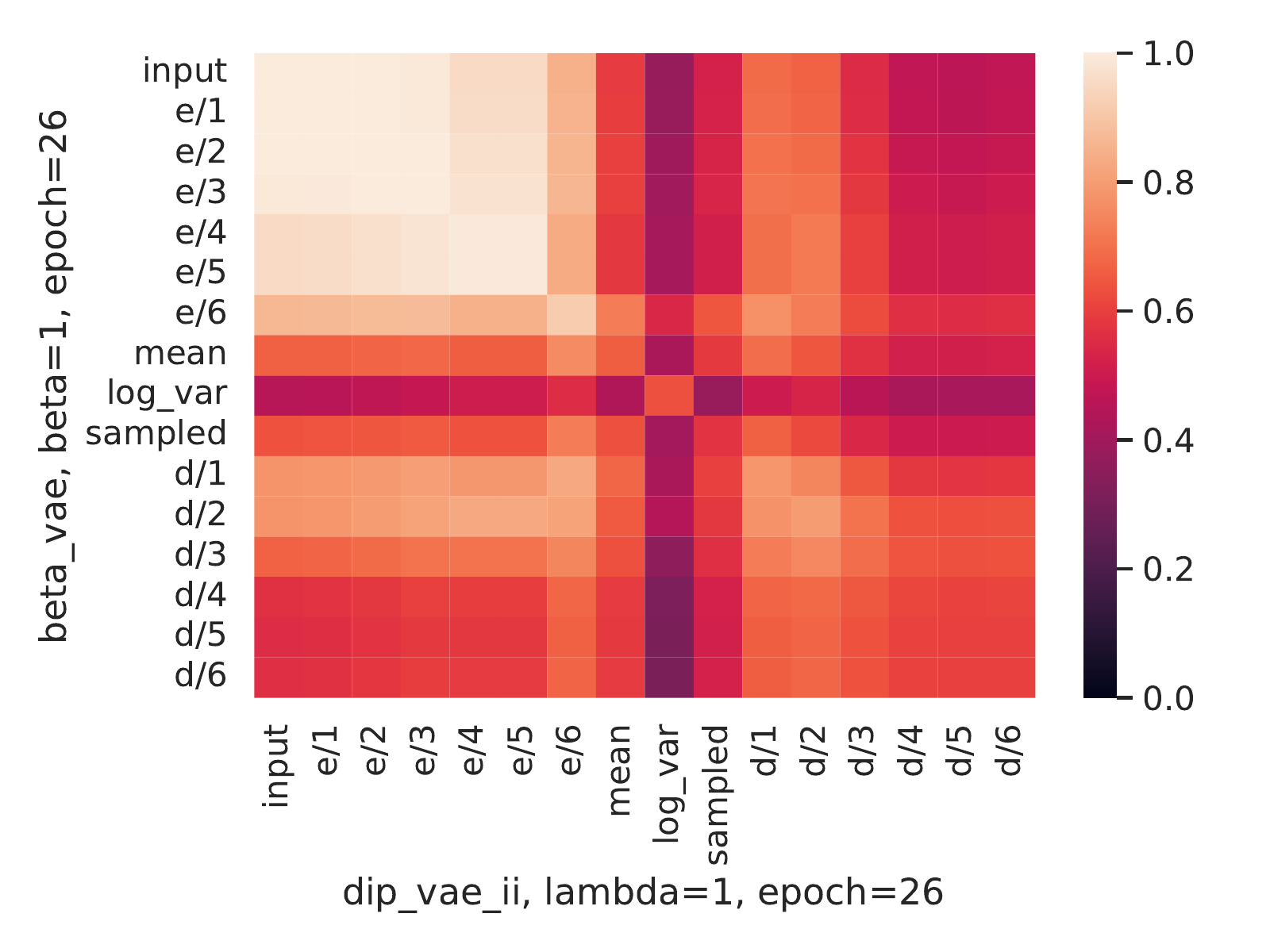}
    }%
    \subcaptionbox{Trained on smallNorb\label{fig:methods-smallnorb}}{
        \includegraphics[width=0.33\textwidth]{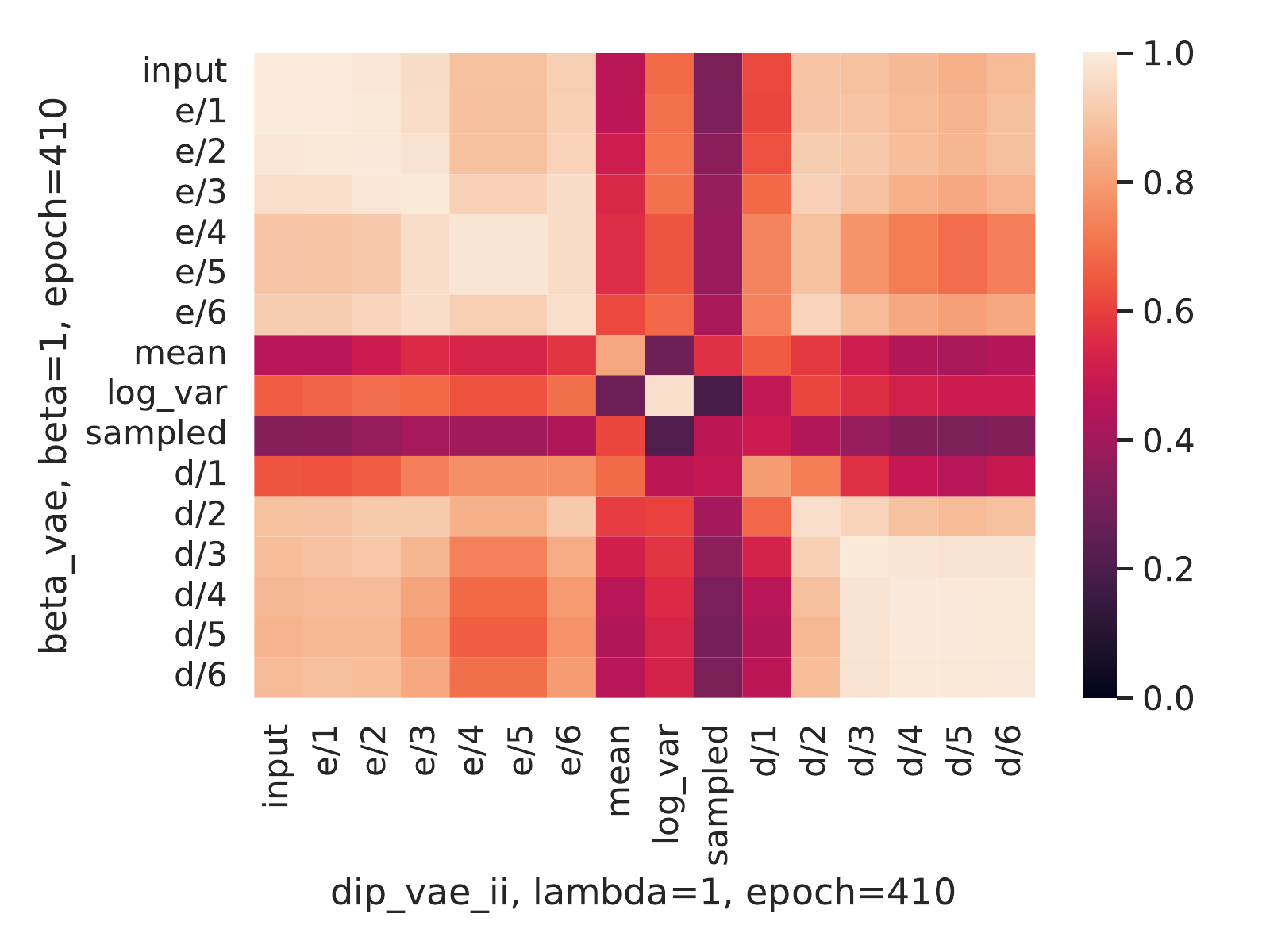}
    }%
    \caption{(a) shows the CKA similarity scores of activations of $\beta$-VAE and DIP-VAE II trained on cars3D with $\beta=1$, and $\lambda=1$, respectively.
             (b) and (c) show the CKA similarity scores of the same learning objectives and regularisation strengths but trained on dSprites and smallNorb.
             These results are averaged over 5 seeds.
             We can see that the representational similarity of all the layers of the encoder (top-left quadrant) except mean and variance is very high (CKA stays close to 1).
             However, the mean, variance, sampled (center diagonal values), and decoder (bottom-right quadrants) representational similarity of different learning objectives seems to vary depending on the dataset.
             In (a) and (c) they have high similarity, while in (b) the similarity is lower.
    }
    \label{fig:methods}
\end{figure}

\paragraph{The representations learned by the early layers of the encoders are similar across learning objectives}
As seen in \Secref{subsec:res-hyperparameters}, the representations learned by the encoder are similar across hyperparameters,
with mean and variance representations being less similar for high regularisation.~In this experiment, we show that this observation also holds across
learning objectives.~We can see in~\Figref{fig:methods} that the representational similarity of different learning objectives with equivalent
regularisation strength is quite high in all the encoder layers except mean and variance.
~This result is consistent across datasets, learning objectives, and well as architectures (see~\Appref{sec:app-fc}), and the
learned representations are also close to what is learned by a classifier with equivalent architecture (see~\Appref{sec:app-clf}).

\paragraph{The representations learned by the mean and variance layers can vary} The similarity of the representations learned
by the mean and variance layers (and consequently by the decoder) tends to vary across learning objectives and seems to be influenced by the dataset.
Indeed, by looking at the center diagonal values of~\Figref{fig:methods}, we can see that
the similarity of the mean, variance, and consequently sampled representations across different learning objectives
is high for cars3D and smallNorb, but low for dSprites.
This may indicate that, for a given dataset, different learning objectives can find different local optima,
leading to lower similarity between mean and variance representations, and ultimately to different representations in the decoder.
~It seems that this phenomenon is very dataset-dependent, and is especially present in dSprites (see~\Figref{fig:methods-dsprites}),
which is one of the most used datasets for disentangled representation learning.
The same phenomenon can be observed with fully-connected architectures in~\Appref{sec:app-fc}.

\paragraph{Implications} The representations learned by most of the encoder layers are very
similar across learning objectives, indicating that the encoder may be learning some general features from the inputs, or
the only features that can be learned when the decoder performs poorly.
Indeed, since the decoder initially struggles to learn, the encoder may only be able to learn very general properties.
~The learning of general representations in early layers is consistent with~\citet{Bansal2021}, who, in the context of
classification, observed that neural networks were learning similar representations regardless of the initialisation,
architecture or learning objective used.~As such, the encoder may be viewed as a feature extractor which is fine-tuned
using a mean and variance layer to produce the sampled representations that will be used by the decoder.
Our observations also suggest that some learning objectives may favour distinct local optima whose existence
has been previously discussed~\citep{Alemi2017,Zietlow2021}.~Such a model-specific choice of local optima may explain
why some learning objectives obtained better disentanglement scores than others on specific datasets but performed worse on
others~\citep{Locatello2019a}.

    \section{Conclusion}\label{sec:conclusion}

\paragraph{Bottom-up learning and posterior collapse}
As reported in~\Secref{subsec:res-training}, the encoder is learned before the decoder,
which could indicate that the decoder struggles to converge before the mean and variance representations are learned.
This would explain why one can observe posterior collapse in a setting where the decoder has access
to the input, and thus can infer the mapping on its own~\citep{Bowman2016,Li2019a}.

\paragraph{Different models encode similar representations}
We have seen in \Secref{sec:results} that the encoders, prior to their mean and variance layers,
learned remarkably similar representations regardless of the initialisation, regularisation, and learning objective used to train
the model.~It is especially intriguing to see that even posterior collapse does not seem to affect these representations.
The representational similarity of the mean, variance, and decoder representations across learning objectives generally
vary depending on the dataset, indicating that different learning objectives may find different local optima
for a given dataset. Note that this behaviour is more visible on some datasets (e.g., dSprites) than others (e.g., cars3D).

\paragraph{Other applications} While our main focus was to compare similarity of models across a variety of
settings, this study also demonstrated that CKA, whose computational cost is very low, can be an efficient tool to detect posterior
collapse.~Indeed, one can directly compare the similarity between mean and sampled representations, which strongly decreases
as the number of collapsed variables grows~\citep{Bonheme2021}.~We believe that this could be a complementary
tool to the more costly mutual information generally used for such purpose.

\paragraph{Limitations} We limited our study to similarity metrics that measure difference in the geometry
of the representations.~While this gave us compelling insights, these metrics have some limitations, discussed in
~\Secref{subsec:bg-limitations}, and may underestimate the similarity between layers with different architectures~\citep{Maheswaranathan2019}.
~We believe that further research using dynamics-based metrics, such as fixed-point
topology~\citep{Maheswaranathan2019}, could provide additional insights into the representations learned by VAEs.
    \ifanonymous
    \else
        \subsubsection*{Acknowledgments}
The authors thank Frances Ding for an insightful discussion on the Procrustes distance, as well as Th\'{e}ophile Champion
and Declan Collins for their helpful comments on the paper.
    \fi
    \clearpage

    \bibliography{main}

    \clearpage
    \appendix
    \section{Disentangled representation learning}\label{sec:app-disentanglement}
As mentioned in ~\Secref{sec:intro}, we are interested in the family of methods modifying the weight on the
regularisation term of~\Eqref{eq:elbo} to encourage disentanglement. In our paper, the term regularisation refers to the moderation of this parameter only.~To achieve this, our experiment will focus on the models described below.

\paragraph{$\beta$-VAE} The goal of this method~\citep{Higgins2017} is to penalise the regularisation
term of~\Eqref{eq:elbo} by a factor $\beta > 1$, such that
\begin{equation}
    \label{eq:Higgins2017}
    \ELBO = \E_{q_{\vphi}(\rvz|\rvx)}\left[\log p_{\vtheta}(\rvx|\rvz)\right]-\beta\KL(q_{\vphi}(\rvz|\rvx)||p(\rvz)).
\end{equation}

\paragraph{Annealed VAE}~\citet{Burgess2018} proposed to gradually increase
the encoding capacity of the network during the training process. The goal is to progressively learn latent variables by
decreasing order of importance.
This leads to the following objective, where \textrm{C} is a parameter that can be understood as a channel capacity and $\gamma$ is a hyper-parameter
penalising the divergence, similarly to $\beta$ in $\beta$-VAE:
\begin{equation}
    \ELBO = \E_{q_{\vphi}(\rvz|\rvx)}\left[\log p_{\vtheta}(\rvx|\rvz)\right]-\gamma\left|\KL(q_{\vphi}(\rvz|\rvx)||p(\rvz))-\mathrm{C}\right|\label{eq:Burgess2018}.
\end{equation}
As the training progresses, the channel capacity $\mathrm{C}$ is increased, going from zero to its maximum channel capacity $\mathrm{C_{max}}$ and allowing a higher value of the KL divergence term.
VAEs that use~\Eqref{eq:Burgess2018} as a learning objective are referred to as Annealed VAEs in this paper.

\paragraph{$\beta$-TC VAE}~\citet{Chen2018} argued that only the distance between the
estimated latent factors and the prior should be penalised to encourage disentanglement, such that
\begin{equation}
    \label{eq:kl-pz-qz}
    \ELBO = \E_{p(\rvx)}\left[\E_{q_{\vphi}(\rvz|\rvx)}\left[\log p_{\vtheta}(\rvx|\rvz)\right]-\KL\left(q_{\vphi}(\rvz|\rvx)||p(\rvz)\right)\right] -\lambda\KL(q_{\vphi}(\rvz)||p(\rvz)).
\end{equation}
Here, $\KL(q_{\vphi}(\rvz)||p(\rvz))$ is approximated by penalising the dependencies between the dimensions of $q_{\vphi}(\rvz)$:
\begin{equation}
    \ELBO \approx \frac{1}{n}\sum_{i=1}^{n}\left[\E_{q_{\vphi}(\rvz|\rvx)}\left[\log p_{\vtheta}(\vx_{i}|\rvz)\right]-\KL\left(q_{\vphi}(\rvz|\vx_{i})||p(\rvz)\right)\right]
    -\underbrace{\lambda\KL(q_{\vphi}(\rvz)||\prod_{j=1}^{\mathrm{D}}q_{\vphi}(\rvz_{j}))}_{\textrm{total correlation}}\label{eq:Kim2018}.
\end{equation}
The total correlation of \Eqref{eq:Kim2018} is then approximated over a mini-batch of samples $\{\vx_1,\dots,\vx_{M}\}$ as follows:
\begin{equation*}
\E_{q_{\vphi}(\rvz)}[\log{q_{\vphi}(\rvz)}]\approx\frac{1}{M}\sum_{i=1}^{M}\left(\log \frac{1}{NM}\sum_{j=1}^{M}q_{\vphi}(z(\vx_i)|\vx_{j})\right),
\end{equation*}
where $z(\vx_{i})$ is a sample from $q_\vphi(\rvz|\vx_i)$, $M$ is the number of samples in the mini-batch, and $N$ the total number of input examples.
$\E_{q_{\vphi}(\rvz_i)}[\log{q_{\vphi}(\rvz_i)}]$ can be computed in a similar way.

\paragraph{DIP-VAE} Similarly to~\citet{Chen2018},~\citet{Kumar2018} proposed to regularise the
distance between $q_\phi(\rvz)$ and $p(\rvz)$ using~\Eqref{eq:kl-pz-qz}.
The main difference is that here $\KL(q_{\phi}(\rvz)||p(\rvz))$ is measured by matching the moments of the learned
distribution $q_{\phi}(\rvz)$ and its prior $p(\rvz)$.
The second moment of the learned distribution is given by
\begin{equation}
    \label{eq:Kumar2018Cov}
    \mathrm{Cov}_{q_{\phi}(\rvz)}[\rvz] = \mathrm{Cov}_{p(\rvx)}\left[\mu_{\phi}(\rvx)\right] + \E_{p(\rvx)}\left[\Sigma_{\phi}(\rvx)\right].
\end{equation}
DIP-VAE II penalises both terms of~\Eqref{eq:Kumar2018Cov} such that
\begin{equation*}
    \lambda\KL(q_{\phi}(\rvz)||p(\rvz)) = \lambda_{od} \sum_{i\neq j}\left(\mathrm{Cov}_{q_{\phi}(\rvz)}\left[\rvz\right]\right)_{ij}^{2}
    + \lambda_{d} \sum_{i} \left(\mathrm{Cov}_{q_{\phi}(\rvz)}\left[\rvz\right]_{ii} -1\right)^{2},
\end{equation*}
where $\lambda_{d}$ and $\lambda_{od}$ are the penalisation terms for the diagonal and off-diagonal values respectively.
    \section{Additional details on mean, variance and sampled representations}\label{sec:app-vae}
\begin{figure}[ht!]
    \centering
    \includegraphics[width=0.7\textwidth]{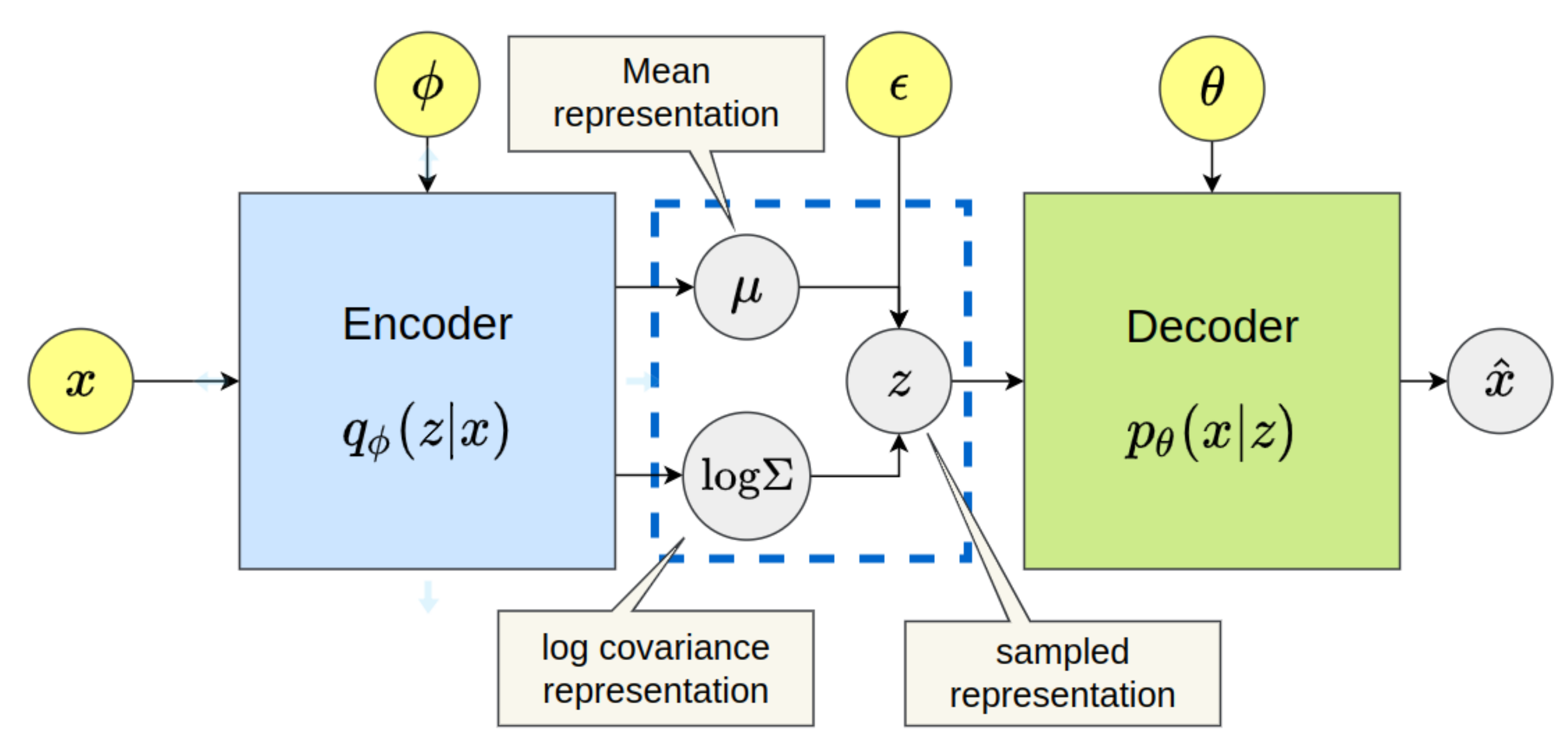}
    \caption{The structure of a VAE}
    \label{fig:vae-archi}
\end{figure}

This section presents a concise illustration of what mean, variance and sampled representations are.
As shown in~\Figref{fig:vae-archi}, the mean, variance and sampled representations are the last 3 layers of the encoder,
where the sampled representation, $\vz$, is the input of the decoder.~These representations, specific to VAEs, influence the models' behaviour that can be quite different from other deep learning models,
as shown by research on polarised regime and posterior collapse, for example.

    \section{Experimental setup}\label{sec:xp-setup}
To facilitate the reproducibility of our experiment, we detail below the Procrustes normalisation process and the configuration used for model training.
\paragraph{Procrustes normalisation} Similarly to~\citet{Ding2021}, given an activation matrix $\mX \in \R^{n \times m}$
containing $n$ samples and $m$ features, we compute the vector $\bar{\vx} \in \R^{m}$ containing the mean values of the columns of $\mX$.
Using the outer product $\otimes$, we get $\bar{\mX} = \vone_n \otimes \bar{\vx}$, where $\vone_n \in \R^{n}$ is a vector of ones
and $\bar{\mX} \in \R^{n \times m}$. We then normalise $\mX$ such that
\begin{equation}
        \dot{\mX} = \frac{\mX - \bar{\mX}}{\norm{\mX - \bar{\mX}}_F}.
\end{equation}
As the Frobenius norm of $\dot{\mX}$ and $\dot{\mY}$ is 1, and $\norm{\dot{\mY}^T\dot{\mX}}_*$ is always positive (1 when
$\dot{\mX} = \dot{\mY}$, smaller otherwise), \Eqref{eq:pd} lies in $[0,2]$, and \Eqref{eq:ps} in $[0, 1]$.

\paragraph{VAE training} Our implementation uses the same hyperparameters as~\citet{Locatello2019a}, and the details are listed in~\Twotablerefs{table:global-hyperparam}{table:model-hyperparam}.
We reimplemented~\citet{Locatello2019a} code base, designed for Tensorflow 1, in Tensorflow 2 using Keras.
The model architecture used is also identical, as described in~\Tableref{table:architecture}.
Each model is trained 5 times, on seeded runs with seed values from 0 to 4.
Intermediate models are saved every 1,000 steps for smallNorb, 6,000 steps for cars3D and 11,520 steps for dSprites.
Every image input is normalised to have pixel values between 0 and 1.

For the fully-connected models presented in~\Appref{sec:app-fc}, we used the same architecture and hyperparameters as those implemented in \texttt{disentanglement lib} of~\citet{Locatello2019a}, and the details are presented in~\Twotablerefs{table:linear-architecture}{table:linear-model-hyperparam}.

\begin{table}[h!]
    \centering
    \caption{Shared hyperparameters}
    \label{table:global-hyperparam}
    \begin{tabular}{ l l }
        \hline
        Parameter & Value \\
        \hline
        Batch size & 64  \\
        Latent space dimension & 10  \\
        Optimizer & Adam \\
        Adam: $\beta_1$ & 0.9 \\
        Adam: $\beta_2$ & 0.999 \\
        Adam: $\epsilon$ & 1e-8 \\
        Adam: learning rate & 0.0001 \\
        Reconstruction loss & Bernoulli \\
        Training steps & 300,000 \\
        Intermediate model saving & every 6K steps\\
        Train/test split & 90/10\\
        \hline
    \end{tabular}
\end{table}

\begin{table}[h!]
    \centering
    \caption{Model-specific hyperparameters}
    \label{table:model-hyperparam}
    \begin{tabular}{ l l l }
        \hline
        Model & Parameter & Value \\
        \hline
        $\beta$-VAE & $\beta$ & [1, 2, 4, 6, 8] \\
        $\beta$-TC VAE & $\beta$ & [1, 2, 4, 6, 8] \\
        DIP-VAE II & $\lambda_{od}$ & [1, 2, 5, 10, 20] \\
        & $\lambda_{d}$ & $\lambda_{od}$ \\
        Annealed VAE & $C_{max}$ & [5, 10, 25, 50, 75] \\
        & $\gamma$ & 1,000 \\
        & iteration threshold & 100,000 \\
        \hline
    \end{tabular}
\end{table}

\begin{table}[h!]
    \centering
    \caption{Shared architecture}
    \label{table:architecture}
    \begin{tabularx}{\linewidth}{ X X }
        \hline
        Encoder & Decoder \\
        \hline
        Input: $\R^{64 \times 63 \times channels}$ & $\R^{10}$ \\
        Conv, kernel=4×4, filters=32, activation=ReLU, strides=2 & FC, output shape=256, activation=ReLU \\
        Conv, kernel=4×4, filters=32, activation=ReLU, strides=2 & FC, output shape=4x4x64, activation=ReLU \\
        Conv, kernel=4×4, filters=64, activation=ReLU, strides=2 & Deconv, kernel=4×4, filters=64, activation=ReLU, strides=2 \\
        Conv, kernel=4×4, filters=64, activation=ReLU, strides=2 & Deconv, kernel=4×4, filters=32, activation=ReLU, strides=2 \\
        FC, output shape=256, activation=ReLU, strides=2 & Deconv, kernel=4×4, filters=32, activation=ReLU, strides=2 \\
        FC, output shape=2x10 & Deconv, kernel=4×4, filters=channels, activation=ReLU, strides=2 \\
        \hline
    \end{tabularx}
\end{table}

\begin{table}[h!]
    \centering
    \caption{Fully-connected architecture}
    \label{table:linear-architecture}
    \begin{tabularx}{\linewidth}{ X X }
        \hline
        Encoder & Decoder \\
        \hline
        Input: $\R^{64 \times 63 \times channels}$ & $\R^{10}$ \\
        FC, output shape=1200, activation=ReLU & FC, output shape=256, activation=tanh \\
        FC, output shape=1200, activation=ReLU & FC, output shape=1200, activation=tanh \\
        FC, output shape=2x10 & FC, output shape=1200, activation=tanh \\
        \hline
    \end{tabularx}
\end{table}

\begin{table}[h!]
    \centering
    \caption{Hyperparameters of fully-connected models}
    \label{table:linear-model-hyperparam}
    \begin{tabular}{ l l l }
        \hline
        Model & Parameter & Value \\
        \hline
        $\beta$-VAE & $\beta$ & [1, 8, 16] \\
        $\beta$-TC VAE & $\beta$ & [2] \\
        DIP-VAE II & $\lambda_{od}$ & [1, 20, 50] \\
        & $\lambda_{d}$ & $\lambda_{od}$ \\
        Annealed VAE & $C_{max}$ & [5] \\
        & $\gamma$ & 1,000 \\
        & iteration threshold & 100,000 \\
        \hline
    \end{tabular}
\end{table}
    \section{Consistency of the results with Procrustes Similarity}\label{sec:procrustes}
\begin{figure}[ht!]
    \centering
    \subcaptionbox{CKA\label{fig:heatmap-cka}}{
        \includegraphics[width=0.5\textwidth]{heatmaps/cars3d/beta_tc_vae_2_epoch_25_beta_tc_vae_2_epoch_1090}
    }%
    \subcaptionbox{Procrustes\label{fig:heatmap-procrustes}}{
        \includegraphics[width=0.5\textwidth]{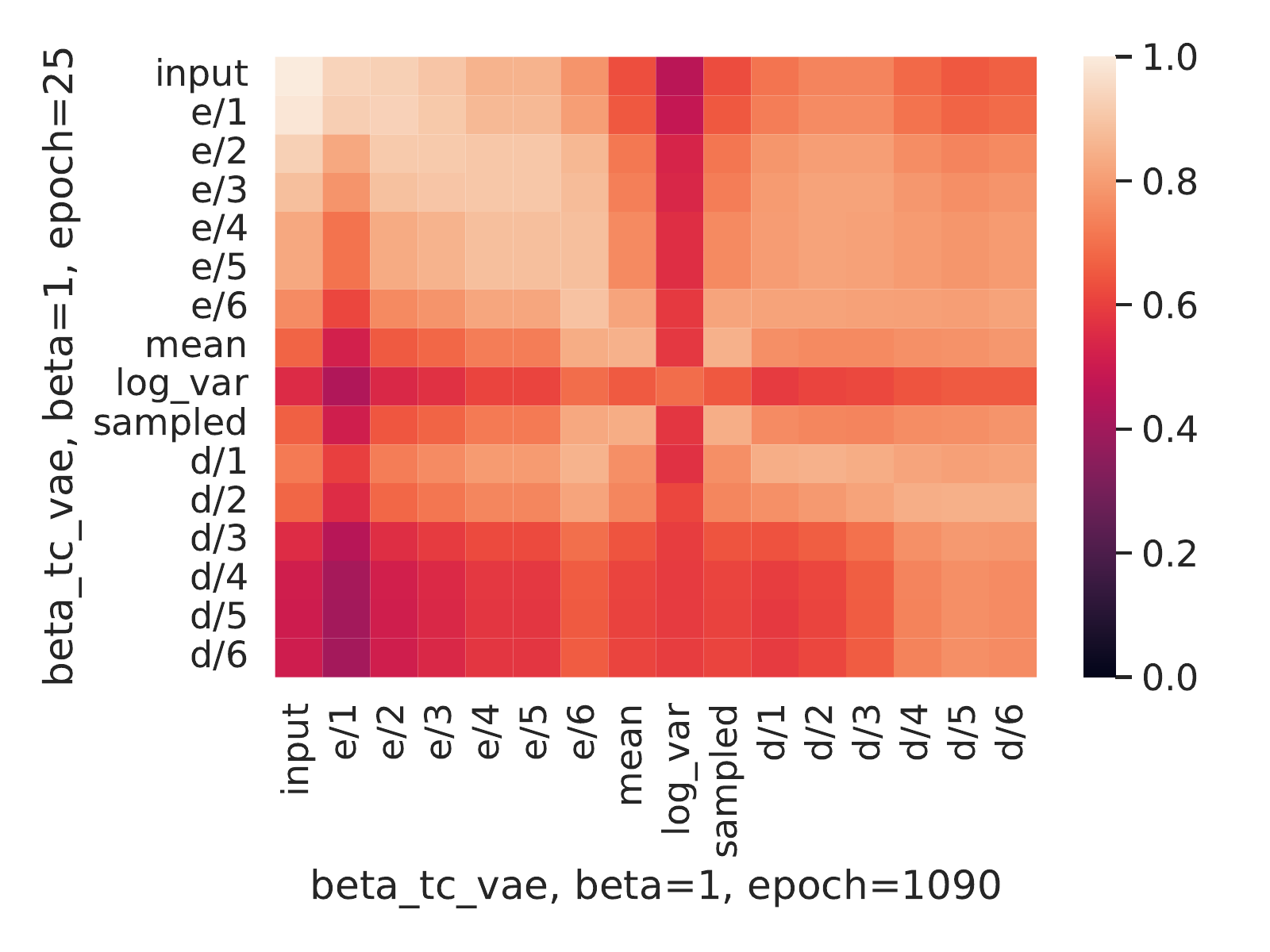}
    }%
    \caption{As reported in~\Figref{fig:heatmap-1} of the main paper, (a) shows the CKA similarity scores of activations
        at epochs 25 and 1090 of $\beta$-TC VAE trained on cars3D with $\beta=2$.
        (b) shows the Procrustes similarity scores of the same configuration.~We observe the same trend with both metrics
        with Procrustes slightly overestimating the similarity between high dimensional activations (bottom-right quadrants), which agrees with the properties of the Procrustes similarity reported in \Secref{subsec:bg-limitations}.
        }
    \label{fig:training-sim-comp}
\end{figure}

\begin{figure}[ht!]
    \centering
    \subcaptionbox{CKA\label{fig:methods-cka}}{
        \includegraphics[width=0.5\textwidth]{heatmaps/cars3d/beta_vae_1_epoch_1090_dip_vae_ii_1_epoch_1090}
    }%
     \subcaptionbox{Procrustes\label{fig:methods-procrustes}}{
        \includegraphics[width=0.5\textwidth]{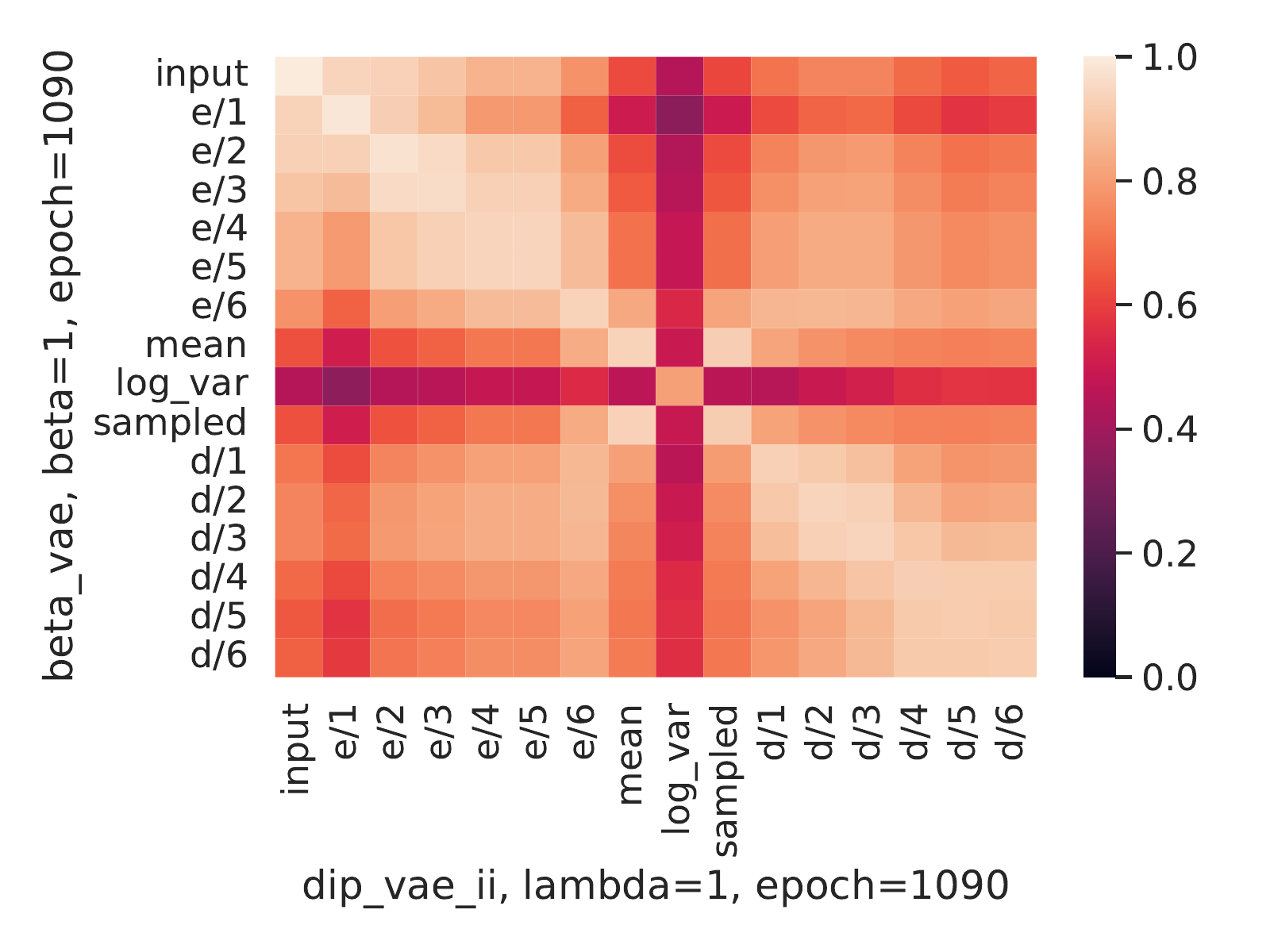}
    }%
    \caption{As in~\Figref{fig:methods-cars}, (a) shows the CKA similarity scores of activations of $\beta$-VAE and DIP-VAE II trained on cars3D with $\beta=1$, and $\lambda=1$, respectively.
        (b) shows the Procrustes similarity scores using the same configuration.~We observe the same trend with both metrics
        with Procrustes slightly overestimating the similarity between high dimensional activations (bottom-right quadrants) (c.f.~\Secref{subsec:bg-limitations}).
    }
    \label{fig:methods-comp}
\end{figure}

\begin{figure}[ht!]
    \centering
    \subcaptionbox{CKA for $\beta=1$\label{fig:enc-1-cka}}{
        \includegraphics[width=0.5\textwidth]{input_encoder/cka_cars3d_beta_vae_1_input_encoder}
    }%
    \subcaptionbox{CKA for $\beta=8$\label{fig:enc-2-cka}}{
        \includegraphics[width=0.5\textwidth]{input_encoder/cka_cars3d_beta_vae_8_input_encoder}
    }%
    \\
    \subcaptionbox{Procrustes for $\beta=1$\label{fig:enc-1-procrustes}}{
        \includegraphics[width=0.5\textwidth]{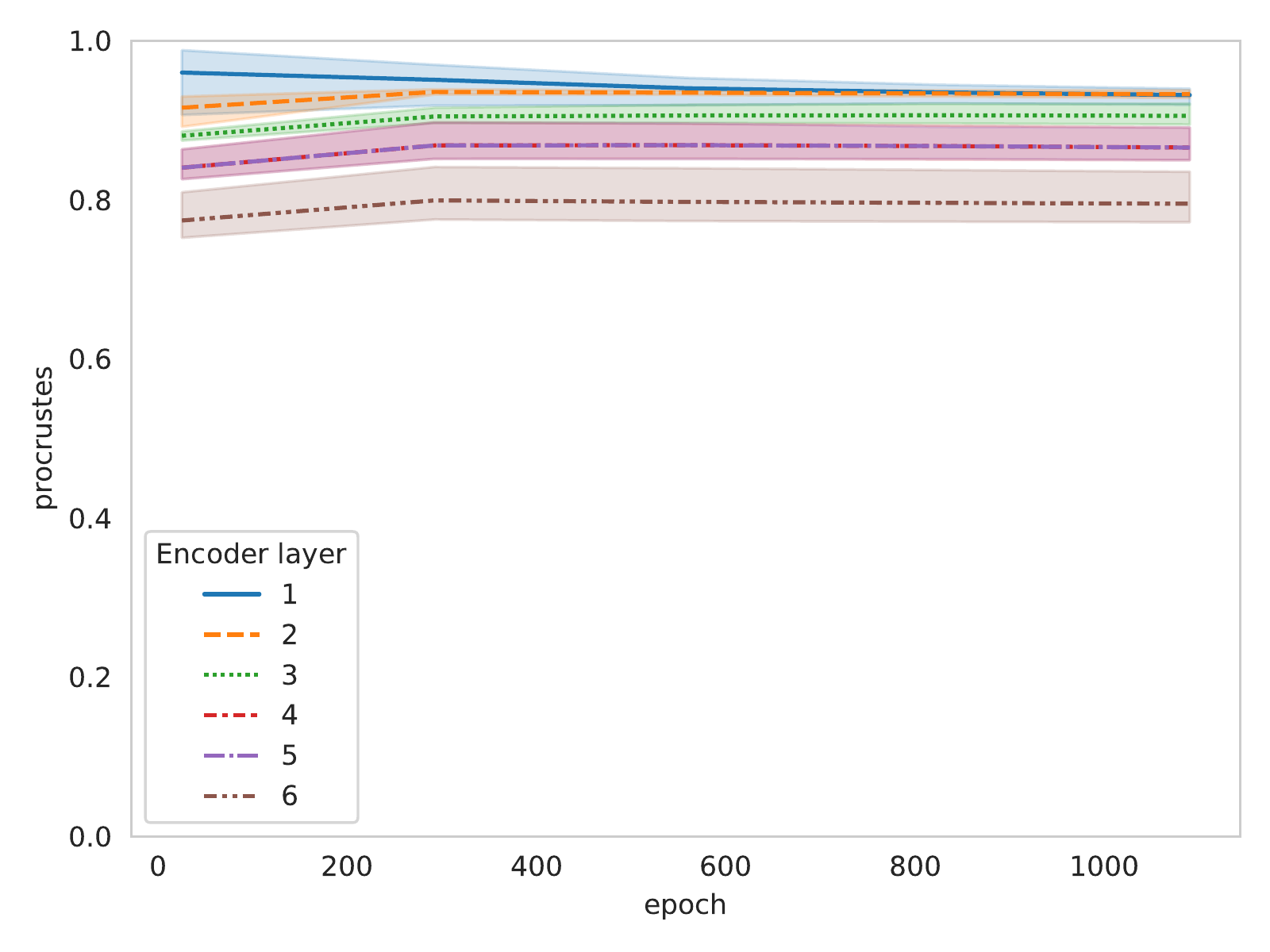}
    }%
    \subcaptionbox{Procrustes for $\beta=8$\label{fig:enc-2-procrustes}}{
        \includegraphics[width=0.5\textwidth]{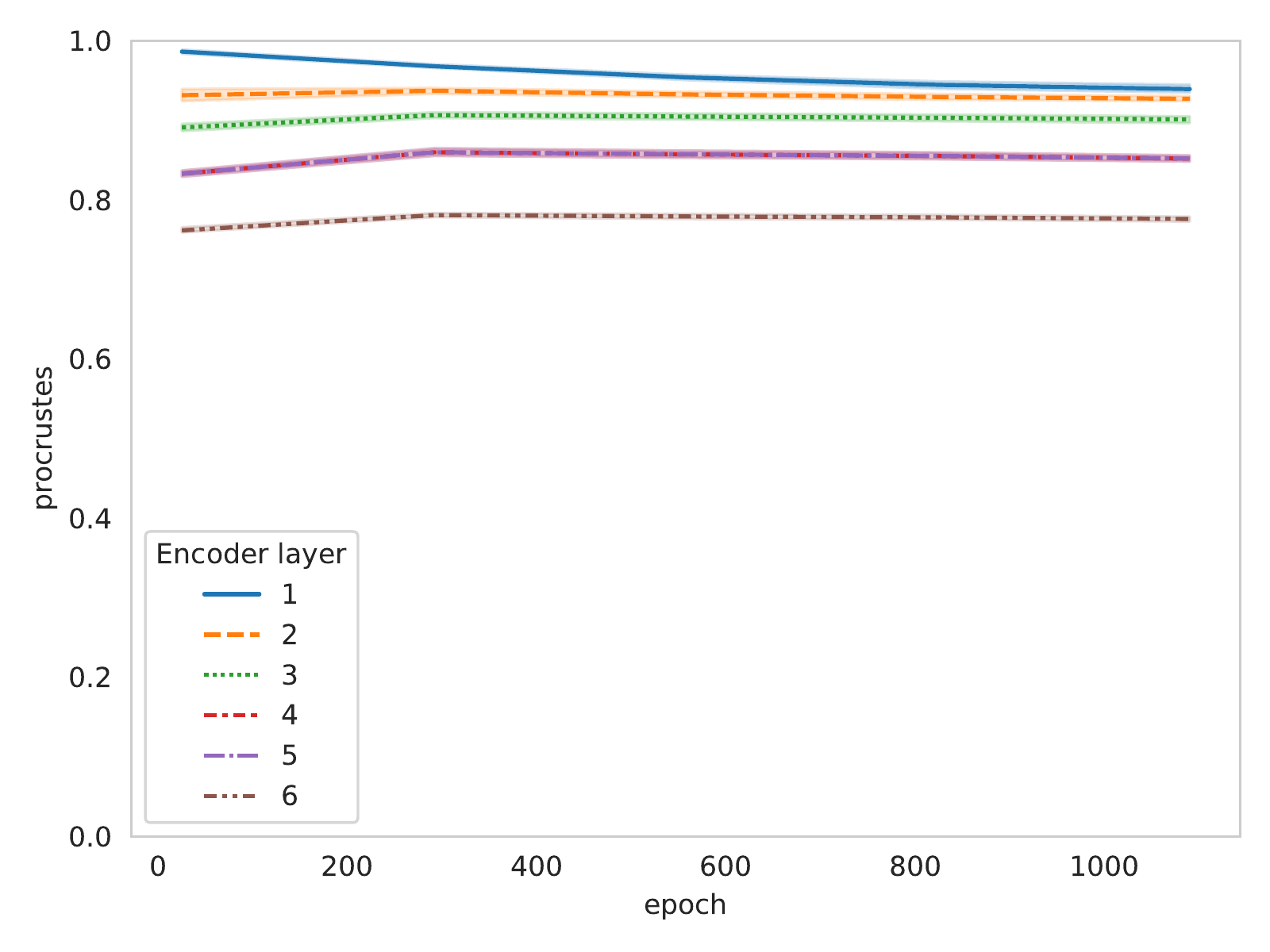}
    }%
     \caption{(a) shows the CKA scores between the inputs and the activations of the first 6 layers of the encoder of a
        $\beta$-VAE trained on cars3D with $\beta=1$.~(b) shows the scores between the same representations with $\beta=8$.
         (a) and (b) are discussed in~\Figref{fig:reg-encoder} of the main paper.
         (c) and (d) are the Procrustes scores of the same configurations.~We observe the same trend for both metrics
        with more variance in (c) for Procrustes with $\beta=1$.~Procrustes also displays a slightly lower similarity for layers 4 to 6 of the encoder,
        possibly due to changes in the representation that are underestimated by CKA (c.f.~\Secref{subsec:bg-limitations}).}
    \label{fig:reg-encoder-comp}
\end{figure}

As mentioned in~\Secref{sec:experiment}, in this section we provide a comparison between the CKA scores reported in the
main paper, and the Procrustes scores for the cars3D dataset.
We can see in \Figrangeref{fig:training-sim-comp}{fig:reg-encoder-comp} that Procrustes and CKA provide similar results.
%~\Figref{fig:tsne-comp} shows a similar clustering pattern between the epoch number and the regularisation strength in both metrics.
~\Figrangetworef{fig:training-sim-comp}{fig:methods-comp} show that Procrustes tends to overestimate the similarity between
high-dimensional inputs, as mentioned in~\Secref{subsec:bg-limitations} (recall the example given in~\Figref{fig:cka-procrustes}).
In~\Figref{fig:reg-encoder-comp}, we observe a slightly lower similarity with Procrustes than CKA on the $5^{th}$ and $6^{th}$ layers
of the encoder, indicating that some small changes in the representations may have been underestimated by CKA,
as discussed in~\Secref{subsec:bg-limitations} and by~\citet{Ding2021}.~Note that the difference between the CKA and
Procrustes similarity scores in~\Figref{fig:reg-encoder-comp} remains very small (around 0.1) indicating consistent
results between both metrics.

%\begin{figure}[ht!]
%    \centering
%    \subcaptionbox{CKA\label{fig:tsne-reg-cka}}{
%        \includegraphics[width=0.5\textwidth]{tsne/cka_cars3d_beta_vae_tsne_regularisation}
%    }%
%    \subcaptionbox{Procrustes\label{fig:tsne-reg-procrustes}}{
%        \includegraphics[width=0.5\textwidth]{tsne/procrustes_cars3d_beta_vae_tsne_regularisation}
%    }%
%    \caption{t-SNE of the similarity between models at different epochs and with different regularisation strength. To improve t-SNE's stability, we applied PCA and reduced the dimensionality of its inputs to 50.
%        (a) is the CKA similarity originally presented in~\Figref{fig:tsne-reg} of the main paper, (b) is the Procrustes similarity.
%        ~We observe the same trend with both metrics, with clear clusters based on the epoch (especially the early ones)
%        and regularisation strength.}
%    \label{fig:tsne-comp}
%\end{figure}

    \clearpage
\section{Resources}\label{sec:app-ressources}
As mentioned in~\Twosecrefs{sec:intro}{sec:experiment}, we released the code of our experiment, the pre-trained models
and similarity scores:
\ifanonymous
\begin{itemize}
    \item The similarity scores can be downloaded from an anonymous Google account using the following tiny URL~\url{https://t.ly/0GLe3}
    \item The code can also be downloaded from an anonymous Google account using another tiny URL~\url{https://t.ly/VMIm}
    \item Our pre-trained models are large (around 80 GB in total), and it was not feasible to make them available to the reviewers using an anonymous link. The URL to the models will, however, be available in the non-anonymised version of this paper.
\end{itemize}
\else
\begin{itemize}
    \item The similarity scores can be downloaded at~\url{https://data.kent.ac.uk/444/}
    \item The pre-trained models can be downloaded at~\url{https://data.kent.ac.uk/428/}
    \item The code is available at~\url{https://github.com/bonheml/VAE_learning_dynamics}
\end{itemize}
\fi
    \section{CKA on fully-connected architectures}\label{sec:app-fc}
\begin{figure}
    \centering
    \subcaptionbox{Epoch 1\label{fig:lep-1}}{
        \includegraphics[width=0.33\textwidth]{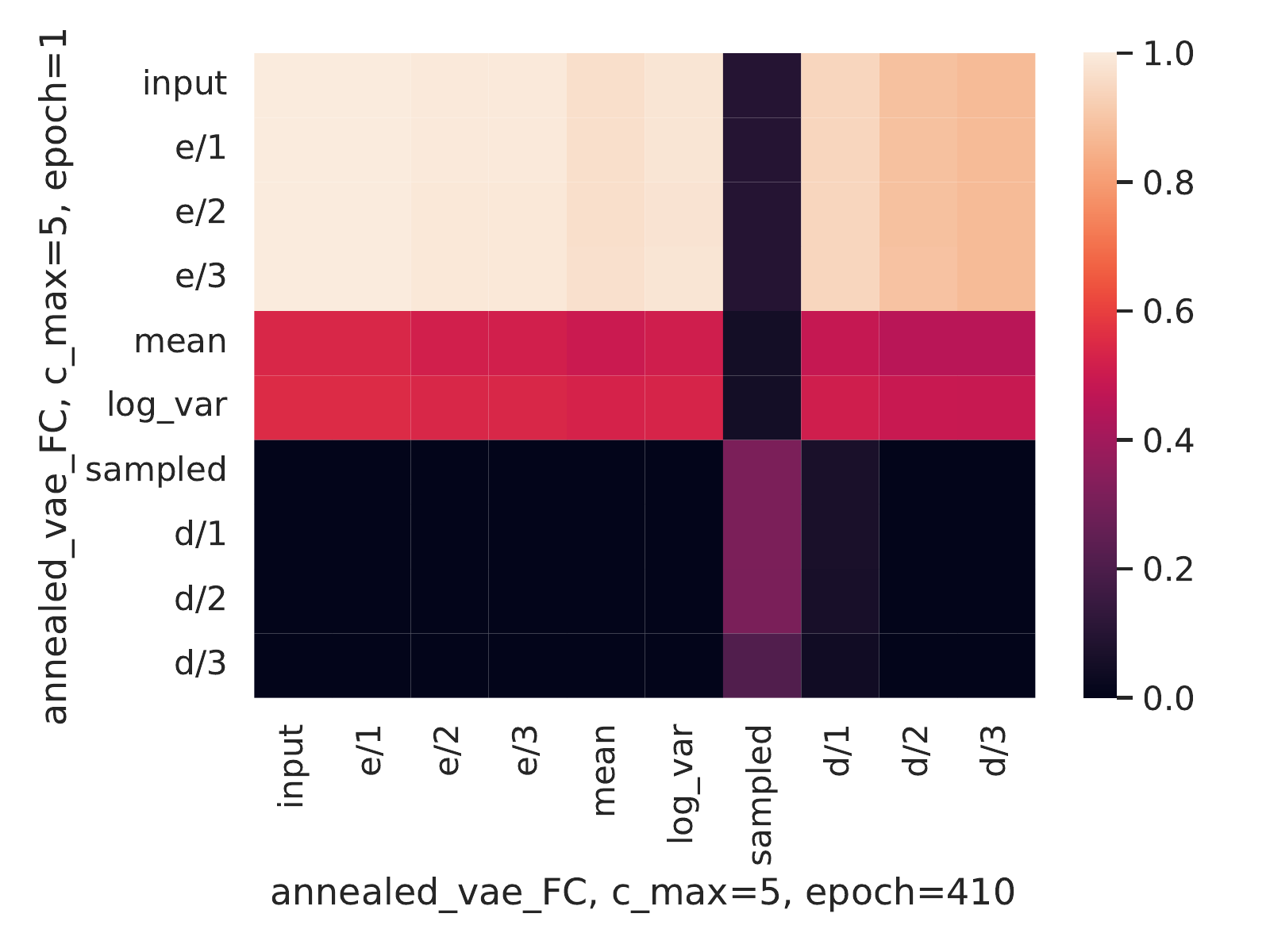}
    }%
    \subcaptionbox{Epoch 10\label{fig:lep-2}}{
        \includegraphics[width=0.33\textwidth]{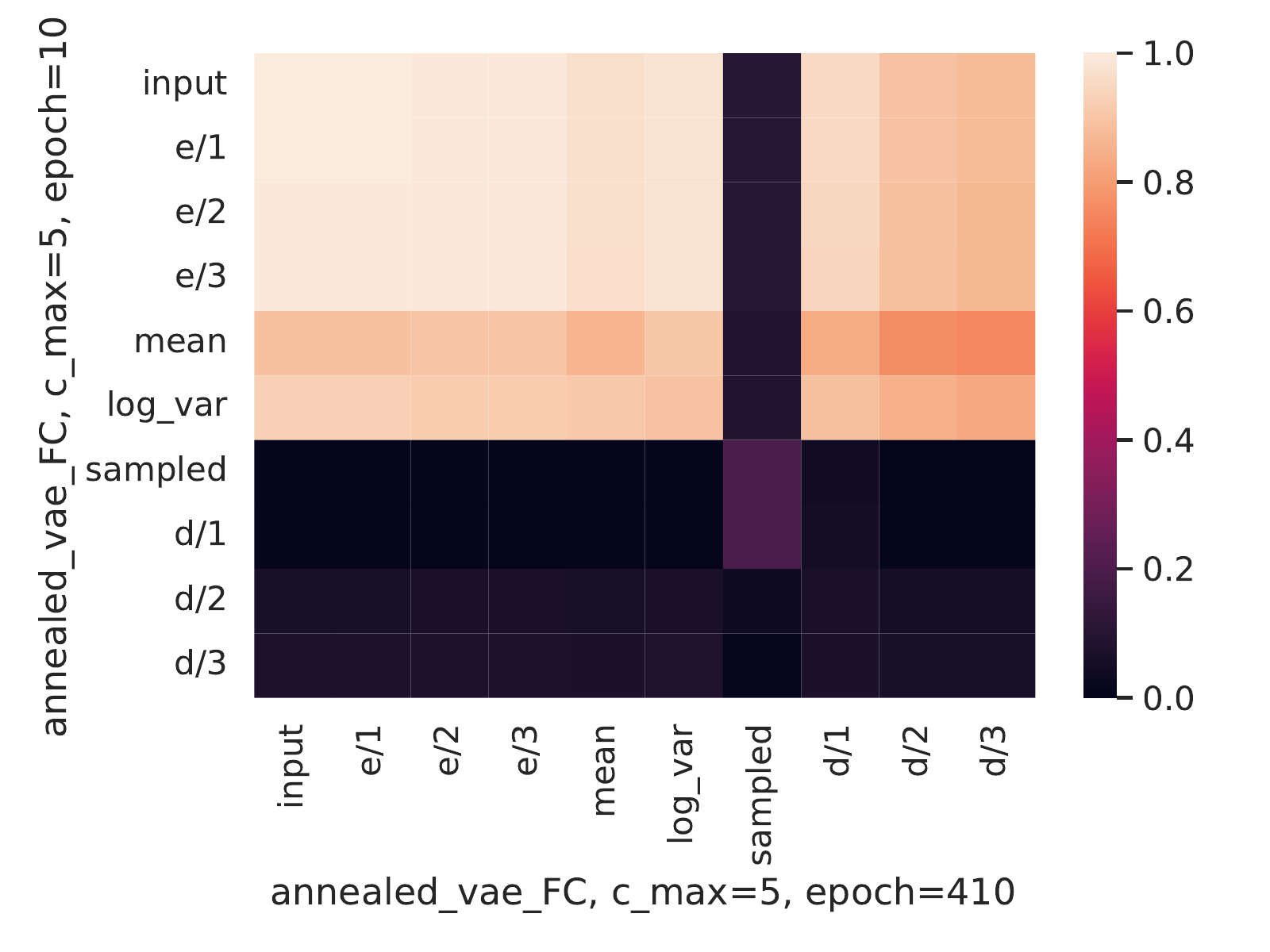}
    }%
    \subcaptionbox{Epoch 410\label{fig:lep-3}}{
        \includegraphics[width=0.33\textwidth]{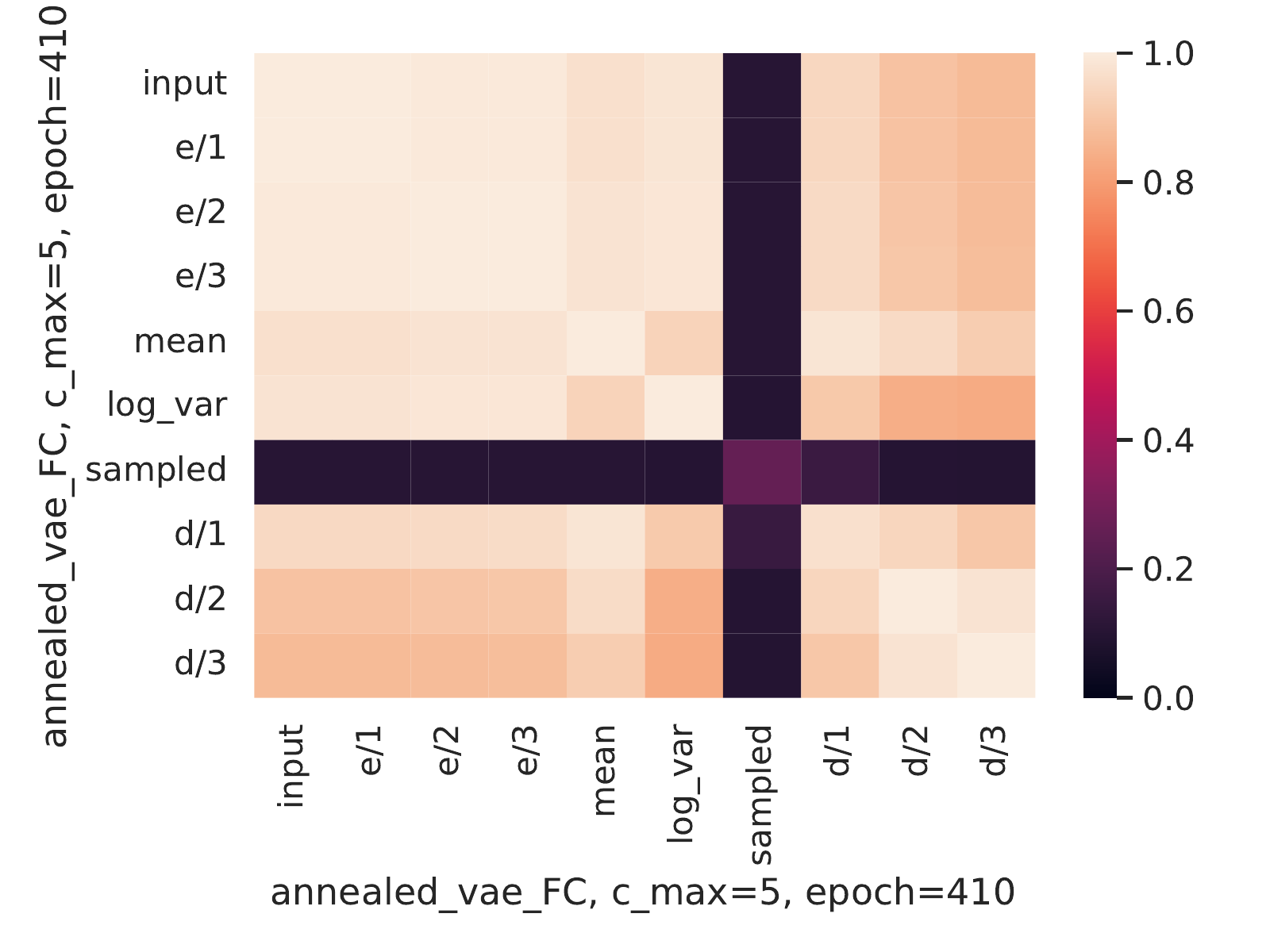}
    }%
    \caption{(a), (b), and (c) show CKA scores between a fully-trained fully-connected Annealed VAE and a fully-connected Annealed VAE trained
    for 1, 10 and 410 epochs, respectively. All the models are trained on smallNorb and the results are averaged over 5 seeds.
     Similarity to~\Figref{fig:heatmap-3} of~\Secref{subsec:res-training}, we can see that there is
        a high similarity between the representations learned by the encoder early in the training and after
        complete training (see the bright cells in the top-left quadrants in (a), (b), and (c)). The
        mean and variance representations similarity with a fully trained model increase after a few more epochs (the purple line in the middle disappear between (a) and (b)),
        and finally the decoder is learned (see bright cells in bottom-right quadrant of (c)).}
    \label{fig:lep}
\end{figure}

In order to assess the generalisability of our findings, we have repeated our observations on the fully-connected VAEs that are described in~\Appref{sec:xp-setup}.
We can see in~\Threefigref{fig:lep}{fig:lreg}{fig:methods-linear} that the same general trend as for the convolutional architectures can be identified
(see~\Threefigref{fig:heatmap-3}{fig:reg-1}{fig:methods} of ~\Secrefs{subsec:res-training}{subsec:res-hyperparameters}{subsec:res-objective} for a comparison with convolutional networks).

\paragraph{Learning in fully-connected VAEs is also bottom-up}
We can see in~\Figref{fig:lep} that, similarly to the convolutional architectures shown in \Figref{fig:heatmap-3}, the encoder is learned early in the
training process.~Indeed between epochs 1 and 10, the encoder representations become highly similar to the representations of the
fully trained model (see~\Twofigref{fig:lep-1}{fig:lep-2}).~The decoder is then learned with its representational similarity
with the fully trained decoder raising after epoch 10 (see~\Figref{fig:lep-3}).

\begin{figure}
    \centering
    \subcaptionbox{$\beta = 8$\label{fig:lreg-1}}{
        \includegraphics[width=0.33\textwidth]{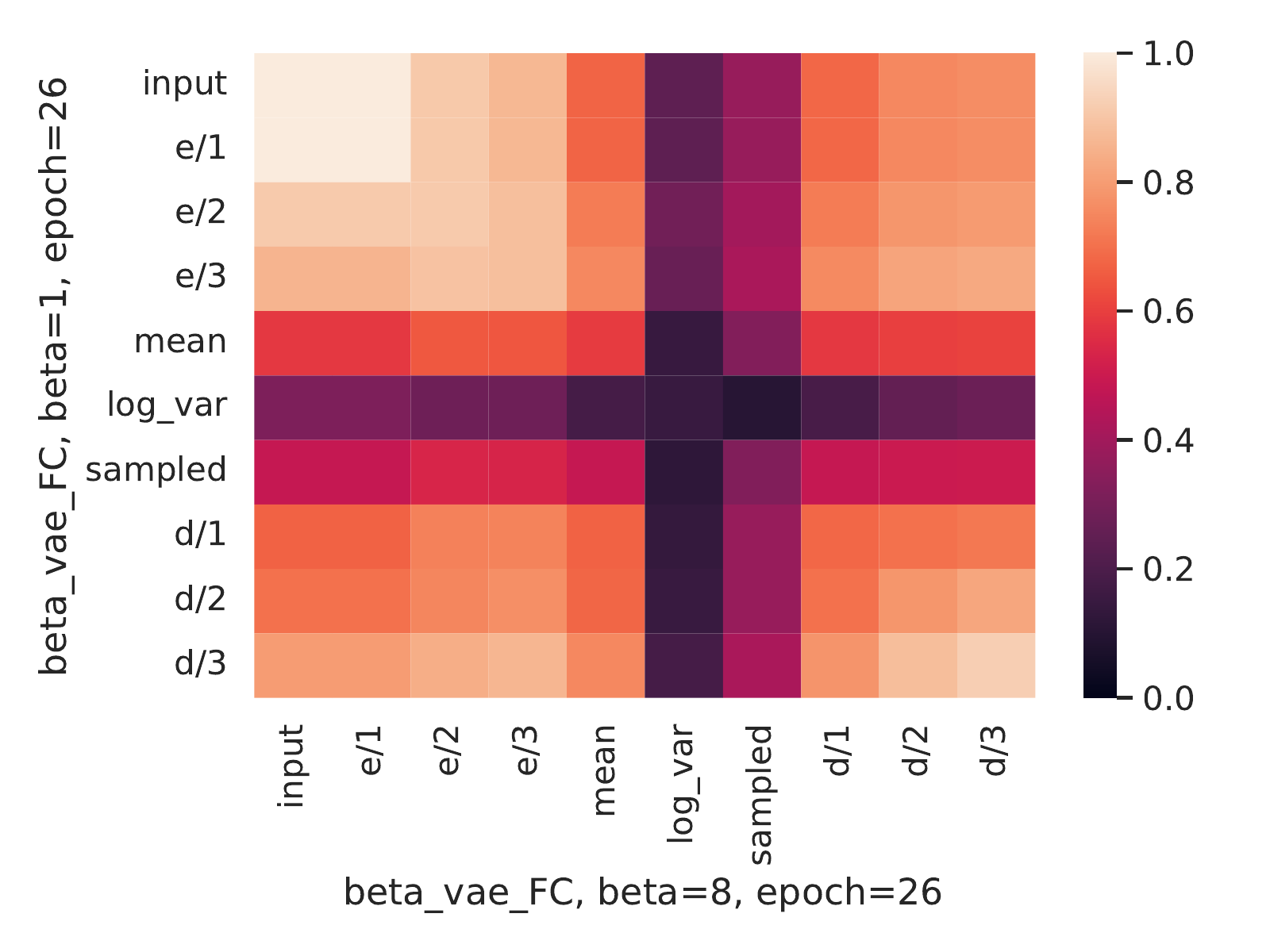}
    }%
    \subcaptionbox{$\beta = 16$\label{fig:lreg-2}}{
        \includegraphics[width=0.33\textwidth]{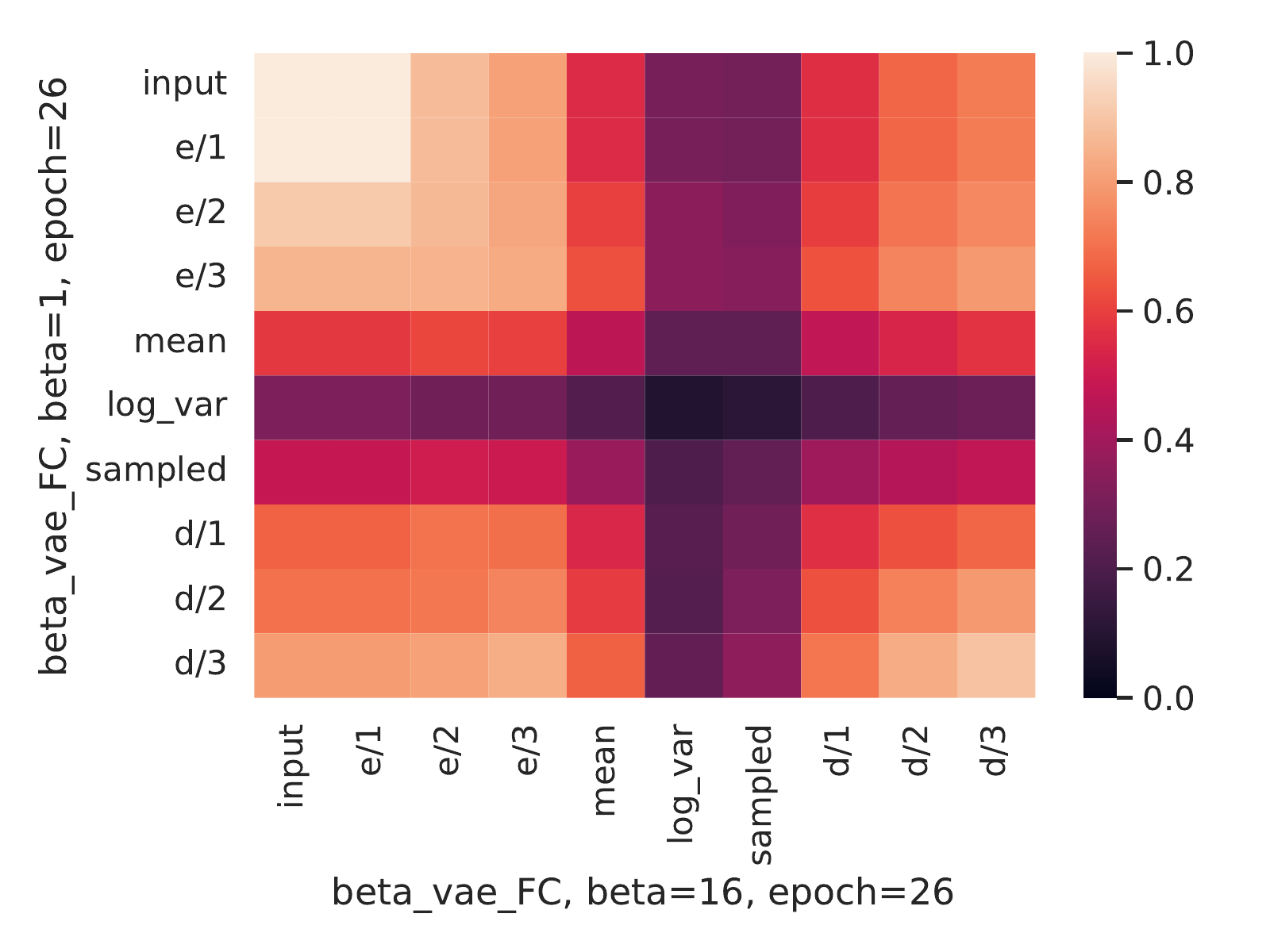}
    }
    \caption{(a) and (b) show the representational similarity between fully-connected $\beta$-VAEs trained with $\beta = 1$,
        and fully-connected $\beta$-VAEs trained with $\beta = 8$ and $\beta = 16$, respectively. All models are trained on dSprites and
        the scores are averaged over 5 seeds. In both figures, the encoder representations stay very similar (bright cells in the top-left quadrants),
        except for the mean, variance and sampled representations. While the variance representation is increasingly different as we increase $\beta$, the
        decoder does not show the dramatic dissimilarity observed in convolutional architectures in~\Figref{fig:reg-1}.}
    \label{fig:lreg}
\end{figure}

\paragraph{Impact of regularisation}
As in convolutional architectures shown in~\Figref{fig:reg-1}, the variance and sampled representations retain little similarity with the encoder representations
in the case of posterior collapse, as shown in~\Figref{fig:lreg}.~Interestingly, in fully-connected architectures the decoder retains more
similarity with its less regularised version than in convolutional architectures, despite suffering from poor reconstruction when heavily regularised.
Thus, CKA of the representations of fully-connected decoders may not be a good predictor of reconstruction quality.
Despite this difference, it can still be used to monitor posterior collapse with fully-connected architecture by relying on the similarity scores
between the encoder representations, and the mean, variance and sampled representations. This property is consistent between both architectures.

\begin{figure}[ht!]
    \centering
    \subcaptionbox{Trained on cars3D\label{fig:lmethods-cars}}{
        \includegraphics[width=0.33\textwidth]{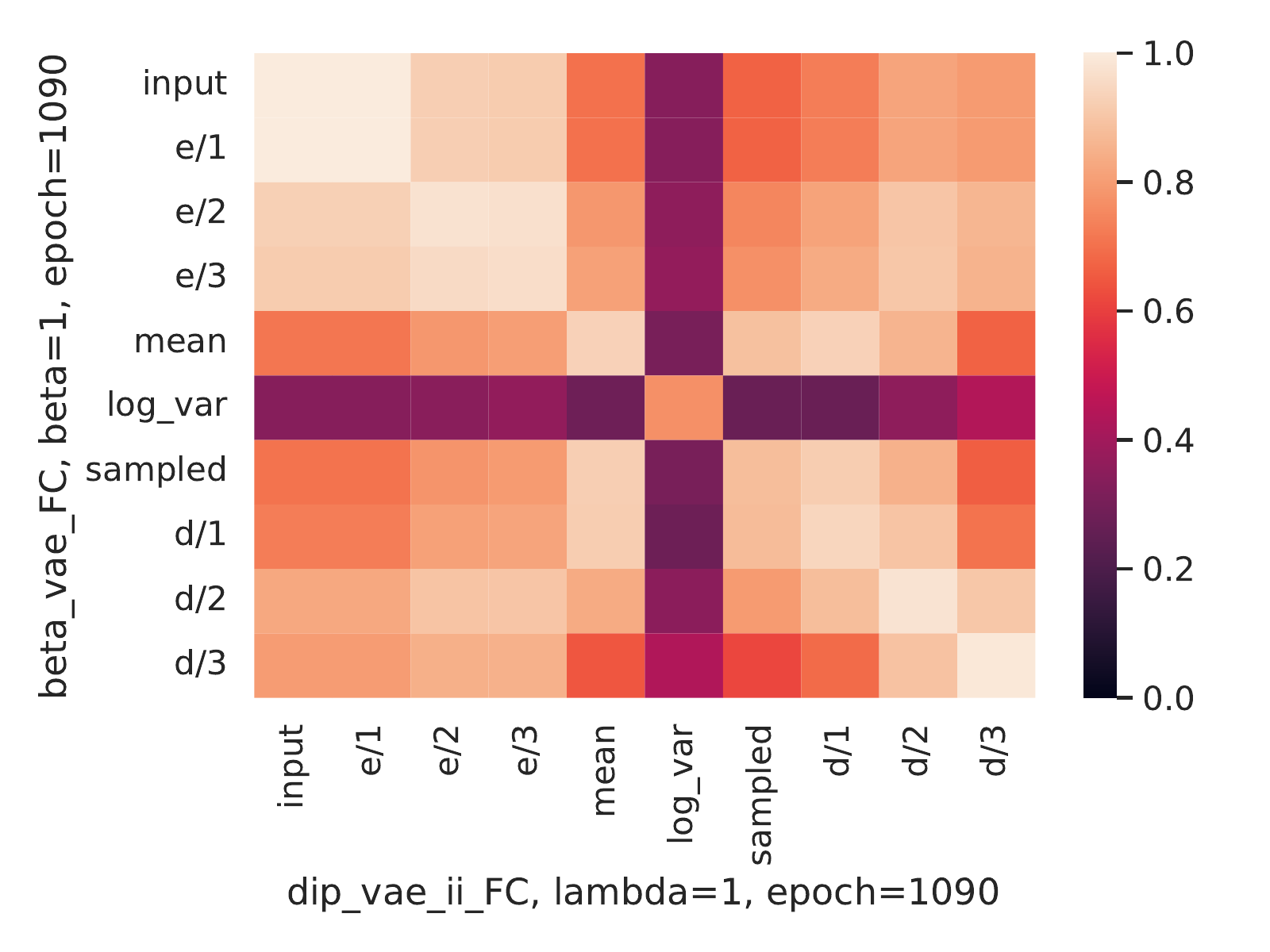}
    }%
    \subcaptionbox{Trained on dSprites\label{fig:lmethods-dsprites}}{
        \includegraphics[width=0.33\textwidth]{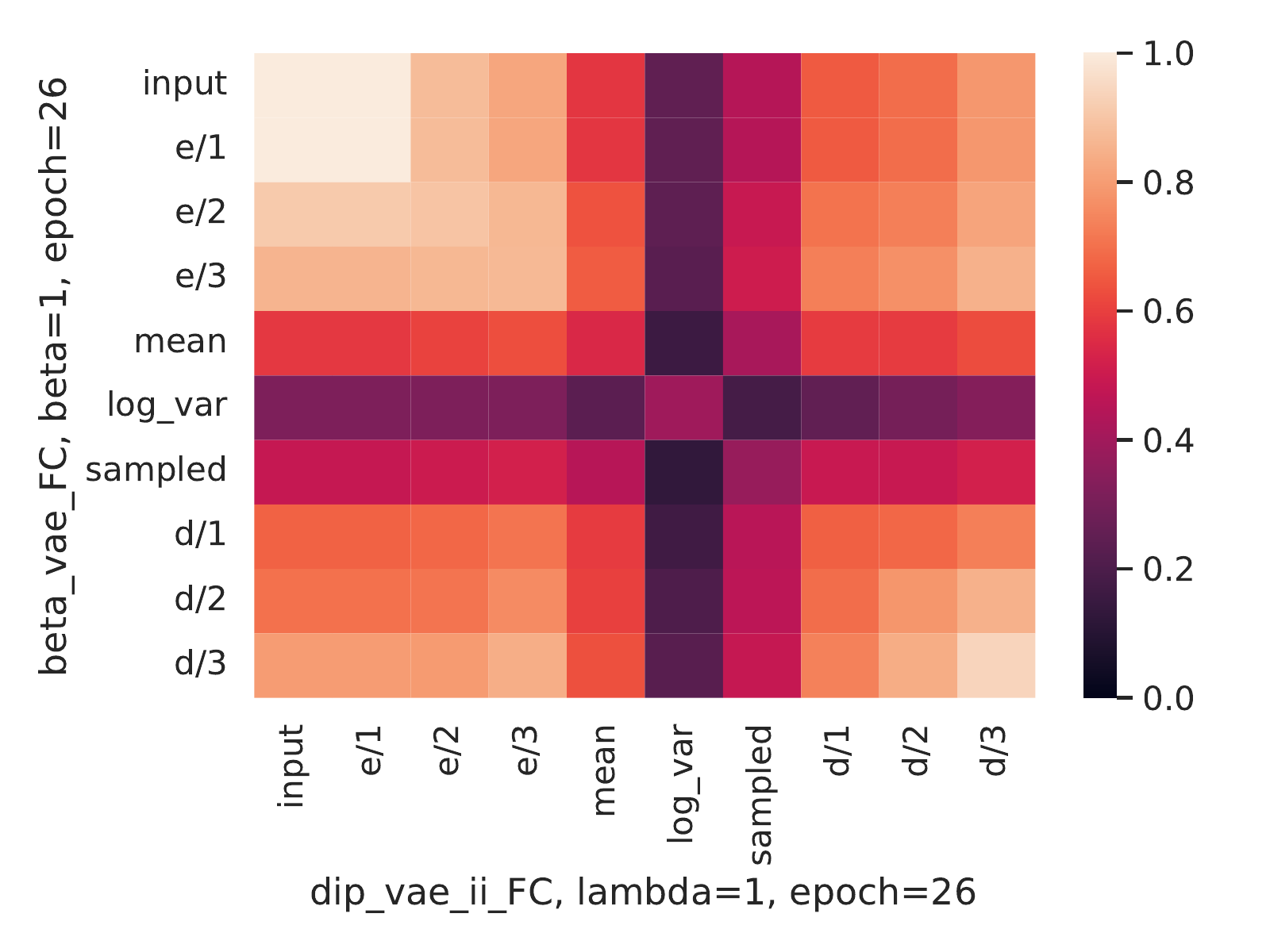}
    }%
    \subcaptionbox{Trained on smallNorb\label{fig:lmethods-smallnorb}}{
        \includegraphics[width=0.33\textwidth]{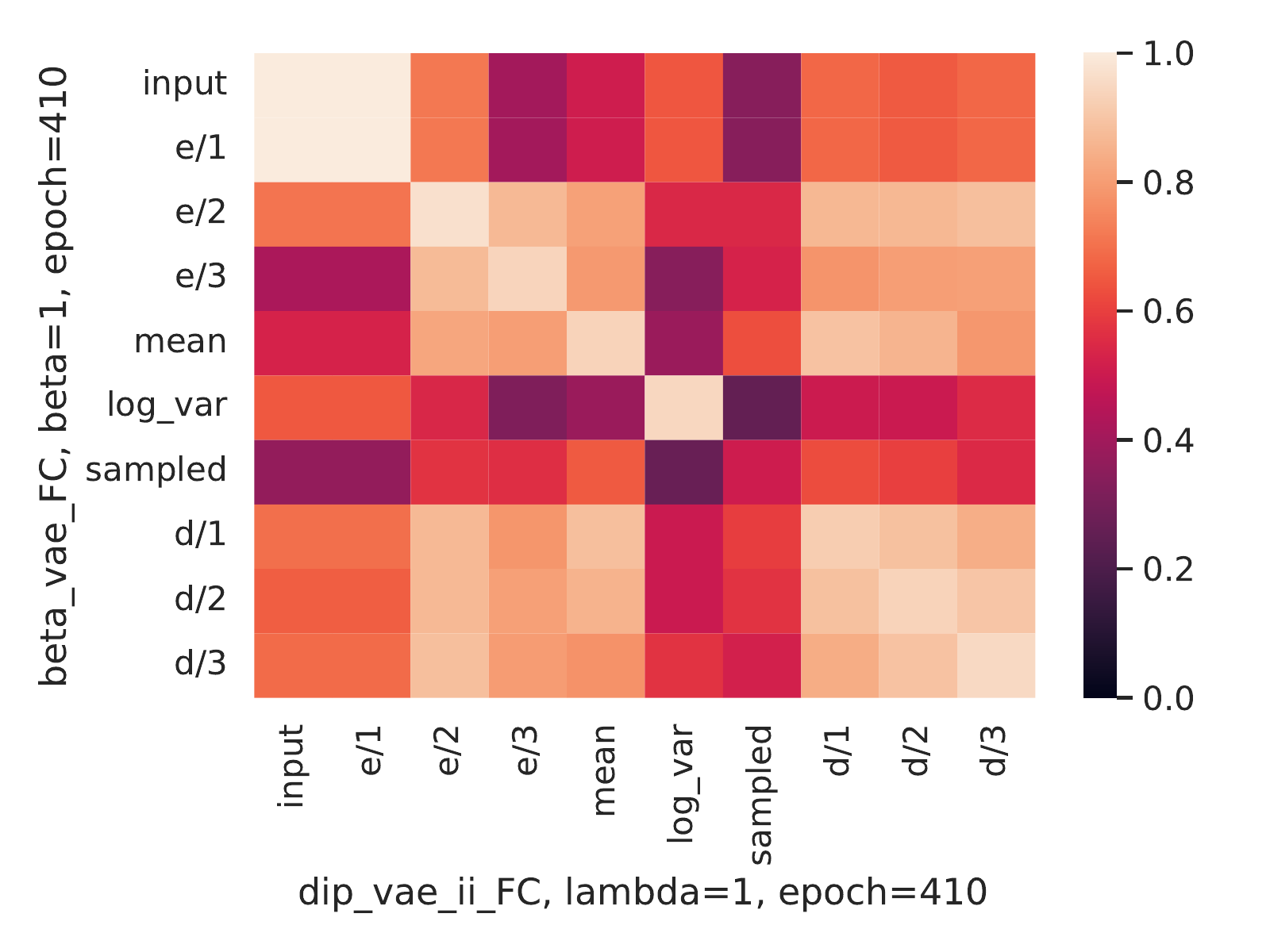}
    }%
    \caption{(a) shows the CKA similarity scores of activations of $\beta$-VAE and DIP-VAE II with fully-connected architectures trained on cars3D with $\beta=1$, and $\lambda=1$, respectively.
             (b) and (c) show the CKA similarity scores of the same learning objectives and regularisation strengths but trained on dSprites and smallNorb.
             All the results are averaged over 5 seeds.
             We can see that the representational similarity of all the layers of the encoder (top-left quadrant) except mean and variance is high (CKA $\geqslant 0.8$).
             However, as in~\Figref{fig:methods}, the mean, variance, sampled (center diagonal values), and decoder (bottom-right quadrants) representational similarity of different learning objectives seems to vary depending on the dataset.
             In (a) and (c) they have high similarity, while in (b) the similarity is lower. Moreover, in (c) the input and first layer of the encoder are quite distinct from the
             other representations, which was not the case in convolutional architectures.
    }
    \label{fig:methods-linear}
\end{figure}

\paragraph{Impact of learning objective}
\Figref{fig:methods-linear} provides results similar to the convolutional VAEs observed in~\Figref{fig:methods}, with a very high similarity between encoder layers
learned from different learning objectives (see diagonal values of the upper-left quadrant).~Here again, the representational similarity
of the decoder seems to vary depending on the dataset, even though this is less marked than for convolutional architectures.
We can also see that the representational similarity between different layers of the encoder vary depending on the dataset,
which was less visible in convolutional architectures.~For example, the similarity between the first and subsequent
layers of the encoder in smallNorb is much lower in fully-connected VAEs. Given that smallNorb is a hard dataset to learn for VAEs~\citep{Locatello2019a},
one could hypothesise that the encoder of fully-connected VAE, being less powerful, is unable to retain as much information as its convolutional counterpart,
leading to lower similarity scores with the representations of the first encoder layer.

    \section{How well does CKA distinguish polarised regime from posterior collapse?}\label{sec:app-latents}

\begin{figure}[ht!]
    \centering
    \subcaptionbox{$\beta=1$\label{fig:latent-20}}{
        \includegraphics[width=0.5\textwidth]{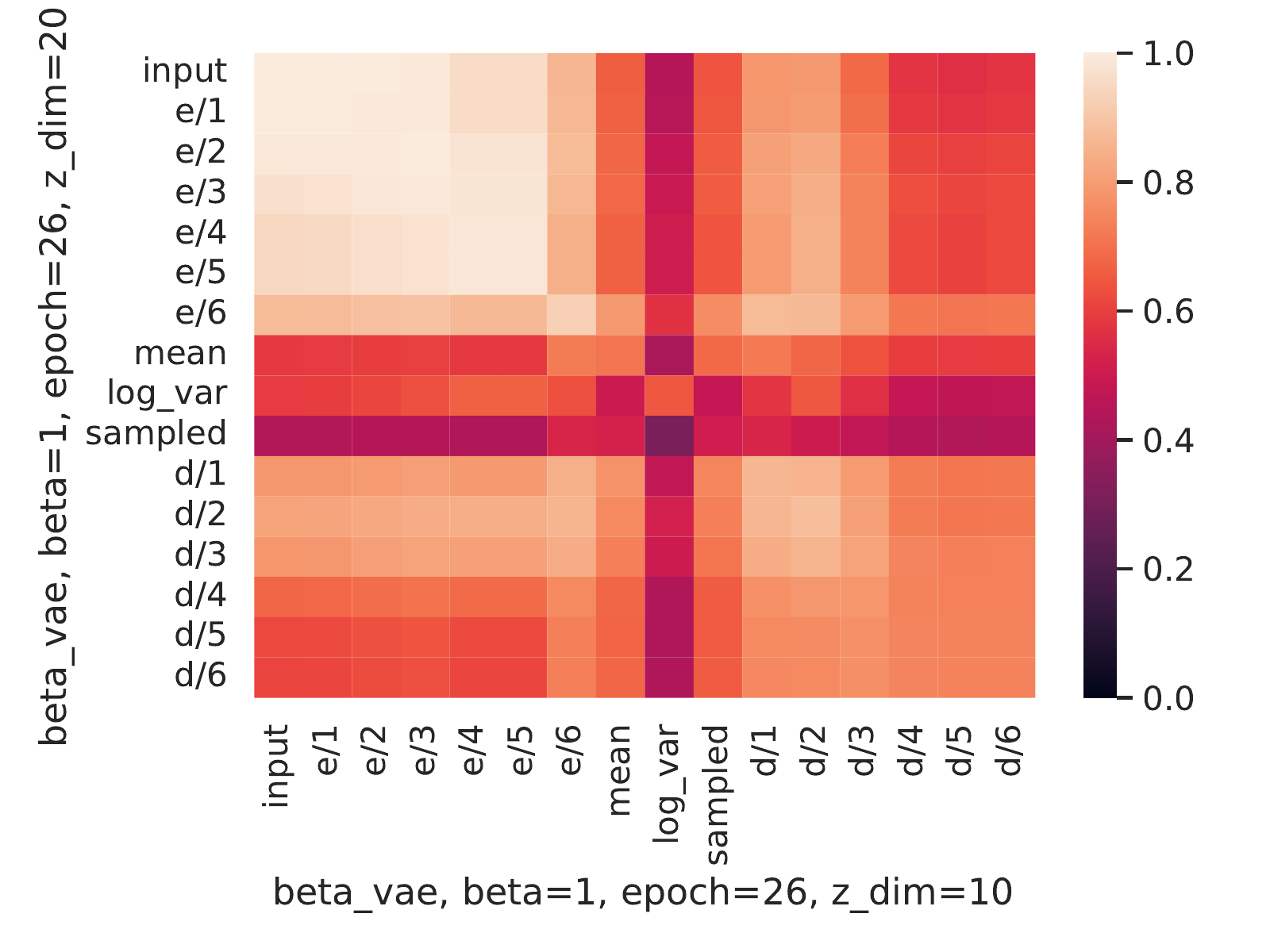}
    }%
    \subcaptionbox{$\beta=8$\label{fig:reg-10}}{
        \includegraphics[width=0.5\textwidth]{heatmaps/dsprites/beta_vae_1_epoch_26_beta_vae_8_epoch_26}
    }%
     \caption{(a) shows the CKA similarity scores between the activations of two $\beta$-VAEs trained on dSprites with $\beta=1$, the first with 10 latent variables, and the second with 20.
        (b) shows the CKA similarity scores between the activations of two $\beta$-VAEs with 10 latent variables trained on dSprites with $\beta=1$, and $\beta=8$.
        For (a) and (b), the activations are taken after complete training, and all the results are averaged over 5 seeds. We can see that we retain a higher similarity in the
        decoder representations in the case of polarised regime (bottom-right quadrant in (a)) than posterior collapse (see the darker bottom-right quadrant in (b)).
        The representational similarity of the mean, variance and sampled representations is also higher in the case of polarised regime than posterior collapse.}
    \label{fig:latent-impact}
\end{figure}

In~\Secref{subsec:res-hyperparameters}, we stated that CKA could be a useful tool to monitor posterior collapse.
However, one can wonder whether CKA can lead to false positives when VAEs contain passive variables for
non-pathological reasons.~For example, due to the polarised regime, if one provides more latent variables than
needed by the VAEs, some of the variables will be collapsed to reduce the KL divergence.~As in posterior collapse, the decoder will
ignore these passive variables.~However, contrary to collapsed models, when passive variables are a result of the
polarised regime, the decoder will still have access to meaningful information and will be able to
correctly reconstruct the image, learning similar representations as a good model with fewer latent variables.
We can see in~\Figref{fig:latent-impact} that, in opposition to posterior collapse, the variance and sampled representations
retain much higher similarity scores with the representations learned by other layers.~Thus, one can differentiate between
the two scenarios using CKA. Given that the CKA scores for the variance and sampled representations vary similarly in fully-connected architectures,
CKA seems to consistently be a good predictor of posterior collapse across learning objectives and architectures while being
robust to false positives.
While one could be tempted to monitor posterior collapse using the changes of similarity scores in the decoder, we have seen in~\Appref{sec:app-fc}
that the fully-connected decoders could retain a relatively high similarity in the case of posterior collapse.
Thus, we recommend relying on the CKA scores of the mean, variance and sampled representations for a better robustness across architectures.

    \section{How similar are the representations learned by encoders and classifiers?}\label{sec:app-clf}
\begin{figure}[ht!]
    \centering
    \subcaptionbox{Trained on cars3D\label{fig:clf-cars}}{
        \includegraphics[width=0.33\textwidth]{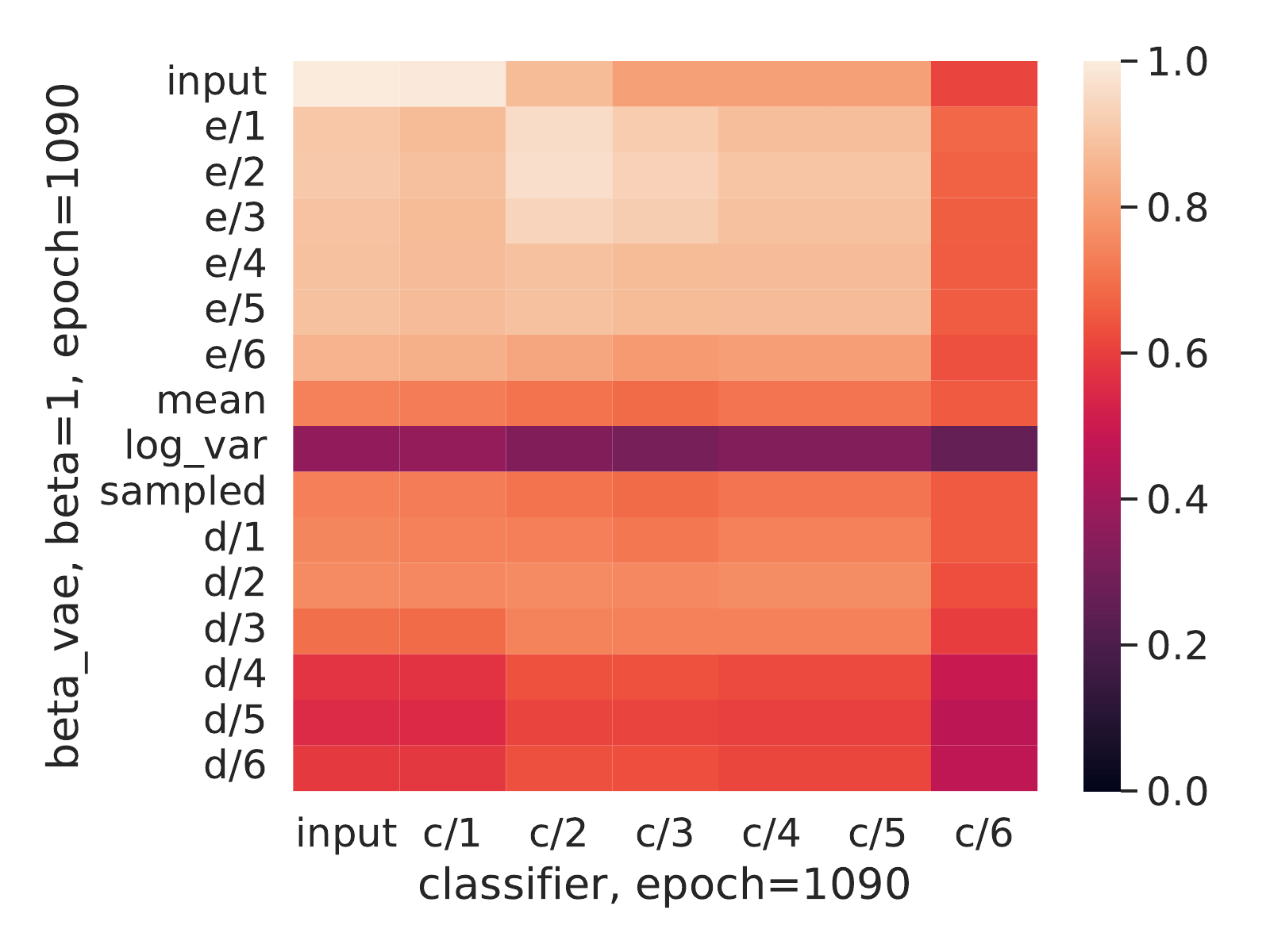}
    }%
    \subcaptionbox{Trained on dSprites\label{fig:clf-dsprites}}{
        \includegraphics[width=0.33\textwidth]{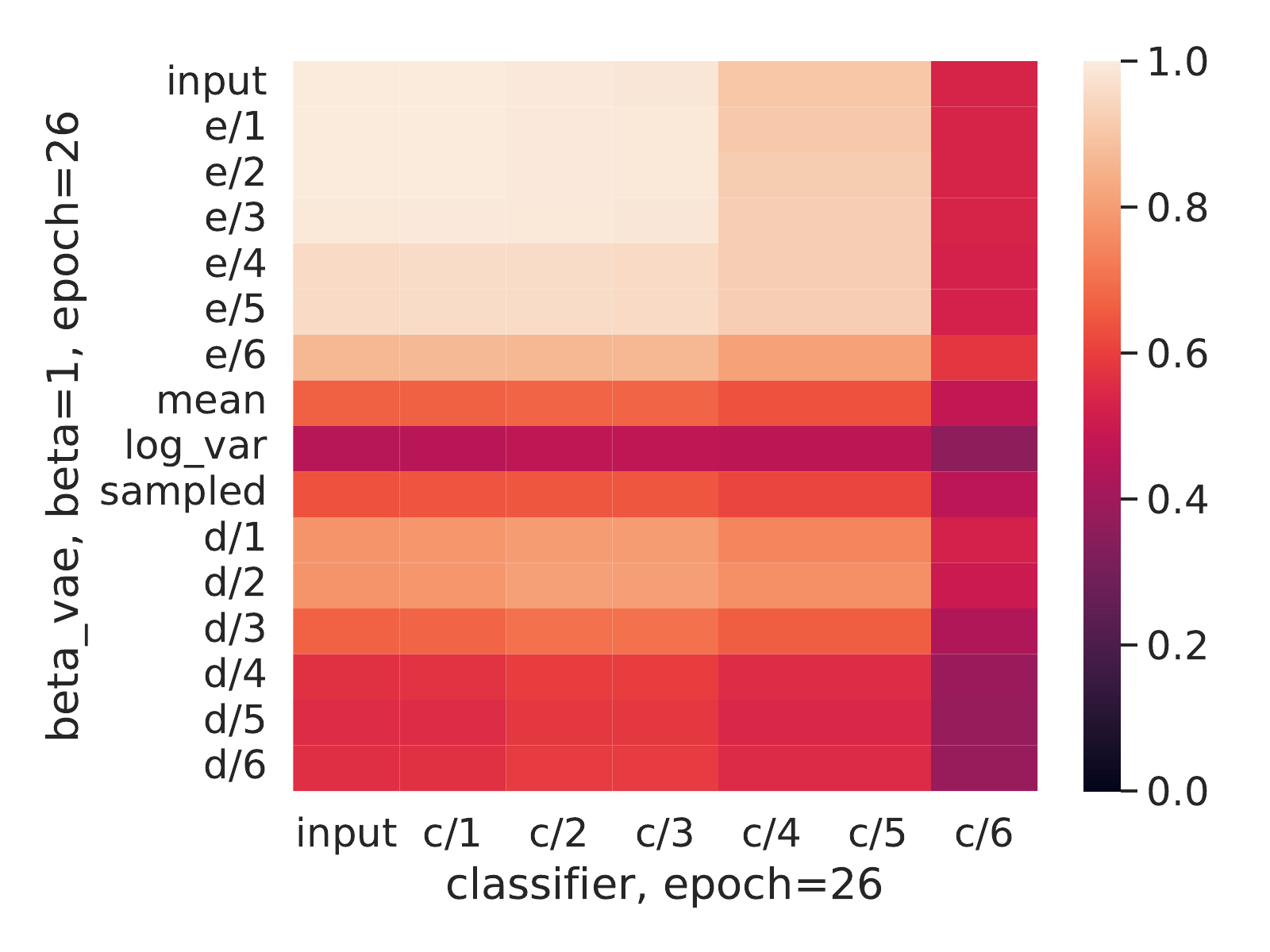}
    }%
    \subcaptionbox{Trained on smallNorb\label{fig:clf-smallnorb}}{
        \includegraphics[width=0.33\textwidth]{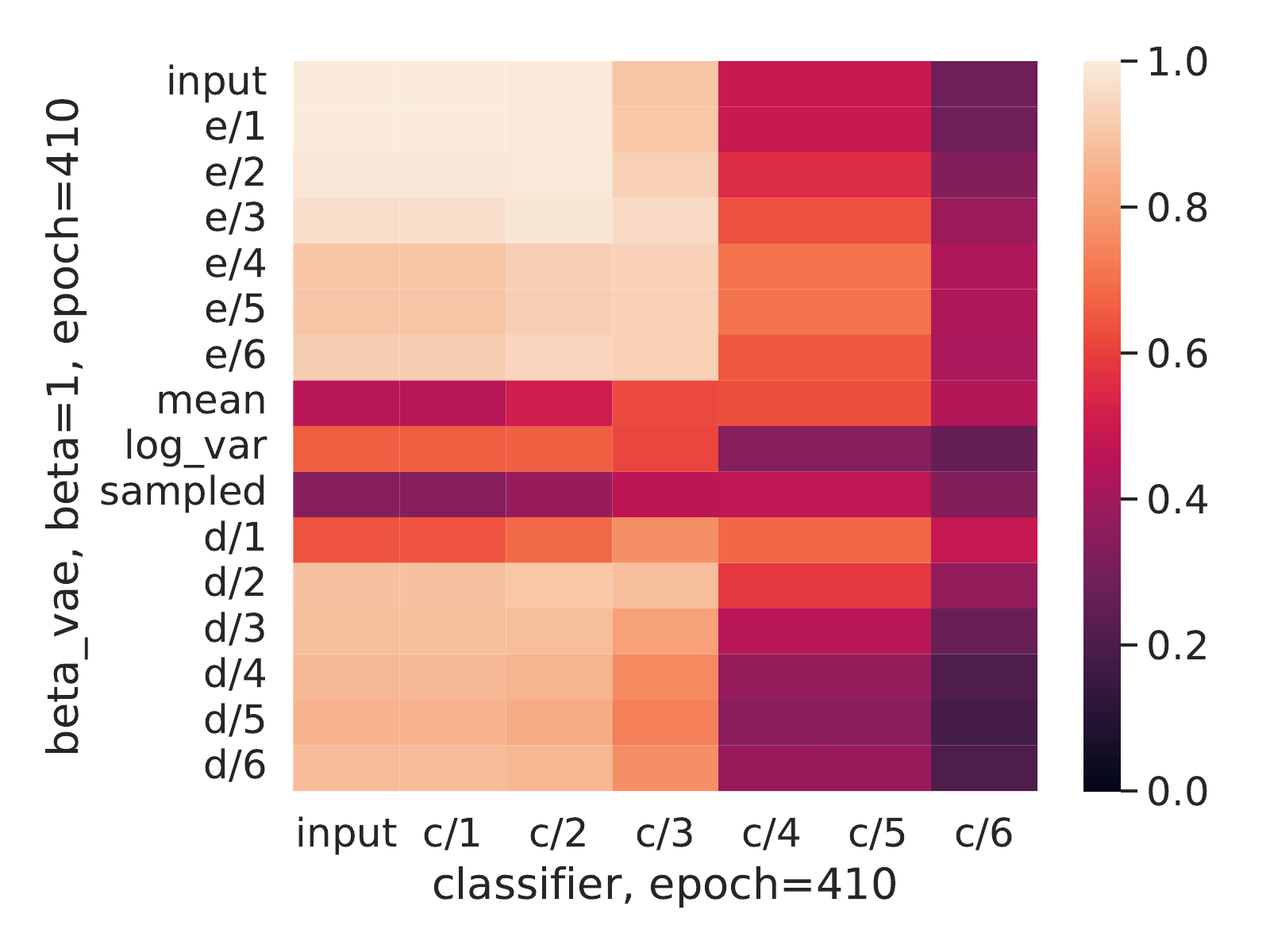}
    }%
    \caption{(a) shows the CKA similarity scores of activations of a classifier and a $\beta$-VAE trained on cars3D with $\beta=1$.
             (b) and (c) show the CKA similarity scores of the same learning objectives and regularisation strengths but trained on dSprites and smallNorb.
             All the results are averaged over 5 seeds.
             We can see that the representational similarity between the layers of the classifiers and of the encoder (top-left quadrant) except mean and variance is very high (CKA stays close to 1).
             However, the mean, variance, sampled representations (center diagonal values) are different from the representations learned by the classifier.
    }
    \label{fig:clf}
\end{figure}

To compare VAEs with classifiers, we used the convolutional architecture of an encoder for classification, replacing
the mean and variance layers by the final classifier layers.
As shown in~\Figref{fig:clf}, we obtain a high representational similarity when comparing VAEs and classifiers indicating,
consistently with the observations of~\citet{Yosinski2015}, that classifiers seem to learn generative features.
This explains why encoders based on pre-trained classifier architectures such as VGG have empirically demonstrated
good performances~\citep{Liu2021} and also suggests that the weights of the pre-trained architecture could be used as-is
without further updates.
While using pre-trained encoders may be beneficial in the context of transfer learning, domain adaptation~\citep{Pan2009},
or simply reconstruction quality~\citep{Liu2021}, one should not expect a dramatic improvement of the training time given that the encoder
is learned very early during the training (see~\Secref{subsec:res-training}).

    \section{Representational similarity of VAEs at different epochs}\label{sec:app-epochs}

The results obtained in~\Secref{subsec:res-training} have shown a high similarity between the encoders at an early stage of training and fully trained.
One can wonder whether these results are influenced by the choice of epochs used in~\Figref{fig:training-sim}.
After explaining our epoch selection process, we show below that it does not influence our results, which are consistent over snapshots taken at
different stages of training.

\paragraph{Epoch selection} For dSprites, we took snapshots of the models at each epoch, but for cars3D and smallNorb, which both train for a
higher number of epochs, it was not feasible computationally to calculate the CKA between every epoch.~We thus
saved models trained on smallNorb every 10 epochs, and models trained on cars3D every 25 epochs.
Consequently, the epochs chosen to represent the early training stage in~\Secref{sec:results} is always the first snapshot taken for each model. Below, we preform additional experiments with a broader range of epoch numbers to show that the results are consistent with our findings in the main paper, and they do not depend on specific epochs.

\begin{figure}[ht!]
    \centering
    \subcaptionbox{Epoch 2\label{fig:dsprites-epoch-1}}{
        \includegraphics[width=0.5\textwidth]{heatmaps/dsprites/dip_vae_ii_5_epoch_2_dip_vae_ii_5_epoch_26}
    }%
    \subcaptionbox{Epoch 7\label{fig:dsprites-epoch-2}}{
        \includegraphics[width=0.5\textwidth]{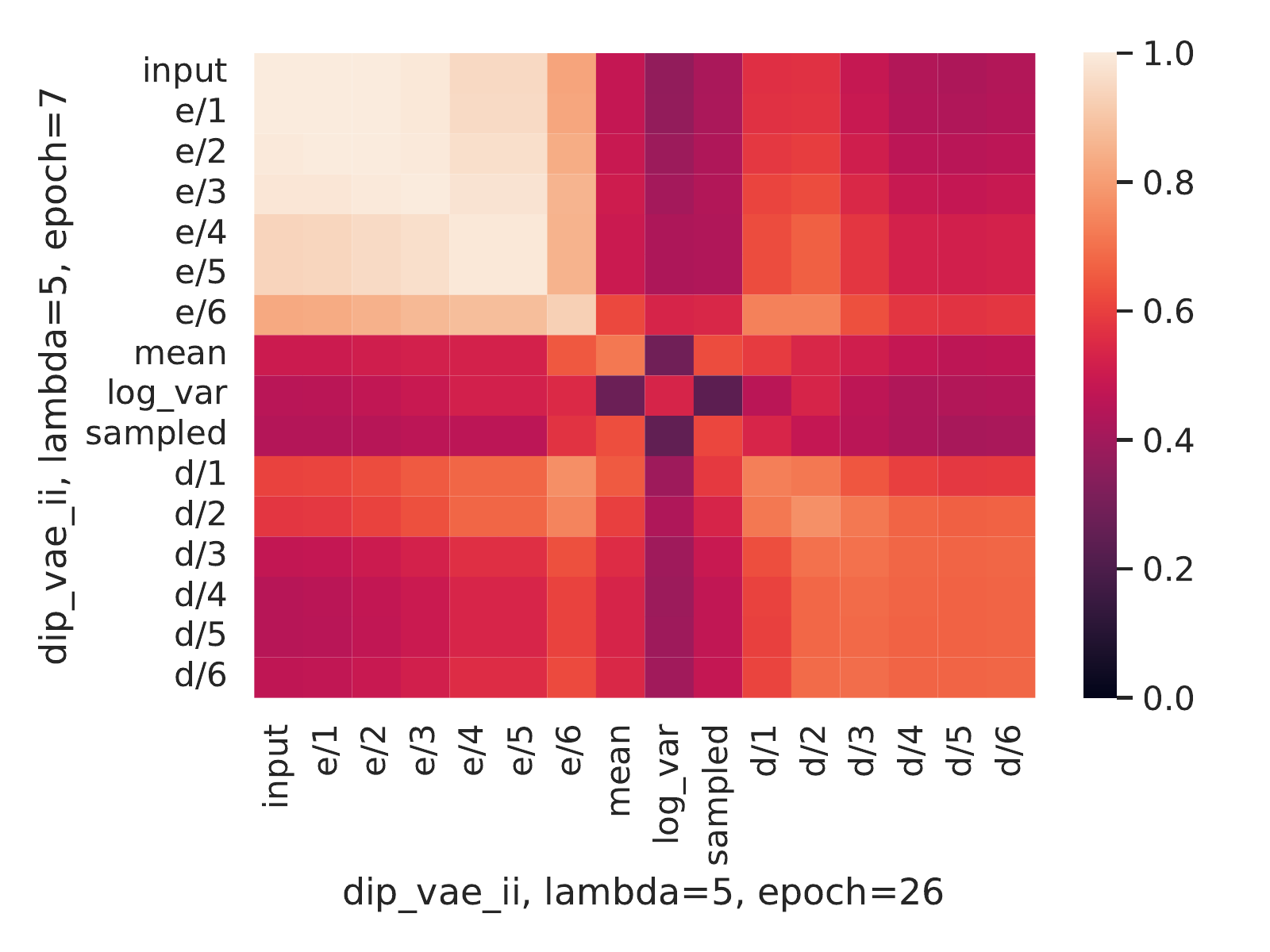}
    }\\
    \subcaptionbox{Epoch 14\label{fig:dsprites-epoch-3}}{
        \includegraphics[width=0.5\textwidth]{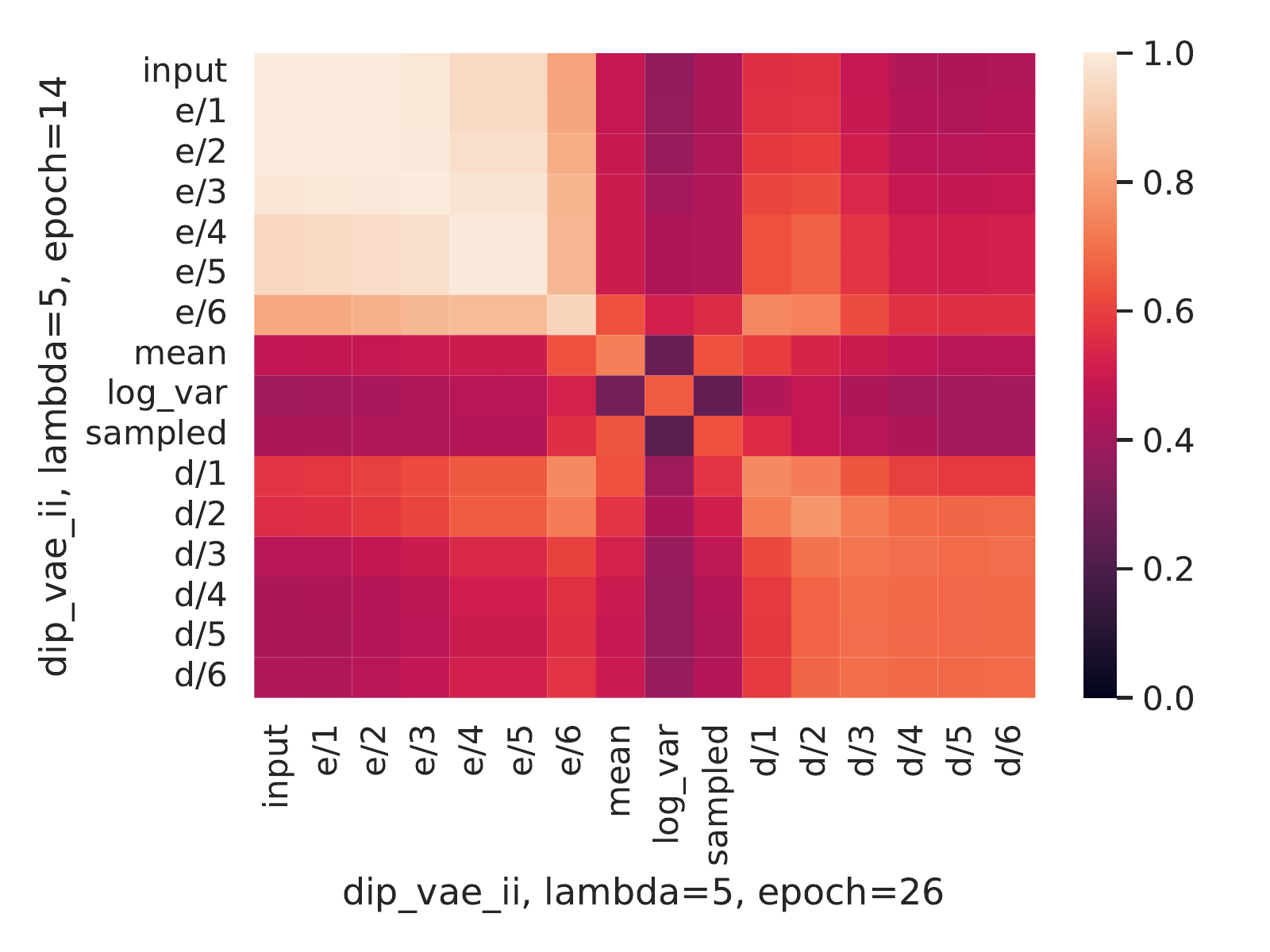}
    }%
    \subcaptionbox{Epoch 19\label{fig:dsprites-epoch-4}}{
        \includegraphics[width=0.5\textwidth]{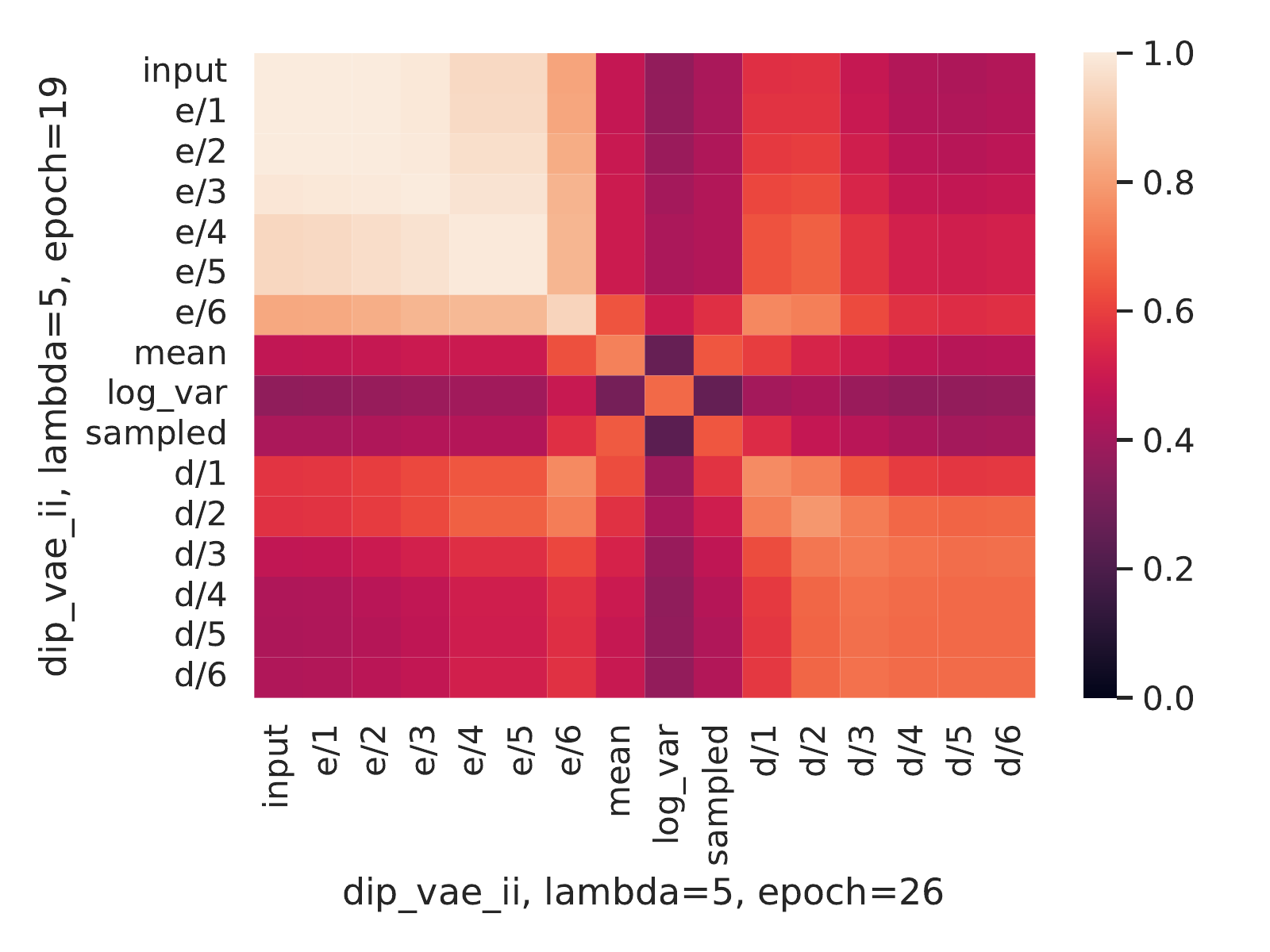}
    }
    \caption{(a), (b), (c), and (d) show the representational similarity between DIP-VAE II after full training, and at epochs 2, 7, 14, and 19, respectively.
    All models are trained on dSprites and the results are averaged over 5 runs.}
    \label{fig:dsprites-epochs}
\end{figure}

\paragraph{Similarity changes over multiple epochs}
In~\Threefigref{fig:dsprites-epochs}{fig:cars-epochs}{fig:norb-epochs}, we can observe the same trend of learning
phases as in~\Figref{fig:training-sim}.~First, the encoder is learned, as shown by the high representational similarity of the upper-left quadrant of
~\Threefigref{fig:dsprites-epoch-1}{fig:cars-epoch-1}{fig:norb-epoch-1}.
Then, the decoder is learned, as shown by the increased representational similarity of the bottom-right quadrant
of~\Threefigref{fig:dsprites-epoch-2}{fig:cars-epoch-2}{fig:norb-epoch-2}.
Finally, further small refinements of the encoder and decoder representations take place in the remaining training time,
as shown by the slight increase of representational similarity in~\Threefigref{fig:dsprites-epoch-3}{fig:cars-epoch-3}{fig:norb-epoch-3},
and~\Threefigref{fig:dsprites-epoch-4}{fig:cars-epoch-4}{fig:norb-epoch-4}.

\begin{figure}[ht!]
    \centering
    \subcaptionbox{Epoch 25\label{fig:cars-epoch-1}}{
        \includegraphics[width=0.5\textwidth]{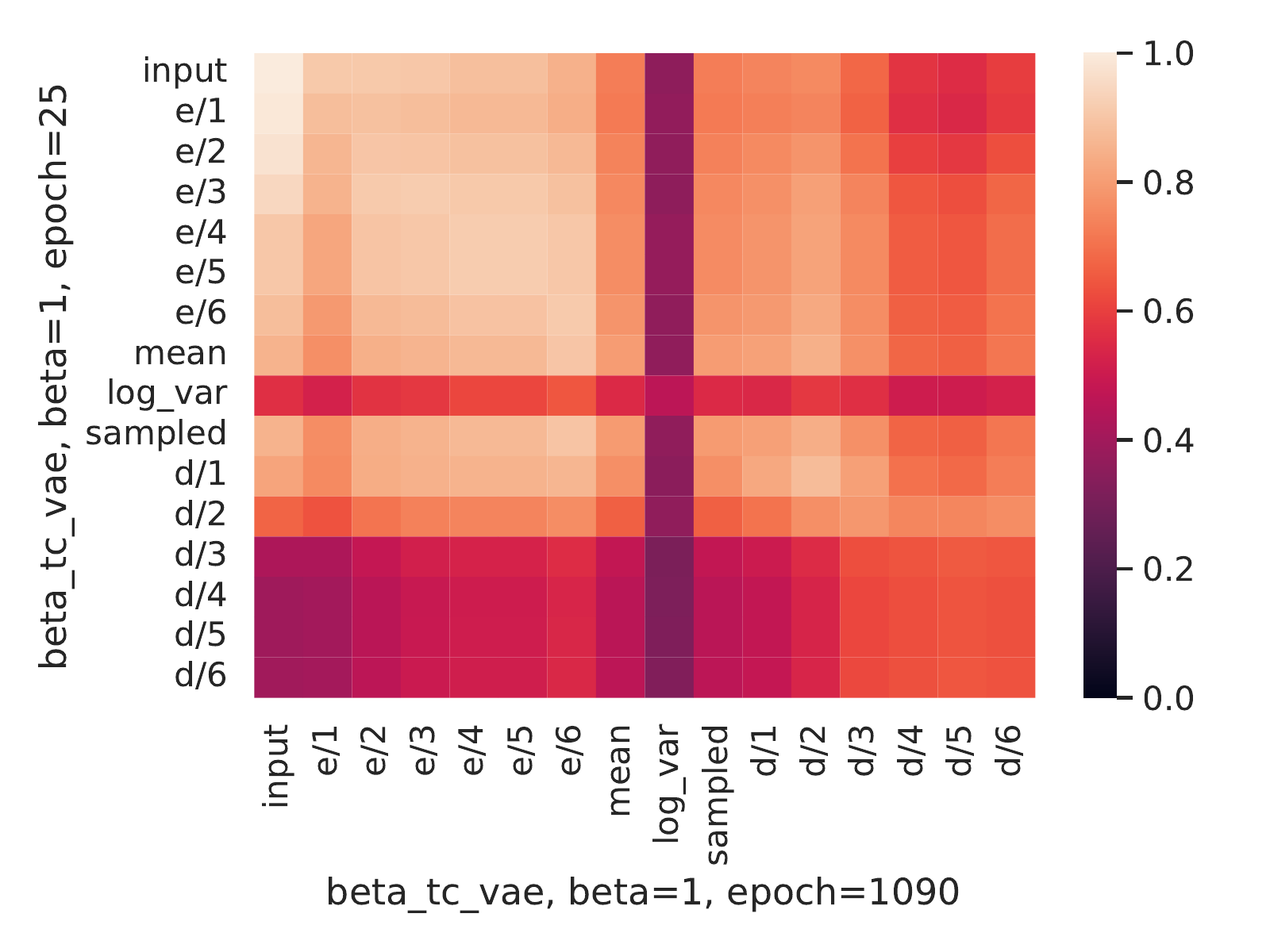}
    }%
    \subcaptionbox{Epoch 292\label{fig:cars-epoch-2}}{
        \includegraphics[width=0.5\textwidth]{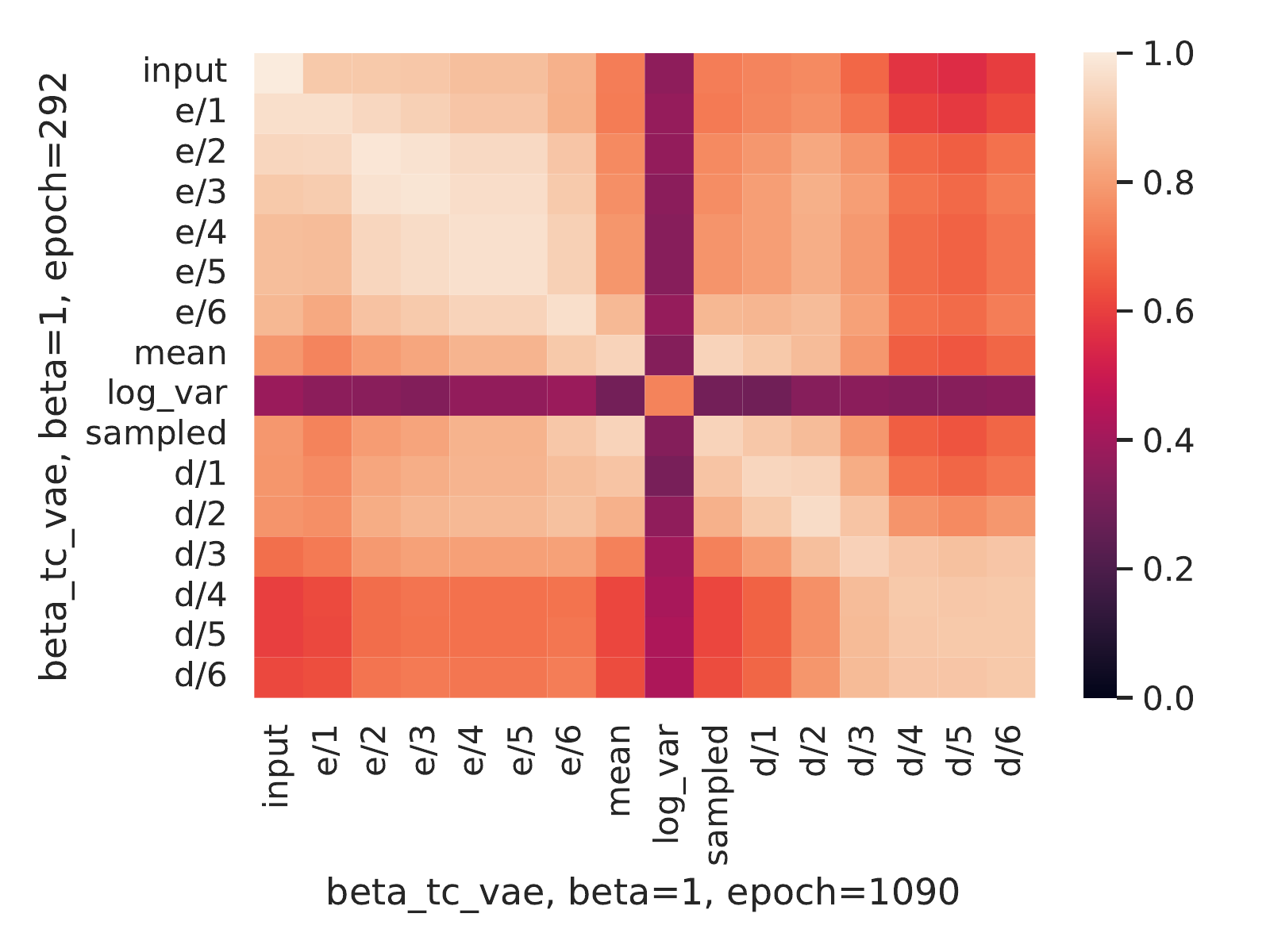}
    }\\
    \subcaptionbox{Epoch 559\label{fig:cars-epoch-3}}{
        \includegraphics[width=0.5\textwidth]{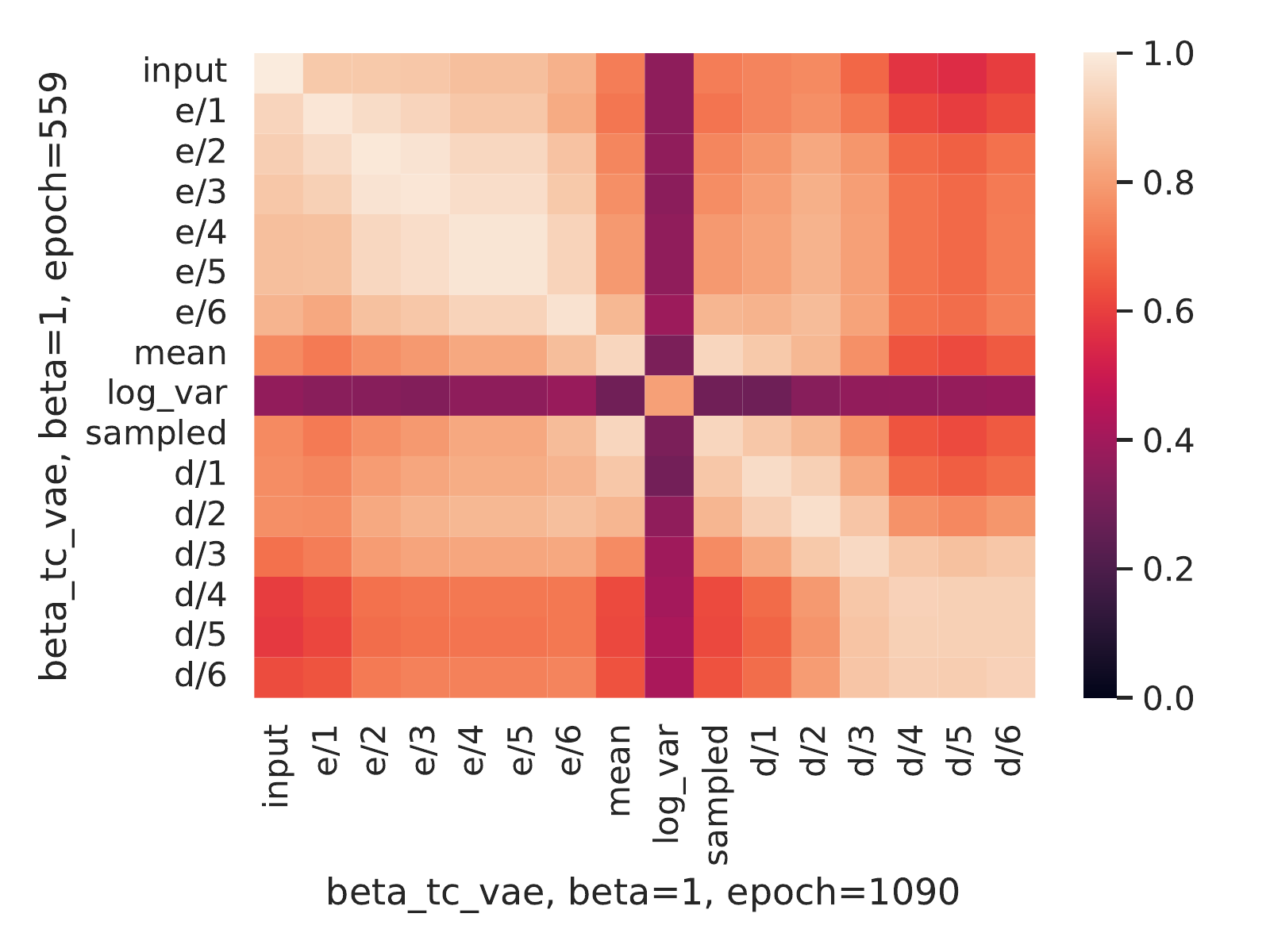}
    }%
    \subcaptionbox{Epoch 826\label{fig:cars-epoch-4}}{
        \includegraphics[width=0.5\textwidth]{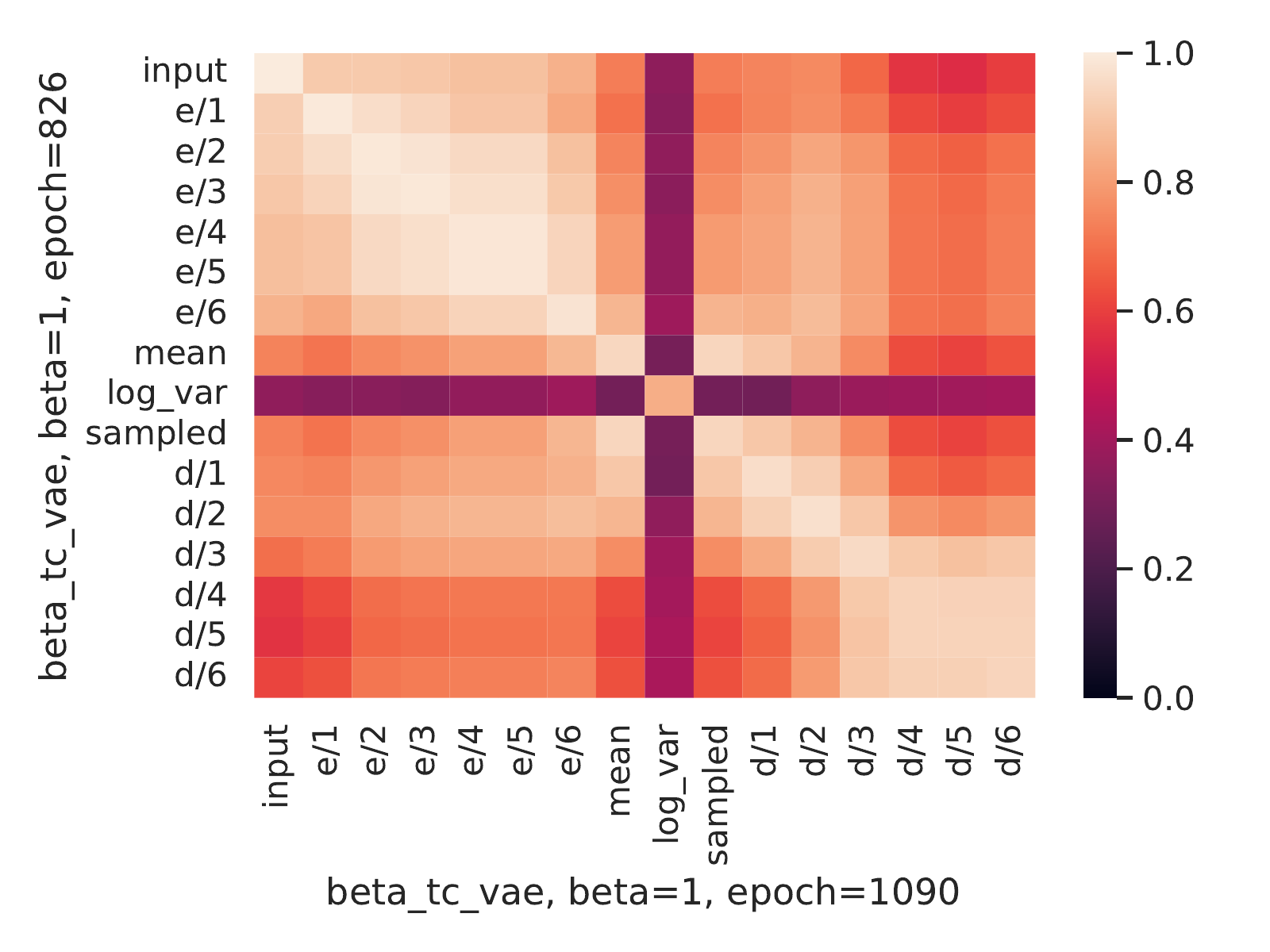}
    }
    \caption{(a), (b), (c), and (d) show the representational similarity between $\beta$-TC VAE after full training, and at epochs 25, 292, 559, and 826, respectively.
    All models are trained on cars3D and the results are averaged over 5 runs.}
    \label{fig:cars-epochs}
\end{figure}

\begin{figure}[ht!]
    \centering
    \subcaptionbox{Epoch 10\label{fig:norb-epoch-1}}{
        \includegraphics[width=0.5\textwidth]{heatmaps/smallnorb/annealed_vae_5_epoch_10_annealed_vae_5_epoch_410}
    }%
    \subcaptionbox{Epoch 110\label{fig:norb-epoch-2}}{
        \includegraphics[width=0.5\textwidth]{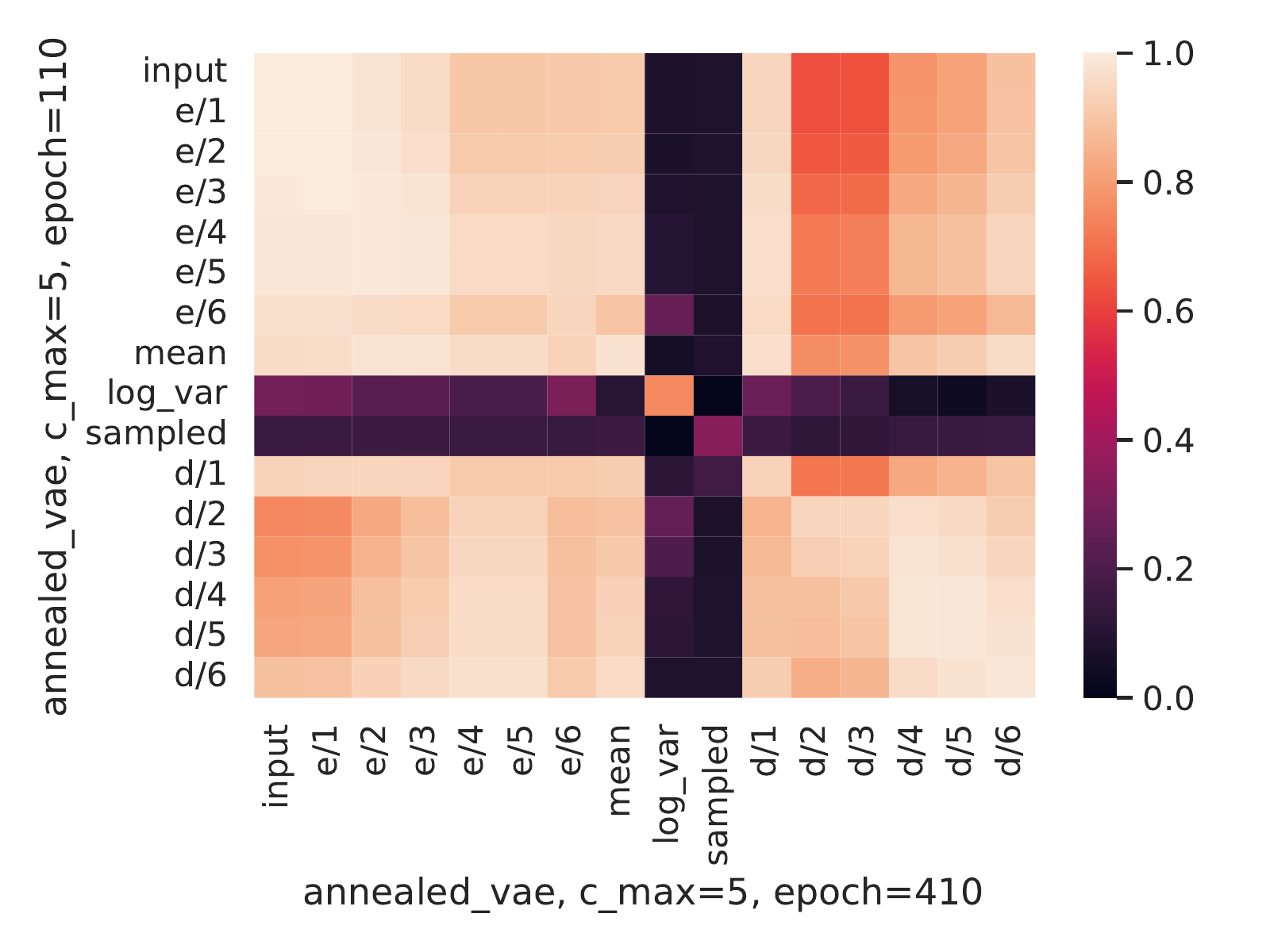}
    }\\
    \subcaptionbox{Epoch 210\label{fig:norb-epoch-3}}{
        \includegraphics[width=0.5\textwidth]{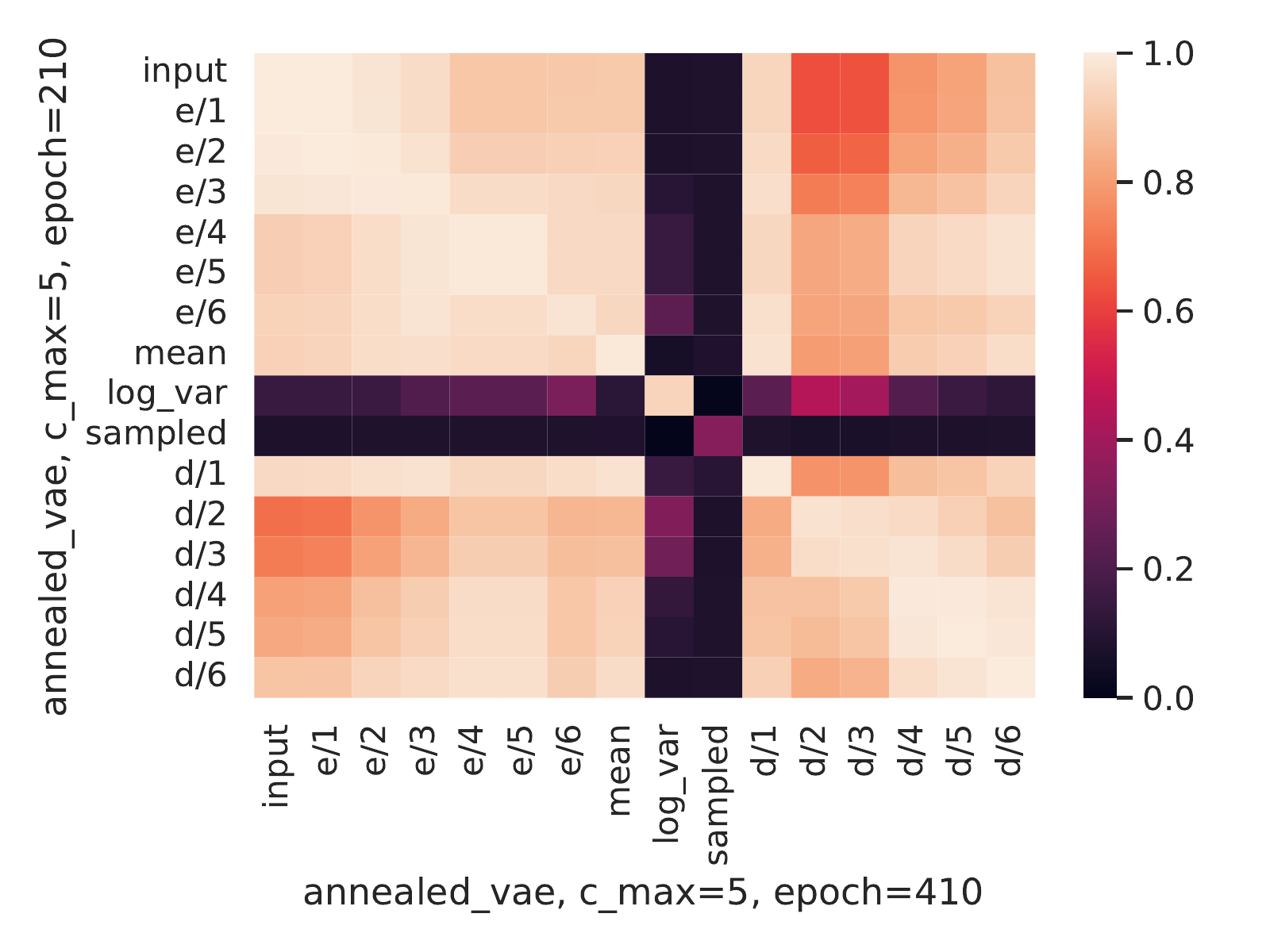}
    }%
    \subcaptionbox{Epoch 311\label{fig:norb-epoch-4}}{
        \includegraphics[width=0.5\textwidth]{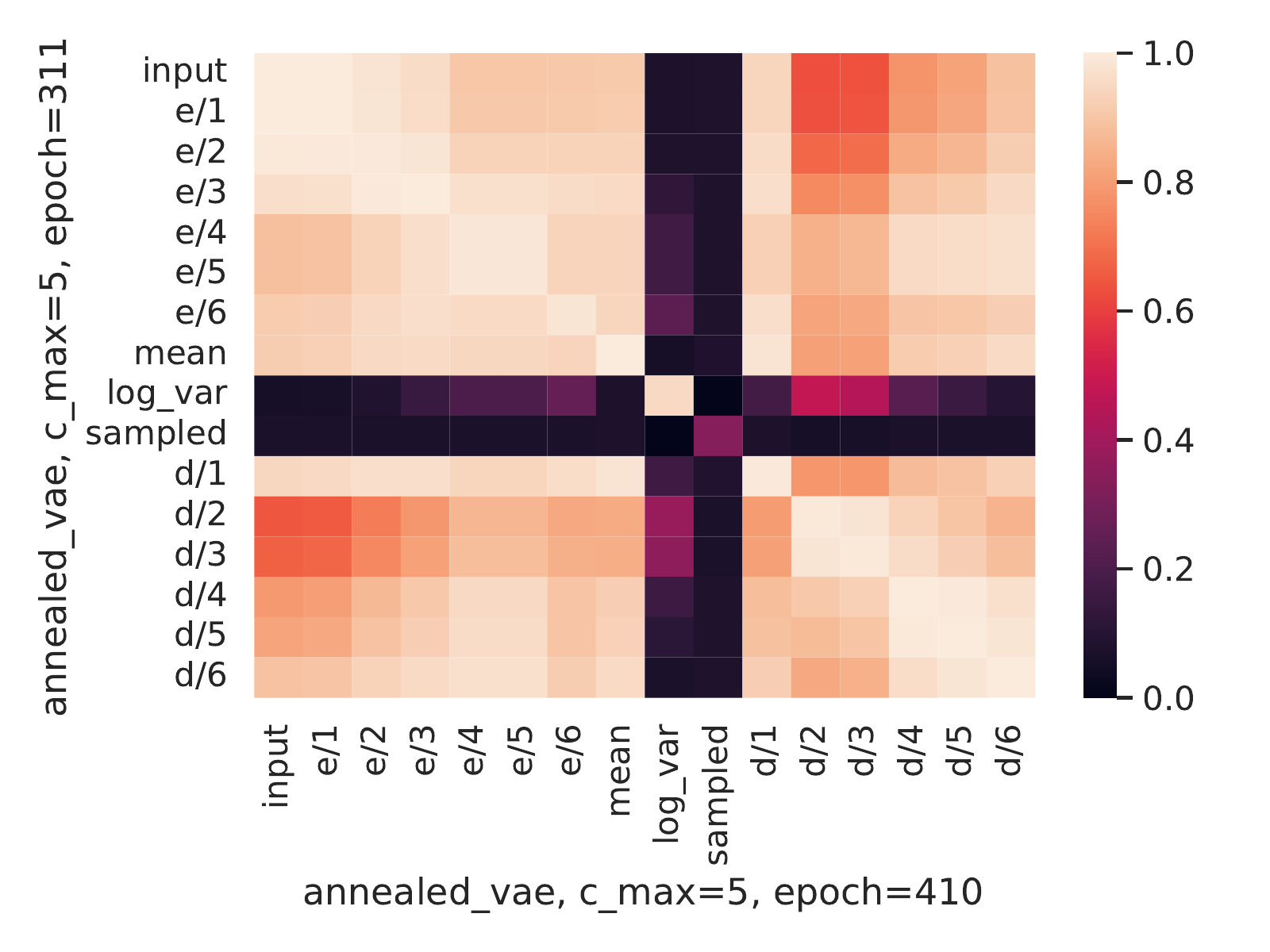}
    }
    \caption{(a), (b), (c), and (d) show the representational similarity between Annealed VAE after full training, and at epochs 10, 110, 210, and 311, respectively.
    All models are trained on smallNorb and the results are averaged over 5 runs.}
    \label{fig:norb-epochs}
\end{figure}
    \clearpage
    \section{Convergence rate of different VAEs}\label{sec:app-convergence}
\begin{figure}[ht!]
    \centering
    \subcaptionbox{Convergence on cars3D\label{fig:conv-cars}}{
        \includegraphics[width=\textwidth]{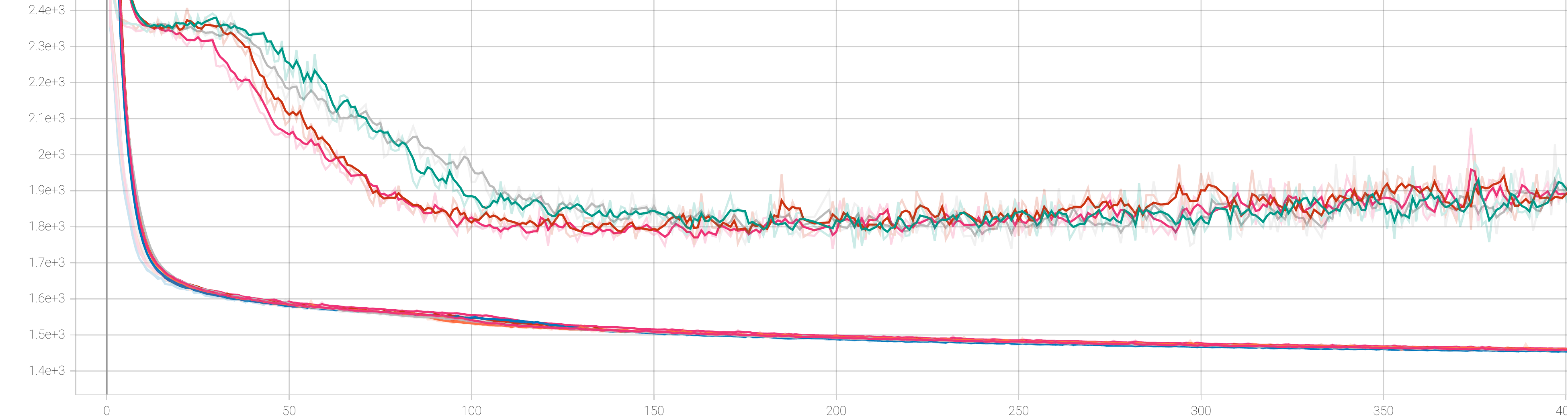}
    }\\
    \subcaptionbox{Convergence on dSprites\label{fig:conv-dsprites}}{
        \includegraphics[width=\textwidth]{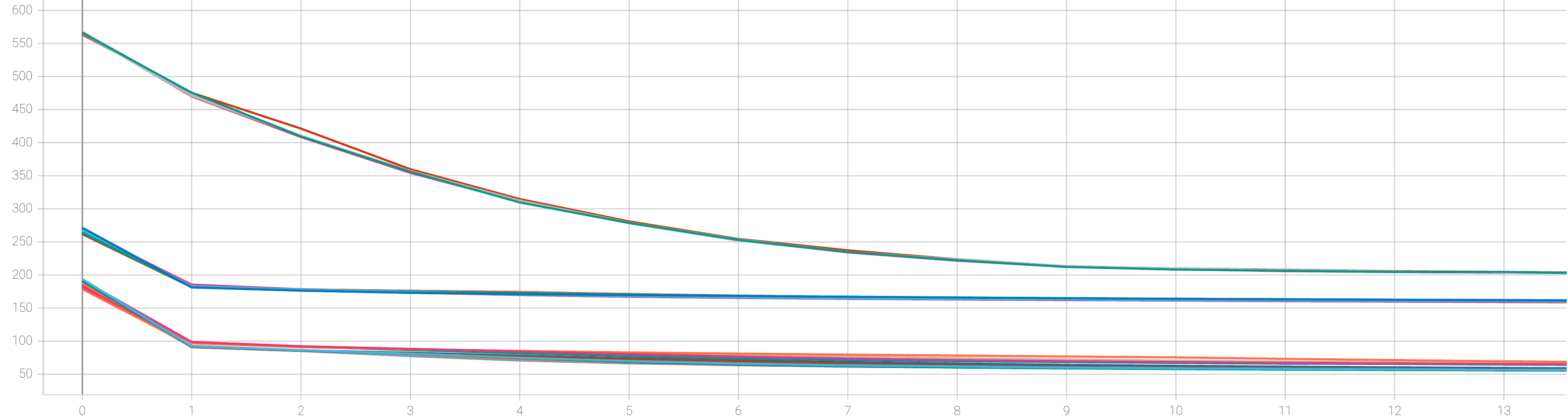}
    }\\
    \subcaptionbox{Convergence on smallNorb\label{fig:conv-smallnorb}}{
        \includegraphics[width=\textwidth]{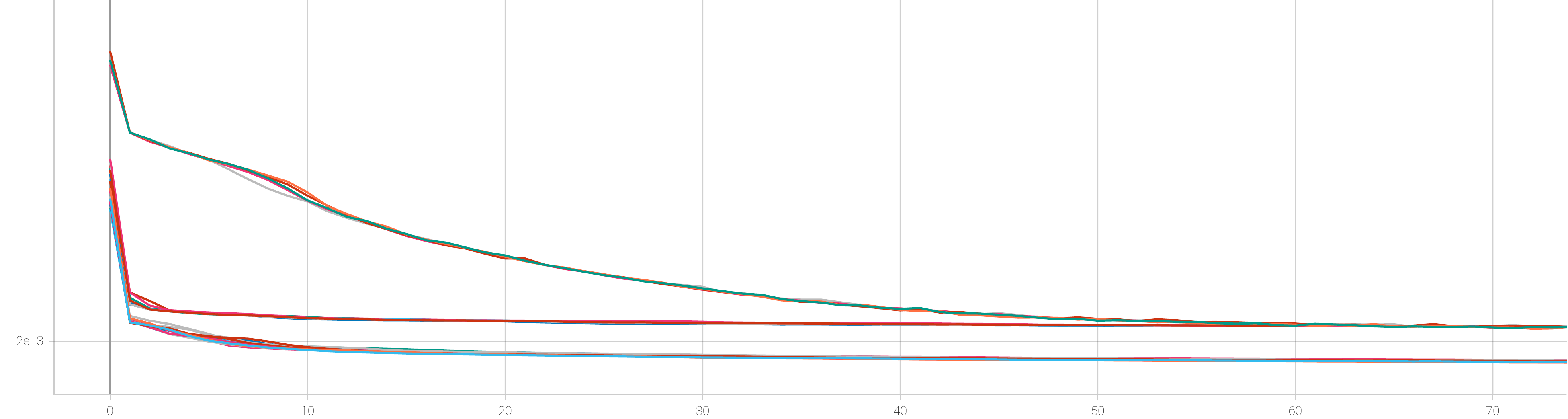}
    }\\
    \caption{ In (a), (b), and (c), we show the model loss of each model that converged when trained on cars3D, dSprites, and smallNorb, respectively.
    For each learning objective, we display 5 runs of the least and most regularised versions.
    }
    \label{fig:conv}
\end{figure}

We can see in~\Figref{fig:conv} that all the models converge at the same epoch, with less regularised models reaching lower losses.
While annealed VAEs start converging together with the other models, they then take longer to plateau, due to the annealing process.
We can see them distinctly in the upper part of~\Figref{fig:conv}.
Overall, the epochs at which the models start to converge are consistent with our choice of epoch for early training in~\Secref{subsec:res-training}.

\end{document}